\documentclass{article}
\usepackage{iclr2027_conference,times}

\usepackage{array,bm,amsmath,amssymb,mathrsfs,amsthm,mathtools,bbm}
\usepackage{booktabs,longtable,multirow}
\usepackage{algorithm}
\usepackage{algpseudocode}
\usepackage{graphicx}
\usepackage{xcolor}
\usepackage{enumitem}
\usepackage{adjustbox}
\usepackage{placeins}
\usepackage{microtype}
\usepackage{url}
\usepackage{hyperref}
\usepackage{wrapfig}

\allowdisplaybreaks
\mathtoolsset{showonlyrefs}


\newcommand{\R}{\mathbb{R}}

\newcommand{\E}{\mathbb{E}}
\newcommand{\Pp}{\mathbb{P}}

\newcommand{\argmin}{\mathop{\mathrm{arg\,min}}}

\newcommand{\cA}{\mathcal{A}}
\newcommand{\cB}{\mathcal{B}}

\newcommand{\cE}{\mathcal{E}}

\newcommand{\cL}{\mathcal{L}}

\newcommand{\cR}{\mathcal{R}}
\newcommand{\wh}{\widehat}
\newcommand{\wt}{\widetilde}

\newcommand{\one}{\mathbf{1}}
\newcommand{\dd}{\mathrm{d}}

\newcommand{\WD}{\operatorname{WD}}
\newcommand{\WDsup}{\ensuremath{\operatorname{WD}_{\mathrm{sup}}}}
\newcommand{\CE}{\operatorname{CE}}

\newtheorem{theorem}{Theorem}[section]
\newtheorem{lemma}[theorem]{Lemma}
\newtheorem{proposition}[theorem]{Proposition}
\newtheorem{corollary}[theorem]{Corollary}
\theoremstyle{definition}
\newtheorem{assumption}[theorem]{Assumption}
\newtheorem{definition}[theorem]{Definition}

\theoremstyle{remark}
\newtheorem{remark}[theorem]{Remark}

\title{Beyond Sub-Gaussian Detector Scores: Robust Weighted Profile-Loss Change Point Detection for Human-LLM Text Segmentation}
\author{
Wan Tian\textsuperscript{1*} \ \ 
Zhongyi Li\textsuperscript{2*} \ \ 
Yawen Li\textsuperscript{3}  \ \ 
Rui Zhang\textsuperscript{4} \ \ 
Yijie Peng\textsuperscript{5$\dagger$}\ \ 
Fuzhen Zhuang\textsuperscript{2$\dagger$} \\[1ex]
\textsuperscript{1}Peking University \quad
\textsuperscript{2}Beihang University \quad
\textsuperscript{3}Capital Normal University  \\
\textsuperscript{4}Renmin University of China \quad
\textsuperscript{5}Nanjing University \\[0.5ex]
\textsuperscript{*}These authors contributed equally to this work. \quad
\textsuperscript{$\dagger$}Corresponding authors. \\
\texttt{Correspondence: pengyijie@nju.edu.cn; zhuangfuzhen@buaa.edu.cn}
}
\iclrfinalcopy

\begin{document}
\maketitle

\begin{abstract}
Mixed human-LLM documents require locating authorship transitions from detector scores whose reliability varies across text units.
Existing weighted mean contrasts are vulnerable to extreme scores, while directly replacing means with robust centers obscures how a misplaced boundary changes the population objective.
We propose Robust Weighted Profile-Loss Change Point Detection (RWCP), which combines capped reliability weights, Huber profile gains, and narrowest-over-threshold search in reliability coordinates.
Our key analysis expresses the population gap between a true and a displaced split as a merge cost, avoiding a closed-form solution for the nonlinear center of a mixed segment.
Under explicit curvature, spacing, and dependence conditions, core RWCP recovers the number of changes and localizes their boundaries; its quadratic-loss limit recovers squared weighted CUSUM.
We also study RWCP-R, a separately evaluated decoder that shares source centers across nonadjacent passages.
Across five retrospective cached-score benchmark families, core RWCP reduces family-macro WindowDiff by 17.6\% relative to weighted change-point detection, and RWCP-R lowers it further.
Boundary recovery improves most clearly for isolated changes, while both fixed configurations miss changes in collaborative and densely alternating text.
\end{abstract}

\section{Introduction}
\label{sec:introduction}

Mixed human--LLM writing calls for locating authorship changes rather than labeling an entire document. Human passages can alternate with generated continuations or model-assisted edits, so the output is a sequence of source-homogeneous segments and boundaries.
Mapping each sentence or short unit \(X_i\) to a detector score \(Y_i=\phi(X_i)\) makes this an ordered change-point problem. WCP uses the unequal reliability of these scores \citep{li2026segmenting}, but weighting alone does not control extreme observations or context-induced dependence.

\begin{figure}[t]
\centering
\includegraphics[width=0.96\linewidth]{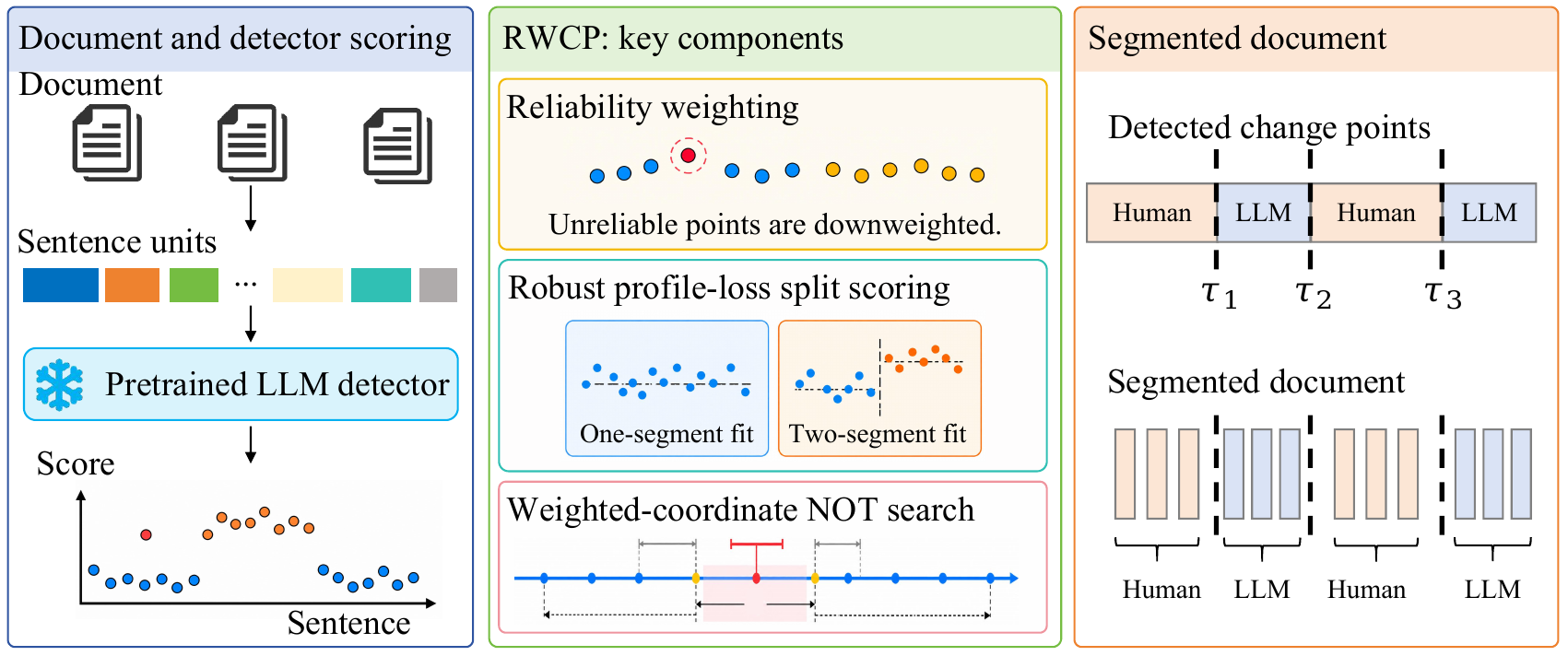}
\caption{Core RWCP: capped reliability weights, Huber profile gains, and weighted-coordinate narrowest-over-threshold search. The recovery theorem covers this procedure. RWCP-R separately decodes recurrent source states using \eqref{eq:recurrent-objective}.}
\label{fig:rwcp-overview}
\end{figure}

Rare tokens, code, equations, paraphrases, and short units can yield extreme scores, while underestimated variance proxies can give a single unit excessive weight.
Huber scores bound their influence, but substituting Huber centers into mean-based CUSUM does not preserve its population decomposition: an incorrect split mixes two source locations, whose optimum is generally not their weighted arithmetic mean. The analysis must compare fitted segments without assuming a formula for that mixed center.

The \emph{profile gain} is the reduction in optimized Huber loss obtained by splitting an interval. With one change, its population difference between the true and a displaced boundary equals a merge cost; local risk curvature makes this gap proportional to misplaced reliability mass and squared jump size.
Capped weights control leverage, bounded Huber scores support dependence-aware concentration without sub-Gaussian raw scores, and reliability-coordinate intervals isolate changes by information mass. Guarded recursion converts these properties into count and localization guarantees.

Core RWCP combines robust weighted profile gains with weighted-coordinate narrowest-over-threshold search; in the unconstrained quadratic limit, its gain is exactly squared weighted CUSUM.
We prove finite-sample exact count recovery and weighted localization under explicit curvature, central-mass, spacing, and dependence conditions, including the logarithmic cost of geometric mixing. The merge-cost analysis distinguishes this result from a direct robust-center substitution in an existing contrast.

Recurring human and LLM passages offer a second opportunity: nonadjacent passages can share a source state. We study RWCP-R, a distinct decoder with shared centers and a fitted residual scale.
We now evaluate an analytic-threshold implementation of core RWCP on the same cached scores as WCP before assessing RWCP-R separately. Core RWCP reduces window error most clearly on controlled single-change texts, but its fixed threshold often misses changes in real collaborative and dense-change documents; RWCP-R improves average window error further without removing this count-recovery limitation.
The theorem covers core RWCP under its stated assumptions, not the empirical RWCP-R decoder or the claim that either fixed configuration satisfies those assumptions on detector scores.

Huber profile costs are established in penalized segmentation \citep{fearnhead2018changepoint}, narrowest-over-threshold search supplies interval selection \citep{baranowski2019not}, and WCP supplies the reliability-weighted detector-score setting \citep{li2026segmenting}.
Our technical step is to analyze the weighted \emph{difference of optimized profile gains} through a merge-cost margin and carry it through dependence-aware control, reliability-coordinate isolation, and guarded recursion. Related work, full protocols, and proofs appear in Appendices~\ref{sec:related}--\ref{app:theoretical-proofs}.

\section{Robust Weighted Profile Segmentation}
\label{sec:method}

We first define a robust gain for testing whether one location or two better describe an interval. We then use that gain in reliability-coordinate interval search; a separate recurrent objective shares centers across passages assigned to the same source state.

\subsection{Robust locations, reliability, and profile gains}

Let \(0=\tau_0<\tau_1<\cdots<\tau_K<\tau_{K+1}=N\) denote unknown boundaries.
Observation \(i\) has a deterministic reliability weight \(w_i>0\), scale \(s_i=w_i^{-1/2}\), and segment index \(z(i)\).
For cutoff \(c>0\), the Huber loss and score are
\[
\rho_c(u)=
\begin{cases}
u^2/2,&|u|\leq c,\\
c|u|-c^2/2,&|u|>c,
\end{cases}
\qquad
\psi_c(u)=\operatorname{sign}(u)\min\{|u|,c\}.
\]

\begin{definition}[Piecewise common Huber location]
\label{def:piecewise-huber}
There are locations \(\theta_1,\ldots,\theta_{K+1}\) in a compact interval \(\Theta\) such that, whenever \(\tau_{j-1}<i\leq\tau_j\), \(\theta_j\) minimizes \(Q_i(\theta)=\E\rho_c\{\sqrt{w_i}(Y_i-\theta)\}\).
For interior unique minimizers this is equivalent to
\begin{equation}
\E\!\left[\sqrt{w_i}\,
\psi_c\{\sqrt{w_i}(Y_i-\theta_j)\}\right]=0.
\label{eq:score-centering}
\end{equation}
Adjacent jump sizes are \(\kappa_j=|\theta_{j+1}-\theta_j|\).
\end{definition}

The location is defined by the detector-score risk rather than by an ordinary expectation.
A symmetric location family \(Y_i=\theta_{z(i)}+\sigma_i\xi_i\) is sufficient, but asymmetric distributions are allowed whenever the centering condition in \eqref{eq:score-centering} holds.
Given a variance proxy \(\sigma_i^2\), the capped oracle weight is
\begin{equation}
w_i^\circ=\Pi_{[w_{\min},w_{\max}]}(\sigma_i^{-2}),
\qquad
\Pi_{[a,b]}(u)=\min\{\max(u,a),b\}.
\label{eq:oracle-weight}
\end{equation}
Practical weights replace \(\sigma_i^2\) by an externally estimated proxy and add a small denominator ridge.
Capping in \eqref{eq:oracle-weight} prevents a single underestimated variance from receiving unbounded leverage.
Write \(S_{a:b}^w=\sum_{i=a}^b w_i\) and \(d_w(u,v)=\sum_{\min(u,v)<i\leq\max(u,v)}w_i\).

For an interval \(A\), define \(\widehat{\cL}_A(\theta)=\sum_{i\in A}\rho_c\{\sqrt{w_i}(Y_i-\theta)\}\) and let \(\widehat\theta_A\) minimize it over \(\Theta\).
For \(s\leq b<e\), RWCP uses the profile improvement
\begin{equation}
\widehat\Gamma_{s,e}(b)
=2\!\left\{
\widehat\cL_{s:e}(\widehat\theta_{s:e})
-\widehat\cL_{s:b}(\widehat\theta_{s:b})
-\widehat\cL_{b+1:e}(\widehat\theta_{b+1:e})
\right\}.
\label{eq:profile-gain}
\end{equation}
The gain in \eqref{eq:profile-gain} is nonnegative because the one-location model is nested in the split model. Its population analogue makes a displaced split pay for merging two distinct source locations, the mechanism quantified in Theorem~\ref{thm:population-margin}.

\begin{proposition}[Quadratic-loss reduction]
\label{prop:quadratic-reduction}
If \(\rho_c\) is replaced by \(\rho_\infty(u)=u^2/2\) and the three fits are unconstrained, or their weighted means all lie in \(\Theta\), then
\begin{equation}
\widehat\Gamma_{s,e}(b)
=\frac{S_{s:b}^wS_{b+1:e}^w}{S_{s:e}^w}
\left(\bar Y_{s:b}^w-\bar Y_{b+1:e}^w\right)^2
=\{W_{s,e}^Y(b)\}^2,
\label{eq:wcp-equivalence}
\end{equation}
where \(\bar Y_{a:b}^w=\sum_{i=a}^bw_iY_i/S_{a:b}^w\) and \(W_{s,e}^Y(b)\) is the weighted CUSUM.
\end{proposition}

Active parameter constraints introduce projection terms (Appendix~\ref{app:reductions}). Thus RWCP retains the familiar WCP contrast when its loss is quadratic, while the finite-cutoff analysis uses bounded Huber scores in place of a raw-score tail condition.

\subsection{Weighted-coordinate search and recurrent states}
\label{subsec:weighted-intervals}

The reliability mass used by the profile gain also defines the search geometry. RWCP samples intervals uniformly in the cumulative coordinate \(W(t)=\sum_{i\leq t}w_i\), rather than uniformly in sentence index.
Within \(I=[s,e]\), it scans splits whose two sides each have mass at least \(m_N\), computes \(T(I)=\max_b\widehat\Gamma_{s,e}(b)\), and retains intervals with \(T(I)>r_N\).
Among them it selects the interval with the least reliability mass, reports its maximizing split \(b^\star\), and recurses.
This is the narrowest-over-threshold principle \citep{baranowski2019not} in the information geometry supplied by the detector.

After accepting a boundary, a guard of mass \(g_N\) prevents duplicate discoveries while preserving well-separated neighboring changes.
For a current search range \([a,b]\), define the nearest mass crossings
\begin{equation}
\begin{aligned}
t_L&=\max\{t\in[a,b^\star]:S_{t:b^\star}^w\geq g_N\},&
\ell_g&=t_L-1,\\
t_R&=\min\{t\in[b^\star+1,b]:S_{b^\star+1:t}^w\geq g_N\},&
r_g&=t_R+1.
\end{aligned}
\label{eq:guard-endpoints}
\end{equation}
If the left or right crossing does not exist, set \(\ell_g=a-1\) or \(r_g=b+1\), respectively. Recursion continues on the nonempty ranges \([a,\ell_g]\) and \([r_g,b]\). Each removed side has mass below \(g_N+w_{\max}\).
The theorem uses the analytic threshold \(r_N=C_r\Lambda_N(\delta)\).
In practice, a block-multiplier alternative forms clipped influence residuals from a conservative pilot, multiplies overlapping residual blocks by Gaussian weights, and takes the conditional quantile of the maximum over the same fixed candidate collection.
This calibration is not part of the recovery theorem.

When source states recur, nonadjacent passages can inform a common center. RWCP-R uses this structure through a separate recurrent-state objective; it replaces the core interval search and does not invoke its threshold or guard.
\label{subsec:recurrent-profiles}
With robustly standardized scores \(\widetilde Y_i\), compressed capped weights \(\widetilde w_i\), \(S\) source states, centers \(\mu\), and scale \(\sigma\geq\sigma_{\min}\), it seeks a local minimum of
\begin{equation}
\mathcal J(z,\mu,\sigma)
=\sum_{i=1}^N\rho_c\!\left(
\frac{\sqrt{\widetilde w_i}(\widetilde Y_i-\mu_{z_i})}{\sigma}
\right)
+N\log\sigma
+\lambda\sum_{i=2}^N\one\{z_i\ne z_{i-1}\}.
\label{eq:recurrent-objective}
\end{equation}
For fixed centers and scale, Viterbi decoding solves the state update in \(O(NS^2)\); for fixed states, weighted Huber fits update the centers and a monotone score equation updates the scale.
Exact conditional updates cannot increase \eqref{eq:recurrent-objective}; the joint problem is nonconvex and the finite-iteration solver has no global-optimality guarantee.
Here \(\sigma\) is common to all states. With residuals \(r_i=\sqrt{\widetilde w_i}(\widetilde Y_i-\mu_{z_i})/\sigma\), an interior scale update solves \(\sum_i\psi_c(r_i)r_i=N\); otherwise the constrained optimum is \(\sigma_{\min}\). We declare \(S=3\) for CoAuthor and \(S=2\) for the controlled tasks. Boundaries are state transitions; \(S\) specifies source classes, not their number of occurrences. Appendix~\ref{subsec:practical-configuration} gives the finite-iteration algorithm.
Section~\ref{sec:theory} analyzes core RWCP; Section~\ref{sec:experiments} evaluates the recurrent decoder as a separate empirical extension.

\section{Theoretical Guarantees}
\label{sec:theory}

The recovery argument follows the algorithm's four steps: profile-loss geometry supplies a population margin; bounded Huber scores and empirical curvature preserve it in the observed data; reliability-coordinate intervals isolate each change; and a guard prevents duplicate detections during recursion.
We state the assumptions and main result here, with verification results, lower bounds, constants, and proofs in Appendix~\ref{app:theoretical-proofs}.

\begin{assumption}[Robust locations]
\label{assump:location}
Definition~\ref{def:piecewise-huber} and \eqref{eq:score-centering} hold with finite risks. There is a fixed radius \(r_0>0\) such that \(\operatorname{dist}(\theta_j,\partial\Theta)\geq r_0\) for every true center.
\end{assumption}

\begin{assumption}[Risk curvature]
\label{assump:curvature}
There are fixed \(0<m_0\leq M_0<\infty\) such that, for every \(i\) in segment \(j\) and every \(|\theta-\theta_j|\leq r_0\),
\begin{equation}
\frac{m_0w_i}{2}(\theta-\theta_j)^2
\leq Q_i(\theta)-Q_i(\theta_j)
\leq\frac{M_0w_i}{2}(\theta-\theta_j)^2.
\label{eq:risk-curvature}
\end{equation}
\end{assumption}

\begin{assumption}[Reliability]
\label{assump:weights}
The deterministic weights satisfy \(0<w_{\min}\leq w_i\leq w_{\max}<\infty\), and the minimum segment mass \(\Delta_w=\min_jS_{\tau_{j-1}+1:\tau_j}^w\) obeys \(w_{\max}\leq\Delta_w/24\).
\end{assumption}

\begin{assumption}[Dependence]
\label{assump:mixing}
Every centered bounded transform \(V_i=f_i(Y_i)\) used for scores or central-mass indicators obeys, on every interval \(A\),
\[
\Pp\!\left(\left|\sum_{i\in A}V_i\right|
>C_{\rm dep}\|V\|_\infty\{\sqrt{|A|x}+q_Nx\}\right)\leq2e^{-x},
\qquad x\geq1.
\]
For a two-sided extension, absolute regularity is measured by
\begin{equation}
\beta(k)=\sup_t\E\!\left[
\sup_{B\in\mathcal F_{t+k}^{\infty}}
\left|\Pp(B\mid\mathcal F_{-\infty}^{t})-\Pp(B)\right|
\right].
\label{eq:beta-mixing-coefficient}
\end{equation}
Independent and fixed-order dependent observations permit \(q_N=O(1)\); the conservative geometric-mixing specialization \(\beta(k)\lesssim e^{-ck}\) uses \(q_N=O\{\log^2(eN)\}\).
\end{assumption}

\begin{assumption}[Central mass]
\label{assump:central-mass}
There is a fixed \(p_0\in(0,1]\) such that every observation \(i\) in segment \(j\) and every \(|u|\leq r_0\) satisfy
\(\Pp\!\left(\left|\sqrt{w_i}(Y_i-\theta_j-u)\right|\leq c/2\right)\geq p_0\). For \(K\geq1\), we impose the local-jump restriction \(\kappa_{\max}\leq p_0r_0/16\); for \(K=0\), set \(\kappa_{\max}=0\).
\end{assumption}

Let \(\cA_N\) contain all nonempty index intervals and let \(L_N=1+N^2+N^3\) bound their number together with all split triples. Write \(x_N(\delta)=\log(64L_N/\delta)\) and define
\begin{equation}
\Lambda_N(\delta)=q_N^2\log\!\left(\frac{64L_N}{\delta}\right).
\label{eq:Lambda}
\end{equation}
Set \(m_{\mathrm{curv}}=C_{\mathrm{curv}}\Lambda_N(\delta)\).

\begin{assumption}[Spacing and signal]
\label{assump:signal}
For \(K\geq1\), set \(\kappa_{\min}=\min_j\kappa_j>0\) and \(\kappa_{\max}=\max_j\kappa_j\). For constants \(c_{m,1},C_g,C_{\mathrm{snr}}>0\) and \(0<c_{m,2}\leq 1/12\), the admissible side mass \(m_N\), guard mass \(g_N\), and signal obey
\[
\begin{gathered}
m_{\mathrm{curv}}\leq m_N\leq
c_{m,1}\frac{\Lambda_N(\delta)}{\kappa_{\max}^2},
\qquad
C_g\frac{\Lambda_N(\delta)}{\kappa_{\min}^2}+w_{\max}
\leq g_N\leq c_{m,2}\Delta_w,\\
m_N\leq \frac{\Delta_w}{12},
\qquad
\kappa_{\min}^2\Delta_w\geq C_{\mathrm{snr}}\Lambda_N(\delta).
\end{gathered}
\]
\end{assumption}

For adjacent pure intervals \(A,B\), let \(D^\star(A,B)\) be the excess population loss caused by fitting their union with one location instead of two.

\begin{lemma}[Mixed-segment excess risk]
\label{lem:mixed-excess}
If \(A\) and \(B\) have locations \(\theta_A,\theta_B\) in the common curvature region, then
\begin{align}
D^\star(A,B)
&\geq\frac{m_0}{2}
\frac{S_A^wS_B^w}{S_A^w+S_B^w}
(\theta_A-\theta_B)^2,
\label{eq:merge-lower}\\
D^\star(A,B)
&\leq\frac{M_0}{2}
\frac{S_A^wS_B^w}{S_A^w+S_B^w}
(\theta_A-\theta_B)^2.
\label{eq:merge-upper}
\end{align}
\end{lemma}

\begin{theorem}[Single-change population margin]
\label{thm:population-margin}
Suppose \([s,e]\) contains one change \(\tau\) of size \(\kappa\).
The population counterpart of \eqref{eq:profile-gain} is maximized at \(\tau\).
For \(b<\tau\),
\begin{equation}
\Gamma^\star_{s,e}(\tau)-\Gamma^\star_{s,e}(b)
=2D^\star([b+1,\tau],[\tau+1,e])
\geq m_0\kappa^2
\frac{S_{b+1:\tau}^wS_{\tau+1:e}^w}
{S_{b+1:\tau}^w+S_{\tau+1:e}^w},
\label{eq:left-margin}
\end{equation}
with the symmetric bound for \(b>\tau\).
When the misclassified mass is no larger than the opposite pure-side mass, the gap is at least \(m_0\kappa^2d_w(b,\tau)/2\).
\end{theorem}

The identity in \eqref{eq:left-margin} converts a difficult mixed-center calculation into an excess-risk comparison. Bounded Huber scores keep empirical same-location merge gains at \(O\{\Lambda_N(\delta)\}\), while different-location merges retain the curvature term in Lemma~\ref{lem:mixed-excess}.

\begin{lemma}[Uniform empirical merge costs]
\label{lem:empirical-merge}
For adjacent intervals \(A,B\) with \(S_A^w,S_B^w\geq m_{\mathrm{curv}}\), define
\begin{equation*}
\widehat D(A,B)=
\inf_\theta\{\widehat{\cL}_A(\theta)+\widehat{\cL}_B(\theta)\}
-\inf_\theta\widehat{\cL}_A(\theta)
-\inf_\theta\widehat{\cL}_B(\theta).
\end{equation*}
On an event of probability at least \(1-\delta/2\), uniformly over the master collection, intervals from the same true segment with masses at least \(m_{\mathrm{curv}}\) satisfy
\begin{equation}
0\leq\widehat D(A,B)\leq C_0\Lambda_N(\delta).
\label{eq:null-merge}
\end{equation}
If instead \(A,B\) lie in adjacent segments separated by \(\kappa\), then
\begin{align}
\widehat D(A,B)
&\geq c_0\kappa^2\frac{S_A^wS_B^w}{S_A^w+S_B^w}
-C_1\Lambda_N(\delta),
\label{eq:alternative-merge}\\
\widehat D(A,B)
&\leq C_2\kappa^2\frac{S_A^wS_B^w}{S_A^w+S_B^w}
+C_3\Lambda_N(\delta).
\label{eq:alternative-merge-upper}
\end{align}
\end{lemma}

Thus the same gain separates homogeneous intervals from intervals containing a sufficiently strong change: bounded scores control the null merge cost, and empirical curvature preserves the \(\kappa^2\)-weighted alternative up to a uniform fluctuation.

Let \(W_N=W(N)\) denote the total reliability mass.
Weighted-coordinate sampling supplies the algorithmic counterpart of the population margin.

\begin{lemma}[Weighted-coordinate isolation]
\label{lem:weighted-isolation}
Under Assumption~\ref{assump:weights}, suppose
\(\max_i w_i\leq\Delta_w/24\).
If the number of sampled intervals satisfies
\begin{equation}
M\geq C_I\left(\frac{W_N}{\Delta_w}\right)^2
\log\!\left\{\frac{4(K\vee1)}{\delta}\right\},
\label{eq:number-intervals}
\end{equation}
then, with probability at least \(1-\delta/4\), every change \(\tau_j\) is the only change in some sampled interval \([s,e]\) for which
\begin{equation}
\frac{\Delta_w}{6}\leq S_{s:\tau_j}^w,\,
S_{\tau_j+1:e}^w\leq\frac{\Delta_w}{3},
\qquad
S_{s:e}^w\leq\frac{2\Delta_w}{3}.
\label{eq:isolated-mass}
\end{equation}
\end{lemma}

\begin{theorem}[Finite-sample count recovery and localization]
\label{thm:exact-recovery}
Under Assumptions~\ref{assump:location}--\ref{assump:signal}, sample \(M\) weighted-coordinate intervals with \(M\) satisfying \eqref{eq:number-intervals}, and choose \(r_N=C_r\Lambda_N(\delta)\) with the constants specified in Appendix~\ref{app:exact-recovery}.
Then, with probability at least \(1-\delta\), core RWCP returns exactly \(K\) changes and, after ordering,
\begin{equation}
d_w(\widehat\tau_j,\tau_j)
\leq C_{\rm loc}\frac{\Lambda_N(\delta)}{\kappa_j^2},
\qquad j\in[K].
\label{eq:main-localization}
\end{equation}
For \(K=0\), Assumptions~\ref{assump:location}--\ref{assump:central-mass}, \(m_N\geq m_{\mathrm{curv}}\), and \(r_N>2C_0\Lambda_N\) suffice for no false boundary; the jump-dependent spacing and interval-count conditions are unnecessary.
\end{theorem}

The theorem describes locally separated changes; feasible side mass requires \(C_{\mathrm{curv}}\kappa_{\max}^2\leq c_{m,1}\), and \(g_N,M\) use signal/spacing bounds rather than fully adaptive tuning.
In the proof, uniform merge-cost bounds suppress homogeneous intervals, while weighted isolation supplies a short over-threshold interval for each change. The narrowest such interval contains one change; the profile margin localizes it, and the guard removes it without losing neighboring isolating intervals. Induction yields exact \(K\) and simultaneous localization.

\begin{corollary}[Estimated-weight transfer]
\label{cor:estimated-weights}
Suppose a weight estimator obeys
\(c_ww_i^\circ\leq\widehat w_i\leq C_ww_i^\circ\) uniformly with probability at least \(1-\delta_w\), and conditional score assumptions hold uniformly on this event.
Then RWCP computed with \(\widehat w_i\) satisfies, with probability at least \(1-\delta-\delta_w\),
\begin{equation*}
\widehat K=K,\qquad
d_{w^\circ}(\widehat\tau_j,\tau_j)
\leq c_w^{-1}d_{\widehat w}(\widehat\tau_j,\tau_j)
\leq C\frac{\Lambda_N(\delta)}{\kappa_j^2}.
\end{equation*}
The spacing, no-dominant-atom, and signal conditions change only by constants depending on \((c_w,C_w)\).
\end{corollary}

Independent and fixed-order dependent scores attain weighted localization order \(O\{\log(N/\delta)/\kappa_j^2\}\); geometric mixing incurs the explicit \(q_N^2\) factor. External proxies or sample splitting can justify Corollary~\ref{cor:estimated-weights}; target-score-derived weights require separate joint analysis.

Under independent Gaussian scores and inactive caps, the two-point lower bound matches the \(\log N/\kappa^2\) order (Appendix~\ref{app:minimax-proof}); it does not address dependence or active caps.

The weighted metric translates to text units: bounded weights make \(d_w(u,v)\) comparable to the number of misplaced sentences, while \(\sigma_i^2\asymp n_i^{-1}\) with inactive caps makes it comparable to misplaced tokens. Formal metric corollaries and the lower-bound argument appear in Appendix~\ref{app:minimax-proof}.

\section{Experiments}
\label{sec:experiments}

\subsection{Protocol and overall comparison}

We compare an analytic-threshold implementation of core RWCP with WCP, then evaluate whether the separate recurrent decoder improves window segmentation and how both methods handle dense changes.
Five cached-score families cover single changes in WikiQA, News, and Story; real CoAuthor sessions; multiple-change Story documents; text attacks; and varying LLM-written proportions. The suite has 25 settings and 2,690 instances, including all WikiQA records under the common singleton convention (Appendix~\ref{subsec:reproduction_setup}).
Comparators are TextTiling, direct and sentence-level prediction, local voting, VCP, and WCP where outputs are available \citep{hearst1997texttiling,zhang2024localization,li2026segmenting}.
Core RWCP, RWCP-R, and WCP share cached scores and reliability proxies on nontrivial inputs; single-change experiments use fine-tuned scores and the other families use normalized scores.
Core RWCP uses one fixed analytic-threshold configuration, specified in Appendix~\ref{subsec:core-evaluation}; its profile search is the subject of Theorem~\ref{thm:exact-recovery}, although these numerical settings are not certified to satisfy the theorem's unknown constants and assumptions.
RWCP-R uses a separately frozen configuration: Huber cutoff $c=1.5$, transition penalty $\lambda=4$, weight exponent $\eta=0.5$, scale floor $0.25$, and six deterministic starts. Appendix~\ref{subsec:selection-provenance} documents its selection on benchmarks that informed earlier development, making the comparisons retrospective.

We use standard raw-boundary WindowDiff (WD, lower is better) to measure window-level disagreement and raw count error $\CE_d=K_d-\widehat K_d$ to show its direction.
Positive CE indicates missed changes and negative CE indicates excess changes; count MAE, matched-boundary F1, and a no-boundary reference complete the recovery assessment in Appendices~\ref{subsec:core-evaluation} and~\ref{subsec:recovery-diagnostics}.
The inherited final-label window error $\WDsup$ is reported separately because it can exceed one and measures a different output.

\begin{table}[!htb]
\centering
\setlength{\abovecaptionskip}{0pt}
\setlength{\belowcaptionskip}{4pt}
\caption{Raw-boundary results averaged over documents, then equally over settings within each family (300, 290, 1,000, 600, and 500 instances). Core RWCP uses the fixed analytic threshold; the historical RWCP-analytic/block variants in Appendix~\ref{app:component-diagnostics} use additional practical refinements. Bold marks the lowest unrounded WD, not significance; CE is signed count bias. Dashes denote unavailable complete-family results.}
\label{tab:main-results}
\small
\setlength{\tabcolsep}{3pt}
\renewcommand{\arraystretch}{1.06}
\begin{adjustbox}{max width=\linewidth}
\begin{tabular}{@{}lrrrrrrrrrr@{}}
\toprule
\multirow{2}{*}{Method} & \multicolumn{2}{c}{Single change} & \multicolumn{2}{c}{CoAuthor} & \multicolumn{2}{c}{Multiple changes} & \multicolumn{2}{c}{Text attacks} & \multicolumn{2}{c}{LLM proportion} \\
\cmidrule(lr){2-3}\cmidrule(lr){4-5}\cmidrule(lr){6-7}\cmidrule(lr){8-9}\cmidrule(lr){10-11}
& $\WD\downarrow$ & CE & $\WD\downarrow$ & CE & $\WD\downarrow$ & CE & $\WD\downarrow$ & CE & $\WD\downarrow$ & CE \\
\midrule
TextTiling & 0.6449 & -2.487 & 0.5445 & 7.314 & 0.6251 & -4.585 & 0.6421 & -2.478 & 0.7611 & -3.730 \\
LLMPred & 0.4446 & -0.807 & 0.4996 & 8.303 & 0.4967 & -0.141 & 0.4508 & -1.013 & 0.4074 & -1.356 \\
SenPred & 0.6297 & -6.717 & 0.5485 & -4.155 & 0.9032 & -23.498 & 0.6791 & -7.390 & 0.9511 & -14.500 \\
Voting & 0.2725 & -0.903 & 0.5487 & 3.286 & 0.4797 & -4.626 & 0.3514 & -1.338 & 0.4939 & -3.164 \\
VCP & 0.3086 & -0.603 & 0.4913 & 9.369 & 0.4613 & -0.579 & -- & -- & -- & -- \\
WCP & 0.3961 & -1.403 & 0.4878 & 7.717 & 0.4363 & 0.772 & 0.3609 & -0.358 & 0.3538 & -0.804 \\
\midrule
Core RWCP & 0.2315 & 0.220 & 0.4897 & 10.152 & 0.4108 & 3.716 & 0.3369 & 0.940 & \textbf{0.2085} & 0.920 \\
RWCP-R & \textbf{0.1895} & 0.280 & \textbf{0.4869} & 10.166 & \textbf{0.3993} & 3.181 & \textbf{0.2676} & 0.475 & 0.2119 & 0.600 \\
\bottomrule
\end{tabular}
\end{adjustbox}
\end{table}

Core RWCP lowers family-macro WD from 0.4070 for WCP to 0.3355 (17.6\% relative) and improves WD in four of five families, but these window-level gains do not imply reliable boundary recovery. Its family-macro count MAE rises from 2.808 to 3.229 and matched-boundary F1 falls from 0.287 to 0.129. Single-change texts are the family-level exception: count MAE improves from 1.637 to 0.393 and F1 from 0.435 to 0.487, although F1 does not improve in every domain. On CoAuthor, multiple-change Story, text attacks, and LLM-proportion documents, core RWCP reports no boundary in 82.4\%, 93.1\%, 94.7\%, and 92.8\% of instances, respectively. Appendix Table~\ref{tab:core-evaluation} reports these diagnostics and conditional localization errors; no CoAuthor document has the correct positive change count under this fixed configuration. These core comparisons are descriptive; the paired intervals below concern RWCP-R.

RWCP-R has lower mean WD than WCP in all five families and lower WD than the fixed core implementation in four (Table~\ref{tab:main-results}).
With equal family weights, WD falls from 0.4070 for WCP to 0.3110, a 23.6\% relative reduction; legacy $\WDsup$ falls from 0.3249 to 0.2990 (8.0\%).
The frozen RWCP-R configuration improves standard WD in 23 of 25 settings against WCP and in 20 against every available non-RWCP baseline.
Nineteen pointwise paired 95\% intervals for the WD difference against WCP lie below zero, one lies above zero, and five include or touch zero.
The joint source-component bootstrap gives a family-macro difference of $-0.0959$ with interval $[-0.1074,-0.0844]$ (Appendix Table~\ref{tab:macro-uncertainty}); these exploratory intervals condition on the saved configurations and available source linkage.

Window-level agreement and recovery of individual changes answer different questions, so we interpret the two together.
RWCP-R improves matched-boundary F1 over WCP on single-change and text-attack families, from 0.435 to 0.544 and from 0.272 to 0.300, respectively.
On CoAuthor and dense-change Story, the fitted transition penalty yields fewer boundaries and lower F1 despite the WD gains; Appendix Table~\ref{tab:count-diagnostics} reports the corresponding count MAE, F1, and no-boundary rates.
This distinction motivates the density analysis below and identifies the regimes where the recurrent model's window-level advantage translates most directly into localized boundaries.

\subsection{Controlled localization and text perturbations}

\begin{table}[!ht]
\centering
\setlength{\abovecaptionskip}{0pt}
\setlength{\belowcaptionskip}{4pt}
\caption{Controlled Claude Haiku 4.5 results: 100 records per domain and condition under the common singleton convention. Bold marks the lowest unrounded WD; CE is signed count bias. Dashes indicate unavailable outputs. Legacy metrics appear in Tables~\ref{tab:single_cp_results} and~\ref{tab:decoherence_paraphrasing_results}.}
\label{tab:main-controlled}
\small
\setlength{\tabcolsep}{8pt}
\renewcommand{\arraystretch}{1.06}
\begin{adjustbox}{max width=\linewidth}
\begin{tabular}{lrrrrrr}
\toprule
\multirow{2}{*}{Method} & \multicolumn{2}{c}{WikiQA} & \multicolumn{2}{c}{News} & \multicolumn{2}{c}{Story} \\
\cmidrule(lr){2-3}\cmidrule(lr){4-5}\cmidrule(lr){6-7}
& $\WD\downarrow$ & CE & $\WD\downarrow$ & CE & $\WD\downarrow$ & CE \\
\midrule
\multicolumn{7}{l}{\textit{Single change}} \\
TextTiling & 0.427 & -0.09 & 0.796 & -4.18 & 0.712 & -3.19 \\
LLMPred & 0.428 & 0.41 & 0.420 & -0.81 & 0.486 & -2.02 \\
PaLD & 0.431 & -1.28 & -- & -- & -- & -- \\
SenPred & 0.257 & -0.90 & 0.707 & -6.74 & 0.924 & -12.51 \\
Voting & 0.286 & 0.00 & 0.107 & -0.32 & 0.424 & -2.39 \\
VCP & 0.267 & 0.20 & 0.265 & -0.99 & 0.394 & -1.02 \\
WCP & 0.284 & 0.07 & 0.370 & -1.73 & 0.534 & -2.55 \\
RWCP-R & \textbf{0.223} & 0.47 & \textbf{0.102} & 0.03 & \textbf{0.243} & 0.34 \\
\midrule
\multicolumn{7}{l}{\textit{Decoherence}} \\
TextTiling & 0.427 & -0.09 & 0.797 & -4.18 & 0.708 & -3.17 \\
LLMPred & 0.416 & 0.37 & 0.437 & -1.22 & 0.542 & -2.96 \\
SenPred & 0.362 & -1.30 & 0.885 & -10.58 & 0.953 & -12.89 \\
Voting & 0.359 & -0.03 & 0.363 & -1.75 & 0.639 & -3.82 \\
WCP & 0.340 & 0.44 & 0.407 & -0.59 & 0.398 & -0.45 \\
RWCP-R & \textbf{0.291} & 0.65 & \textbf{0.335} & 0.60 & \textbf{0.335} & 0.59 \\
\midrule
\multicolumn{7}{l}{\textit{Paraphrasing}} \\
TextTiling & 0.431 & -0.06 & 0.785 & -4.16 & 0.704 & -3.21 \\
LLMPred & 0.429 & 0.30 & 0.395 & -0.44 & 0.486 & -2.13 \\
SenPred & 0.273 & -0.96 & 0.696 & -6.53 & 0.905 & -12.08 \\
Voting & 0.267 & 0.04 & \textbf{0.091} & -0.24 & 0.389 & -2.23 \\
WCP & 0.266 & 0.25 & 0.326 & -1.05 & 0.428 & -0.75 \\
RWCP-R & \textbf{0.189} & 0.37 & 0.146 & 0.12 & \textbf{0.310} & 0.52 \\
\bottomrule
\end{tabular}
\end{adjustbox}
\end{table}

The single-change and attack settings test whether the same configuration transfers across three domains and two text perturbations (Table~\ref{tab:main-controlled}).
RWCP-R attains the lowest available standard WD in all three unperturbed domains, all three decoherence settings, and two of three paraphrasing settings; Voting is better on paraphrased News.
All nine paired WD intervals against WCP exclude zero in favor of RWCP-R.
Count bias adds a useful qualification: RWCP-R is closer to zero than WCP on unperturbed News and Story, while it is more conservative on WikiQA and the decoherence settings.
Together with the higher family-level boundary F1, these results identify single changes and attacked text as the clearest empirical successes of the recurrent decoder.
They compare complete frozen pipelines; the component analysis below examines individual design choices at that configuration.
The GPT-5-mini single-change aggregates remain a separate supplemental panel (Table~\ref{tab:single-gpt-supp}), outside the primary means and paired analysis.

\subsection{Change density and real collaborative text}

\begin{table}[!ht]
\centering
\setlength{\abovecaptionskip}{0pt}
\setlength{\belowcaptionskip}{4pt}
\caption{Story change-density results: 100 documents per generator and $K$. True $K$ is used only for evaluation. WD is standard raw-boundary WindowDiff; CE is signed count bias. Bold compares WCP and RWCP-R within each generator.}
\label{tab:main-density}
\small
\setlength{\tabcolsep}{4pt}
\renewcommand{\arraystretch}{1.06}
\begin{adjustbox}{max width=\linewidth}
\begin{tabular}{lrrrrr}
\toprule
\multirow{2}{*}{Generator} & \multirow{2}{*}{$K$} & \multicolumn{2}{c}{WCP} & \multicolumn{2}{c}{RWCP-R} \\
\cmidrule(lr){3-4}\cmidrule(lr){5-6}
& & WD & CE & WD & CE \\
\midrule
\multirow{5}{*}{Claude Haiku 4.5} & 1 & 0.400 & -0.52 & \textbf{0.288} & 0.58 \\
 & 2 & 0.404 & -0.19 & \textbf{0.374} & 1.44 \\
 & 3 & 0.425 & -0.11 & \textbf{0.378} & 2.10 \\
 & 5 & \textbf{0.425} & 0.39 & 0.429 & 4.26 \\
 & 8 & \textbf{0.430} & 2.85 & 0.448 & 7.21 \\
\midrule
\multirow{5}{*}{GPT-5-mini} & 1 & 0.434 & -0.38 & \textbf{0.339} & 0.79 \\
 & 2 & 0.430 & 0.01 & \textbf{0.408} & 1.51 \\
 & 3 & 0.470 & 0.31 & \textbf{0.418} & 2.42 \\
 & 5 & 0.475 & 1.26 & \textbf{0.447} & 4.28 \\
 & 8 & 0.471 & 4.10 & \textbf{0.464} & 7.22 \\
\bottomrule
\end{tabular}
\end{adjustbox}
\end{table}

The multiple-change settings probe the transition from isolated changes to short alternating passages (Table~\ref{tab:main-density}).
Relative to WCP, standard WD improves for Claude at $K\in\{1,2,3\}$ and for GPT-5-mini at every evaluated $K$.
The advantage narrows at high density: at $K=8$, the WD difference is $+0.018$ for Claude and $-0.006$ for GPT-5-mini.
CE exceeds seven in both cases, corresponding to fewer than one detected boundary on average despite eight reference changes.
The trend is consistent with the practical difficulty of separating short passages under a fixed transition penalty and shows where the window-level advantage becomes small.

On CoAuthor, WD changes only from 0.4878 to 0.4869, while mean CE increases from 7.717 to 10.166.
The no-boundary predictor is close on CoAuthor and the LLM-proportion family: its WD is 0.4893 and 0.2141, compared with 0.4869 and 0.2119 for RWCP-R.
Thus the strongest evidence for localized boundary improvement comes from the single-change and perturbation families, whereas these two families emphasize the distinction between WD and exact boundary recovery.
Appendix Tables~\ref{tab:wd_ce_results}, \ref{tab:percentage_results}, and~\ref{tab:count-diagnostics} give the full baseline, proportion, and recovery diagnostics.

\subsection{Component contributions}

\begin{table}[!ht]
\centering
\setlength{\abovecaptionskip}{0pt}
\setlength{\belowcaptionskip}{4pt}
\caption{Frozen component comparisons, averaged over settings and then families. Fixed scale uses $\sigma=1$, uniform weights exponent zero, and quadratic loss $c=10^6$. No variant is retuned. Bold marks the lowest window error; signed CE is not ranked.}
\label{tab:main-ablation}
\small
\setlength{\tabcolsep}{4pt}
\renewcommand{\arraystretch}{1.06}
\begin{adjustbox}{max width=\linewidth}
\begin{tabular}{lrrr}
\toprule
\multirow{2}{*}{Variant} & \multicolumn{2}{c}{Raw boundaries} & Final labels \\
\cmidrule(lr){2-3}\cmidrule(lr){4-4}
& WD$\downarrow$ & Mean CE & $\WDsup\downarrow$ \\
\midrule
RWCP-analytic & 0.3466 & 2.915 & 0.3233 \\
RWCP-block & 0.3371 & 3.050 & 0.3158 \\
RWCP-R & \textbf{0.3110} & 2.940 & \textbf{0.2990} \\
R: fixed scale & 0.3170 & 2.984 & 0.3038 \\
R: uniform weights & 0.3118 & 2.756 & 0.3058 \\
R: quadratic approximation & 0.3179 & 2.782 & 0.3130 \\
\bottomrule
\end{tabular}
\end{adjustbox}
\end{table}

Table~\ref{tab:main-ablation} probes three design choices at the selected recurrent configuration.
Fitting the scale and retaining the finite Huber cutoff each improve both window errors relative to their fixed-configuration alternatives.
Uniform weights nearly match standard WD: the paired difference for full RWCP-R minus uniform weights is $-0.0008$, with interval $[-0.0052,0.0038]$.
This contrast makes the empirical role of weighting metric-specific: the full model has lower legacy window error, while uniform weights have lower count bias and stronger boundary matching in Appendix Table~\ref{tab:ablation}.

RWCP-R also reduces family-macro standard WD by 10.3\% and 7.7\% relative to the historical analytic and block-calibrated variants. These are different from the fixed analytic-threshold core evaluation in Table~\ref{tab:main-results}.
Because those historical variants differ in search, scale, and other settings together, we interpret this comparison as a pipeline result; the fixed component variants provide the more focused evidence about scale, loss, and weighting.
Family-wise comparisons, auxiliary metrics, and the post-freeze stability check are reported in Appendix~\ref{app:component-diagnostics}.

\section{Discussion and Conclusion}
\label{sec:discussion}

RWCP turns robust change-point detection into a comparison of optimized segment risks. The merge-cost identity supplies a population margin without solving for nonlinear mixed-segment Huber centers, and the quadratic limit recovers weighted CUSUM. With bounded-score concentration, reliability-coordinate isolation, and guarded recursion, this yields exact count recovery and weighted localization for core RWCP under the stated conditions. The new fixed-configuration core evaluation directly measures the corresponding profile search on cached detector scores, but it does not establish that those scores or numerical thresholds meet the theorem's assumptions.

The core evaluation lowers average WindowDiff relative to WCP, with a simultaneous count and boundary-F1 gain only on the single-change family. Its high no-boundary rates on CoAuthor and dense or altered texts demonstrate the limits of a fixed analytic threshold in this retrospective score suite. The recurrent extension shares information across nonadjacent passages and lowers family-macro WindowDiff further, with the clearest boundary-recovery gain on single changes; its count/F1 tradeoff persists in other families. These results motivate count-aware calibration on independent detector streams.

The two contributions have complementary scopes: Theorem~\ref{thm:exact-recovery} covers analytic-threshold core RWCP under specified conditions, while RWCP-R evaluates a distinct recurrent objective on retrospective cached scores. The empirical results support window-error reductions, not general exact-count recovery on this benchmark suite; gradual editing, within-sentence mixing, reversed score semantics, long-memory dependence, and target-dependent weight estimation remain outside the present score model.

\section*{Ethics Statement} RWCP is a research tool for provenance segmentation, not automatic evidence of misconduct.
Detector errors can vary across language, genre, dialect, and demographic writing style; deployment therefore requires domain-specific validation, uncertainty reporting, human review, and respect for dataset and model-provider terms.

\section*{Reproducibility Statement}
The appendices document assumptions, proofs, frozen settings, and per-setting results. The accompanying local records preserve cached inputs and outputs, metric-recomputation scripts, and the complete-record correction; they do not constitute a fresh independent test or establish exact regeneration from model names.

\section*{AI Usage Statement}
Generative AI assisted with manuscript editing, \LaTeX{}, code, and consistency checks, including the reporting correction. Numerical comparisons are derived from saved predictions and explicit evaluation conventions. The authors remain responsible for validating the mathematics, implementation, citations, and empirical claims.

\bibliography{reference_iclr,iclr_additions}

@inproceedings{rajpurkar2016squad,
  title={{SQuAD}: 100,000+ Questions for Machine Comprehension of Text},
  author={Rajpurkar, Pranav and Zhang, Jian and Lopyrev, Konstantin and Liang, Percy},
  booktitle={Proceedings of the 2016 Conference on Empirical Methods in Natural Language Processing},
  pages={2383--2392},
  year={2016},
  doi={10.18653/v1/D16-1264}
}

@inproceedings{narayan2018dont,
  title={Don't Give Me the Details, Just the Summary! Topic-Aware Convolutional Neural Networks for Extreme Summarization},
  author={Narayan, Shashi and Cohen, Shay B. and Lapata, Mirella},
  booktitle={Proceedings of the 2018 Conference on Empirical Methods in Natural Language Processing},
  pages={1797--1807},
  year={2018},
  doi={10.18653/v1/D18-1206}
}

@inproceedings{fan2018hierarchical,
  title={Hierarchical Neural Story Generation},
  author={Fan, Angela and Lewis, Mike and Dauphin, Yann},
  booktitle={Proceedings of the 56th Annual Meeting of the Association for Computational Linguistics},
  pages={889--898},
  year={2018},
  doi={10.18653/v1/P18-1082}
}

@article{hearst1997texttiling,
  title={{TextTiling}: Segmenting Text into Multi-paragraph Subtopic Passages},
  author={Hearst, Marti A.},
  journal={Computational Linguistics},
  volume={23},
  number={1},
  pages={33--64},
  year={1997}
}

@article{pevzner2002critique,
  title={A Critique and Improvement of an Evaluation Metric for Text Segmentation},
  author={Pevzner, Lev and Hearst, Marti A.},
  journal={Computational Linguistics},
  volume={28},
  number={1},
  pages={19--36},
  year={2002}
}

@article{li2026segmenting,
  title={Segmenting Human--{LLM} Co-authored Text via Change Point Detection},
  author={Li, Mengchu and Zhu, Jin and Li, Jinglai and Shi, Chengchun},
  journal={arXiv preprint arXiv:2605.03723},
  year={2026}
}

@article{huber1964,
  title={Robust Estimation of a Location Parameter},
  author={Huber, Peter J.},
  journal={The Annals of Mathematical Statistics},
  volume={35},
  number={1},
  pages={73--101},
  year={1964}
}

@book{huber2009,
  title={Robust Statistics},
  author={Huber, Peter J. and Ronchetti, Elvezio M.},
  edition={2},
  publisher={Wiley},
  year={2009}
}

@article{baranowski2019not,
  title={Narrowest-over-threshold Detection of Multiple Change Points and Change-point-like Features},
  author={Baranowski, Rafal and Chen, Yining and Fryzlewicz, Piotr},
  journal={Journal of the Royal Statistical Society: Series B},
  volume={81},
  number={3},
  pages={649--672},
  year={2019}
}

@article{fearnhead2018changepoint,
  title={Changepoint Detection in the Presence of Outliers},
  author={Fearnhead, Paul and Rigaill, Guillem},
  journal={Journal of the American Statistical Association},
  volume={114},
  number={525},
  pages={169--183},
  year={2019},
  doi={10.1080/01621459.2017.1385466}
}

@article{fryzlewicz2014wbs,
  title={Wild Binary Segmentation for Multiple Change-point Detection},
  author={Fryzlewicz, Piotr},
  journal={The Annals of Statistics},
  volume={42},
  number={6},
  pages={2243--2281},
  year={2014}
}

@article{wang2020univariate,
  title={Univariate Mean Change Point Detection: Penalization, {CUSUM} and Optimality},
  author={Wang, Daren and Yu, Yi and Rinaldo, Alessandro},
  journal={Electronic Journal of Statistics},
  volume={14},
  number={1},
  pages={1917--1961},
  year={2020}
}

@article{yu2020review,
  title={A Review on Minimax Rates in Change Point Detection and Localisation},
  author={Yu, Yi},
  journal={arXiv preprint arXiv:2011.01857},
  year={2020}
}

@inproceedings{li2021adversarial,
  title={Adversarially Robust Change Point Detection},
  author={Li, Mengchu and Yu, Yi},
  booktitle={Advances in Neural Information Processing Systems},
  volume={34},
  pages={22955--22967},
  year={2021}
}

@article{cho2022twostage,
  title={Two-stage Data Segmentation Permitting Multiscale Change Points, Heavy Tails and Dependence},
  author={Cho, Haeran and Kirch, Claudia},
  journal={Annals of the Institute of Statistical Mathematics},
  volume={74},
  number={4},
  pages={653--684},
  year={2022}
}

@book{doukhan1994mixing,
  title={Mixing: Properties and Examples},
  author={Doukhan, Paul},
  publisher={Springer},
  year={1994}
}

@incollection{merlevede2009bernstein,
  title={Bernstein Inequality and Moderate Deviations under Strong Mixing Conditions},
  author={Merlev{\`e}de, Florence and Peligrad, Magda and Rio, Emmanuel},
  booktitle={High Dimensional Probability V: The Luminy Volume},
  pages={273--292},
  publisher={Institute of Mathematical Statistics},
  year={2009}
}

@article{kunsch1989bootstrap,
  title={The Jackknife and the Bootstrap for General Stationary Observations},
  author={K{\"u}nsch, Hans R.},
  journal={The Annals of Statistics},
  volume={17},
  number={3},
  pages={1217--1241},
  year={1989}
}

@inproceedings{gehrmann2019gltr,
  title={{GLTR}: Statistical Detection and Visualization of Generated Text},
  author={Gehrmann, Sebastian and Strobelt, Hendrik and Rush, Alexander M.},
  booktitle={Proceedings of the 57th Annual Meeting of the Association for Computational Linguistics: System Demonstrations},
  pages={111--116},
  year={2019}
}

@inproceedings{mitchell2023detectgpt,
  title={{DetectGPT}: Zero-shot Machine-generated Text Detection Using Probability Curvature},
  author={Mitchell, Eric and Lee, Yoonho and Khazatsky, Alexander and Manning, Christopher D. and Finn, Chelsea},
  booktitle={International Conference on Machine Learning},
  pages={24950--24962},
  year={2023}
}

@inproceedings{bao2024fast,
  title={{Fast-DetectGPT}: Efficient Zero-shot Detection of Machine-generated Text via Conditional Probability Curvature},
  author={Bao, Guangsheng and Zhao, Yanbin and Teng, Zhiyang and Yang, Linyi and Zhang, Yue},
  booktitle={International Conference on Learning Representations},
  year={2024}
}

@inproceedings{hans2024binoculars,
  title={Spotting {LLMs} with Binoculars: Zero-shot Detection of Machine-generated Text},
  author={Hans, Abhimanyu and Schwarzschild, Avi and Cherepanova, Valeriia and Kazemi, Hamid and Saha, Aniruddha and Goldblum, Micah and Geiping, Jonas and Goldstein, Tom},
  booktitle={International Conference on Machine Learning},
  year={2024}
}

@inproceedings{tulchinskii2023intrinsic,
  title={Intrinsic Dimension Estimation for Robust Detection of {AI}-generated Texts},
  author={Tulchinskii, Eduard and Kuznetsov, Kristian and Kushnareva, Laida and Cherniavskii, Daniil and Nikolenko, Sergey and Burnaev, Evgeny and Barannikov, Serguei and Piontkovskaya, Irina},
  booktitle={Advances in Neural Information Processing Systems},
  volume={36},
  pages={39257--39276},
  year={2023}
}

@inproceedings{lee2022coauthor,
  title={{CoAuthor}: Designing a Human--{AI} Collaborative Writing Dataset for Exploring Language Model Capabilities},
  author={Lee, Mina and Liang, Percy and Yang, Qian},
  booktitle={Proceedings of the 2022 CHI Conference on Human Factors in Computing Systems},
  pages={1--19},
  year={2022}
}

@inproceedings{zeng2024hybrid,
  title={Detecting {AI}-generated Sentences in Human--{AI} Collaborative Hybrid Texts: Challenges, Strategies, and Insights},
  author={Zeng, Zijie and Liu, Shuyu and Sha, Lei and Li, Zhuang and Yang, Kaixuan and Liu, Sheng and Ga{\v{s}}evi{\'c}, Dragan and Chen, Guanliang},
  booktitle={International Joint Conference on Artificial Intelligence},
  year={2024}
}

@inproceedings{zhang2024localization,
  title={Machine-generated Text Localization},
  author={Zhang, Zhenyu and Qin, Wen and Plummer, Bryan A.},
  booktitle={Findings of the Association for Computational Linguistics: ACL 2024},
  pages={8357--8371},
  year={2024}
}

@inproceedings{lei2025pald,
  title={{PaLD}: Detection of Text Partially Written by Large Language Models},
  author={Lei, Enzo and Hsu, Hsiang and Chen, Ching-Feng},
  booktitle={International Conference on Learning Representations},
  year={2025}
}
\bibliographystyle{iclr2027_conference}

\appendix

\section{Related Work}
\label{sec:related}

This section positions RWCP at the intersection of machine-text detection, structured authorship localization, robust change point analysis, and dependence-aware calibration. The comparison clarifies which ingredients are inherited from detector scoring and weighted segmentation, and which are introduced by the profile-loss formulation.

\paragraph{Detection and mixed-authorship localization.}

Document-level machine-text detection has been studied through supervised classifiers, likelihood and rank statistics, perturbation tests, rewriting distances, intrinsic-dimension summaries, and watermark-based mechanisms. Representative zero-shot procedures include GLTR-style token-rank diagnostics, DetectGPT and Fast-DetectGPT probability-curvature statistics, Binoculars, and intrinsic-dimension or divergent $n$-gram tests \citep{gehrmann2019gltr,mitchell2023detectgpt,bao2024fast,hans2024binoculars,tulchinskii2023intrinsic}. Supervised approaches train classifiers on paired human and machine corpora or adapt detection rules to a target generator. These methods differ in their assumptions and transfer behavior, but most produce a scalar score or probability that can be used as an input to the present framework. RWCP is detector-agnostic: it changes how ordered scores are segmented, not how a single text unit is scored.

Mixed human--AI writing has motivated sentence-level prediction, token-to-sentence aggregation, window voting, supervised sequence segmentation, and partial-LLM detectors. The CoAuthor data record detailed human interactions with model suggestions and provide a particularly important real-world test bed \citep{lee2022coauthor}. Existing localization procedures often classify each unit independently, aggregate token predictions, or smooth local predictions over a fixed window \citep{zeng2024hybrid,zhang2024localization,lei2025pald}. Change point formulations impose a stronger structural prior: consecutive units share a source until a boundary occurs. This structure reduces isolated label flips and aligns the optimization objective with boundary localization.

The foundational WCP formulation makes a further advance by incorporating heterogeneous score variability. Its upper and lower bounds show that inverse-variance information, rather than raw sentence count, determines localization difficulty. Our work preserves this geometry but replaces the mean-based split statistic by a robust profile-loss gain and extends the stochastic analysis to locally dependent bounded-influence scores.

\paragraph{Change points and robust profile methods.}

CUSUM statistics are central to univariate mean-change detection \citep{wang2020univariate,yu2020review}. Binary segmentation, wild binary segmentation, and narrowest-over-threshold (NOT) algorithms use multiscale or random intervals to isolate individual changes \citep{fryzlewicz2014wbs,baranowski2019not}. NOT selects the shortest interval whose maximal contrast exceeds a threshold and then recurses, thereby reducing cancellation among multiple changes. Non-asymptotic localization theory typically combines three ingredients: a deterministic contrast margin, a uniform noise bound, and an interval-isolation event.

Our proof follows that architecture in reliability coordinates. The deterministic object is a profile gain rather than a linear contrast; the noise control is based on bounded Huber scores under an explicit dependence-aware Bernstein condition; and random intervals are sampled in cumulative reliability mass. The recursive proof also has to control boundary leakage because an estimated split need not coincide exactly with the true change.

Huber loss interpolates between quadratic and absolute loss and yields bounded score influence \citep{huber1964,huber2009}. Robust mean and location estimators have been developed under finite moments, contamination, and dependence. Median-of-means and Catoni-type estimators offer alternative guarantees, while Huber estimators are especially convenient here because their optimized objective defines a profile likelihood-ratio analogue. Robust change point methods include median and rank contrasts, robustified CUSUMs, M-estimation losses, and procedures designed for heavy-tailed or contaminated sequences \citep{li2021adversarial,cho2022twostage}.

Profile costs of the form $\min_\theta\sum_i\rho(Y_i-\theta)$, including Huber losses, are established in robust penalized segmentation \citep{fearnhead2018changepoint}. Bounded influence alone does not provide immunity to arbitrary outliers; in particular, an unbounded Huber loss and a bounded robust loss have different contamination behavior. Our guarantees require finite risk, central mass, bounded weights, local jumps, and the stated concentration condition. The specific distinction analyzed here is between a robust center-difference contrast and a reliability-weighted profile-loss gain. The former requires understanding a nonlinear mixed-distribution center at every candidate split. The latter compares optimized risks and converts an incorrect split into a merge-cost problem. This distinction is not cosmetic; it is the main reason a transparent deterministic margin can be proved.

\paragraph{Dependence and calibration.}

Text scores may be dependent because adjacent sentences share topic, vocabulary, discourse structure, and detector context. Change point theory under mixing and other weak-dependence conditions replaces independent concentration by blocking, coupling, martingale, or spectral arguments \citep{doukhan1994mixing,merlevede2009bernstein}. Block bootstrap and dependent multiplier methods are widely used to approximate distributions of statistics under serial dependence \citep{kunsch1989bootstrap}. In our main theory, dependence enters through an explicit Bernstein condition. Independent and fixed-order dependent sequences have the usual logarithmic complexity, while the geometric beta-mixing specialization retains the published logarithmic penalty. The bootstrap is kept secondary: it is a practical calibration mechanism, while exact recovery is established using an analytic threshold.

\section{Experimental Protocol and Additional Results}
\label{app:remaining-experiments}

This appendix documents the cached-score inventory, fixed RWCP-R configuration,
selection history, and evaluation conventions before presenting the full
benchmark comparisons. It then reports recovery and component diagnostics,
post-freeze stability, implementation details, and paired uncertainty estimates.
All 25 primary settings and 2,690 instances use the same reporting conventions
as Section~\ref{sec:experiments}.

\subsection{Data, comparators, and evaluation scope}
\label{subsec:reproduction_setup}

The controlled domains are denoted WikiQA, News, and Story following the
reference segmentation pipeline \citep{li2026segmenting}; their source
corpora are SQuAD, XSum, and WritingPrompts,
respectively \citep{rajpurkar2016squad,narayan2018dont,fan2018hierarchical}.
In the primary suite, single-change and attack data use cached Claude Haiku 4.5 texts
(\texttt{claude-haiku-4-5}); multiple-change Story also uses
\texttt{gpt-5-mini}. CoAuthor retains its existing
868/289/290 training/validation/test session partition and three source
classes: human, collaborative, and AI \citep{lee2022coauthor}.
The evaluated split has 290 sessions and 7,042 sentences.

\begin{table}[!htbp]
\centering
\caption{Completed benchmark inventory. Counts are evaluated document
instances; transformed copies can share a source. FT is the repository's
fine-tuned score and NFT its normalized score. The retained benchmarks use
the unknown-count protocol for segmentation.}
\label{tab:protocol-summary}
\small
\begin{tabular}{lrrll}
\toprule
Experiment & Settings & Instances & Score & Count protocol \\
\midrule
Single change & 3 & 300 & FT & Unknown $K$ \\

CoAuthor & 1 & 290 & NFT & Unknown $K$ \\
Multiple changes & 10 & 1,000 & NFT & Unknown $K$ \\
Text attacks & 6 & 600 & NFT & Unknown $K$ \\
LLM proportion & 5 & 500 & NFT & Unknown $K$ \\
\bottomrule
\end{tabular}
\end{table}

Every controlled setting contains 100 records, including WikiQA, News, and Story. In the three Claude WikiQA settings (single change and two attacks), records 80 and 86 each contain one sentence and no reference boundary. We retain these records: the only admissible boundary set is empty, both window errors and count MAE are zero, and empty--empty boundary F1 is one. This is a boundary-only convention, not a claim of correct source classification. It is applied equally to every method. The other 98 records per setting retain their saved predictions. Thus the primary suite contains 2,690 instances; six trivial instances enter reporting only and do not alter historical parameter selection. The multiple-change settings use
$K\in\{1,2,3,5,8\}$ for each generator; the proportion settings use target
fractions of 5\%, 10\%, 20\%, 40\%, and 80\%. We reuse the saved hybrid
texts, predictions, and detector scores. Generation revisions, dates, and
decoding metadata are incomplete; reproducibility is from these caches,
rather than exact regeneration from a model name.

Comparators are TextTiling \citep{hearst1997texttiling}, direct LLM prediction
(LLMPred), independent sentence prediction (SenPred), local majority voting
(Voting), VCP, and WCP. VCP/WCP use the reference change-point implementation
\citep{li2026segmenting}; sentence prediction and voting follow the
localization pipeline \citep{zhang2024localization}. PaLD
\citep{lei2025pald} has a complete local output only on WikiQA in the
single-change benchmark. SegFormer outputs on CoAuthor and VCP outputs for
the attack/proportion experiments are unavailable. We retain saved baseline
predictions and recompute metrics consistently; RWCP-R is the rerun method.
The new core RWCP comparison is also recomputed from the same caches, but uses
its own fixed analytic threshold rather than the historical practical
RWCP-analytic/block configurations.

For every nontrivial input in the primary suite, core RWCP and RWCP-R consume exactly the WCP cached score sequence and reliability proxies. FT and NFT are distinct score pipelines, and other
baselines need not share their representations. Historical WCP re-scores
concatenated predicted segments with the language model, whereas RWCP-R
labels its states from fitted score centers. Thus final-label metrics
compare complete pipelines. Standard WD on raw boundaries and the
component ablations provide complementary evidence about segmentation.

\subsection{Fixed analytic-threshold core RWCP evaluation}
\label{subsec:core-evaluation}

To evaluate the method analyzed in Theorem~\ref{thm:exact-recovery}, we run the
weighted Huber profile gain, reliability-coordinate interval search, analytic
threshold, and guarded recursion without the scale estimation, additional
multiscale intervals, global penalized fitting, or state sharing of the practical
variants. The fixed numerical configuration is $c=1.345$, capped weights in
$[0.2,5]$, minimum side and guard masses both 1, 500 sampled intervals,
$\delta=0.05$, threshold multiplier $C_r=1$, dependence multiplier 1, and
seed 42. Cached WCP reliability proxies are clipped directly without
per-document renormalization. The implementation additionally includes the full
document as a deterministic candidate interval and minimizes each Huber loss
over the observed score span; these details are outside the theorem's literal
fixed-$\Theta$, sampled-interval specification. The run does not verify the
unknown signal, spacing, curvature, or dependence constants required by the
theorem, and should not be read as a test of its probability guarantee.

\begin{table}[!htbp]
\centering
\caption{Fixed analytic-threshold core RWCP versus saved WCP boundaries on the same cached scores. Family means average documents within settings and then settings equally. F1 matches boundaries within one sentence. Zero CP is the fraction with no detected boundary. Loc. coverage is the fraction of documents with at least one true change whose estimated change count is exactly correct. Weighted loc. MAE averages ordered boundary distances $d_w$ only over those eligible documents; -- means none is eligible. The conditional localization mean must be interpreted together with coverage. These are descriptive results without newly estimated paired intervals.}
\label{tab:core-evaluation}
\small
\setlength{\tabcolsep}{3pt}
\begin{adjustbox}{max width=\linewidth}
\begin{tabular}{@{}lrrrrrrrrr@{}}
\toprule
& \multicolumn{3}{c}{WCP} & \multicolumn{6}{c}{Core RWCP} \\
\cmidrule(lr){2-4}\cmidrule(lr){5-10}
Family & WD & MAE & F1 & WD & MAE & F1 & Zero CP & Loc. cov. & Weighted loc. MAE \\
\midrule
Single change & 0.3961 & 1.637 & 0.435 & 0.2315 & 0.393 & 0.487 & 31.3\% & 61.4\% & 1.636 \\
CoAuthor & 0.4878 & 7.917 & 0.302 & 0.4897 & 10.159 & 0.057 & 82.4\% & 0.0\% & -- \\
Multiple changes & 0.4363 & 2.130 & 0.205 & 0.4108 & 3.716 & 0.026 & 93.1\% & 1.0\% & 0.892 \\
Text attacks & 0.3609 & 0.998 & 0.272 & 0.3369 & 0.940 & 0.038 & 94.7\% & 5.4\% & 2.788 \\
LLM proportion & 0.3538 & 1.360 & 0.220 & 0.2085 & 0.936 & 0.037 & 92.8\% & 6.4\% & 4.395 \\
\bottomrule
\end{tabular}
\end{adjustbox}
\end{table}

Across the 25 settings, the core implementation has lower family-macro WD
than WCP, but its count MAE and F1 are worse overall. On single-change texts,
the 61.4\% localization coverage is higher than WCP's 30.9\%, yet the
conditional weighted localization error is 1.636 for core RWCP versus 0.821
for WCP; these means use different eligible subsets and cannot establish a
per-boundary improvement. Only 1.0\% of multiple-change documents have the
correct positive change count under core RWCP. The apparent conditional
weighted error of 0.892 in that family rests on just 10 eligible documents.
The zero-boundary rates and low coverage limit any empirical claim of exact
recovery on these cached detector scores.

\begin{longtable}{@{}llrrrrr@{}}
\caption{Per-setting standard WD for WCP and fixed analytic-threshold core RWCP, with core count MAE, boundary F1 at one-sentence tolerance, and no-boundary fraction. Each setting contains 100 documents except CoAuthor (290). The six one-sentence WikiQA records follow the common empty-boundary convention.}
\label{tab:core-per-setting}\\
\toprule
Family & Setting & WCP WD & Core WD & Core MAE & Core F1 & Zero CP \\
\midrule
\endfirsthead
\toprule
Family & Setting & WCP WD & Core WD & Core MAE & Core F1 & Zero CP \\
\midrule
\endhead
Single & WikiQA & 0.284 & 0.279 & 0.580 & 0.380 & 60\% \\
Single & News & 0.370 & 0.119 & 0.130 & 0.760 & 8\% \\
Single & Story & 0.534 & 0.296 & 0.470 & 0.320 & 26\% \\
CoAuthor & Sessions & 0.488 & 0.490 & 10.159 & 0.057 & 82\% \\
Multiple & Claude $K=1$ & 0.400 & 0.319 & 0.920 & 0.070 & 92\% \\
Multiple & Claude $K=2$ & 0.404 & 0.393 & 1.900 & 0.047 & 91\% \\
Multiple & Claude $K=3$ & 0.425 & 0.419 & 2.840 & 0.051 & 88\% \\
Multiple & Claude $K=5$ & 0.425 & 0.447 & 4.830 & 0.035 & 87\% \\
Multiple & Claude $K=8$ & 0.430 & 0.460 & 7.820 & 0.022 & 86\% \\
Multiple & GPT-5-mini $K=1$ & 0.434 & 0.333 & 1.000 & 0.000 & 100\% \\
Multiple & GPT-5-mini $K=2$ & 0.430 & 0.395 & 1.970 & 0.020 & 97\% \\
Multiple & GPT-5-mini $K=3$ & 0.470 & 0.424 & 2.970 & 0.009 & 98\% \\
Multiple & GPT-5-mini $K=5$ & 0.475 & 0.453 & 4.960 & 0.006 & 97\% \\
Multiple & GPT-5-mini $K=8$ & 0.471 & 0.464 & 7.950 & 0.002 & 95\% \\
Attack & WikiQA decoh. & 0.340 & 0.365 & 0.980 & 0.020 & 100\% \\
Attack & News decoh. & 0.407 & 0.336 & 1.000 & 0.000 & 100\% \\
Attack & Story decoh. & 0.398 & 0.321 & 0.860 & 0.060 & 86\% \\
Attack & WikiQA para. & 0.266 & 0.361 & 0.970 & 0.030 & 99\% \\
Attack & News para. & 0.326 & 0.314 & 0.930 & 0.070 & 93\% \\
Attack & Story para. & 0.428 & 0.325 & 0.900 & 0.050 & 90\% \\
Proportion & 5\% & 0.321 & 0.076 & 0.980 & 0.007 & 97\% \\
Proportion & 10\% & 0.338 & 0.137 & 0.970 & 0.017 & 96\% \\
Proportion & 20\% & 0.375 & 0.259 & 0.920 & 0.040 & 91\% \\
Proportion & 40\% & 0.404 & 0.301 & 0.870 & 0.090 & 86\% \\
Proportion & 80\% & 0.330 & 0.269 & 0.940 & 0.030 & 94\% \\
\bottomrule
\end{longtable}

\subsection{Practical RWCP configuration}
\label{subsec:practical-configuration}

RWCP-R implements Section~\ref{subsec:recurrent-profiles}, with three declared source states on CoAuthor and two elsewhere. Given positive
cached reliability proxies $r_i$, we compute
\begin{equation*}
\widetilde w_i=\operatorname{clip}_{[0.05,10]}
\left(\frac{r_i^{\eta}}{N^{-1}\sum_j r_j^{\eta}}\right),\qquad \eta=0.5,
\end{equation*}
and do not renormalize after clipping. The cache proxy is $n_i^2/v_i$:
$n_i$ is the scoring tokenizer's unit token count and $v_i$ the NFT
score-variance proxy; FT uses $v_i=1$. Compression tempers extreme
reliability ratios without discarding their ordering. All settings use the same exponent.

Scores are centered by their median and divided by an initial robust scale.
For adjacent differences, let
\begin{equation*}
D_i=\frac{Y_{i+1}-Y_i}{\sqrt{1/\widetilde w_i+1/\widetilde w_{i+1}}},\qquad
s_D=\frac{\operatorname{median}_i|D_i-\operatorname{median}(D)|}{0.67448975}.
\end{equation*}
The initial scale is the maximum of $s_D$, 0.05 times the Gaussian-scaled
MAD of score levels, and $10^{-6}$ times the score range; a constant
sequence uses scale one. The fitted dimensionless scale in
\eqref{eq:recurrent-objective} is constrained by $\sigma\geq0.25$.

The frozen main configuration uses Huber cutoff $c=1.5$ and transition
penalty $\lambda=4$. Initial centers are equally spaced quantiles between
$(q,1-q)$, for $q\in\{0.05,0.20,0.35\}$, combined with initial scales
one and 0.25, giving six deterministic starts. Each start runs at most 50
alternating updates, stopping when the objective change is below
$10^{-8}(1+|\mathcal J|)$. Center and scale bisection use tolerance
$10^{-8}$. We retain the smallest attained objective, breaking exact ties
by fewer transitions. Empty states retain their previous centers. Fitted centers are ranked in ascending order to assign source-state labels; this assumes score ordering agrees with source semantics. The smallest allowed state run is one observation;
no reference boundary, target LLM proportion, generator identity, or
reference change count enters an unknown-count prediction.

\begin{algorithm}[t]
\caption{Recurrent profile decoding (RWCP-R)}
\label{alg:recurrent}
\begin{algorithmic}[1]
\Require Scores, reliability proxies, declared state count \(S\), and frozen \(c,\eta,\lambda,\sigma_{\min}\).
\State If \(N=1\), return no boundary (the boundary-only evaluation convention).
\State Standardize scores and construct capped weights as specified above.
\For{each of the six quantile/scale initializations}
  \For{\(t=1,\ldots,50\)}
    \State Decode \(z\) by Viterbi with Huber emissions and penalty \(\lambda\).
    \State Fit each occupied center by weighted Huber bisection; retain empty centers.
    \State Let \(F(\sigma)=\sum_i\psi_c(r_i(\sigma))r_i(\sigma)\).
    \State Set \(\sigma=\sigma_{\min}\) if \(F(\sigma_{\min})\leq N\); otherwise solve \(F(\sigma)=N\).
    \State Stop if the objective change is below \(10^{-8}(1+|\mathcal J|)\).
  \EndFor
\EndFor
\State Keep the smallest attained objective; break exact ties by fewer transitions.
\State Rank centers to label states and return \(\{i:z_i\ne z_{i+1}\}\).
\end{algorithmic}
\end{algorithm}
The implementation uses at most 64 bisection iterations per conditional fit with absolute tolerance \(10^{-8}\). For \(B\) bisection steps and \(T\) alternating iterations, six starts cost \(O(6TN(S^2+SB))\) time and \(O(NS)\) decoding memory. The scale ablation fixes \(\sigma=1\) and consequently uses only the three center initializations. The dynamic program is exact conditional on centers and scale; the overall fit is a deterministic multistart approximation.

For continuity, RWCP-analytic and RWCP-block denote the preceding
\texttt{analytic\_block\_v3} implementation, using $c=3$, clipped
mean-one weights in $[0.02,10]$, three refinement passes, and common
threshold multiplier 0.85. These practical core variants also use implementation refinements beyond the unmodified analytic-threshold theorem. Their existing results are retained in
Tables~\ref{tab:standard-wd-summary} and~\ref{tab:ablation}.
The five retained comparisons use RWCP-R throughout. The historical core configurations fix 500 sampled intervals (with exhaustive intervals for $N\leq24$), minimum side mass and guard mass one, nominal level 0.05, dependence factor one, seed 42, and optimization tolerance $10^{-6}$. Block calibration uses length four and 499 multiplier draws; its threshold mixes the bootstrap value with the analytic reference at weight 0.35 on the latter and is capped at 1.5 times that reference. Both variants use curvature floor 0.05, curvature shrinkage 0.5, and the common threshold multiplier 0.85. These empirical values are not asserted to satisfy the theorem's unknown signal/spacing requirements.

\subsection{Development path and leakage controls}
\label{subsec:selection-provenance}

Configuration selection minimized the family-balanced development mean of
\[
\frac12\mathrm{WD}_d
+\frac12\min\{\WDsup(d),1\}
+\frac14\frac{|K_d-\widehat K_d|}{K_d+1}
+\frac14\{1-\mathrm{F1}_{d,h=1}\}.
\]
The search was shared across all reported settings rather than repeated separately for each dataset.
Table~\ref{tab:development-path} records the adaptive sequence.
The final recurrent sweep was the Cartesian product \(c\in\{1.5,3\}\), \(\eta\in\{0.25,0.5,1\}\), \(\lambda\in\{0.25,0.5,1,2,4\}\), and \(\sigma_{\min}\in\{0.25,0.5\}\), giving 60 candidates.
The selected point is the frozen configuration used in the primary results and the reference point for the component ablations.

\begin{table}[!htbp]
\centering
\caption{Adaptive development path under the original broader protocol. The objective is the criterion above averaged first within families and then across families; lower is better. These are development-selection values, not test-set performance estimates.}
\label{tab:development-path}
\small
\setlength{\tabcolsep}{6pt}
\renewcommand{\arraystretch}{1.06}
\begin{adjustbox}{max width=\linewidth}
\begin{tabular}{lclc}
\toprule
Stage & Candidates & Main change & Best dev.\ objective \\
\midrule
Existing RWCP-block & fixed & NOT with block calibration & 0.62712 \\
Global profile study & 45 & global Huber profile and state pruning & 0.61382 \\
Recurrent RWCP-R & 60 & recurrent centers and fitted scale & \textbf{0.58506} \\
\bottomrule
\end{tabular}
\end{adjustbox}
\end{table}

The historical nontrivial subset contains 643 linked source components, with 716 instances assigned to development and 1,968 to the internal check by a fixed hash. Six singleton instances restored for complete 100-record reporting do not enter either split or parameter selection.
RWCP-R parameters and all component ablations were frozen before that check.
No reference boundary, reference change count, target LLM proportion, or generator identity enters an unknown-count prediction, and paired intervals resample source components rather than individual transformed records.
The search was adaptive rather than preregistered, and the internal check still uses benchmark sources that influenced earlier development; it therefore supports a post-freeze stability check, not an independent confirmatory claim.

\subsection{Metrics, selection, and provenance}

Standard boundary WindowDiff \citep{pevzner2002critique}, defined in
\eqref{eq:windowdiff}, compares raw detected boundary counts in windows of
width
\begin{equation}
k_d=\max\!\left\{1,\operatorname{round}\left(\frac{N_d}{2(K_d+1)}\right)\right\},
\label{eq:windowdiff-width}
\end{equation}
with ties-to-even rounding. It lies in $[0,1]$. For comparison with the
inherited tables, $\WDsup$ instead averages the absolute discrepancy
between reference and final-label boundary counts, using width
$\max\{1,\lfloor N_d/[2(K_d+1)]\rfloor\}$. It can exceed one and is not
standard WD. Old aggregate fields are not mixed with new RWCP values.

Raw count error is $\CE_d=K_d-\widehat K_d$, positive for missed changes,
and count MAE averages $|K_d-\widehat K_d|$ over documents.
CoAuthor also retains its historical final-label error $\CE_{\rm label}$.
Boundary F1 uses one-to-one matching of estimated and reference boundaries
within tolerance $h=1$ observation; $h=0$ is an exact-match diagnostic.
Two empty sets receive F1 one and exactly one empty set receives zero.
For singleton documents, which have no evaluable window, both window errors are defined as zero. The zero-detection fraction includes the retained singletons; outside these known no-change records, a high value diagnoses missed boundaries.
Table boldface is computed before rounding and does not denote significance.

The reported suite focuses on sentence-level segmentation with unknown
change counts. This scope was narrowed after inspecting the broader
development results, so its comparisons are retrospective. The selected
configuration and predictions were retained without retuning for this subset.

The development objective, adaptive search path, source-group split,
freezing decisions, and leakage controls are summarized in
Appendix~\ref{subsec:selection-provenance}; this appendix retains the
complete metric definitions and benchmark-specific results.
Source IDs are incomplete, so component linkage reduces obvious reuse but
does not establish independence.

Pointwise paired 95\% intervals against WCP use 2,000 bootstrap repetitions,
resampling source components within each setting. We report them in
Appendix~\ref{app:paired-results}, without multiplicity correction.
Family summaries average settings equally; the component summary then
averages the five retained families equally.

\subsection{Five benchmark comparisons}
\label{sec:experimental_reproduction}

\paragraph{Single-change localization.}
In the primary Claude evaluation, RWCP-R has the lowest $\WDsup$ on all three domains
(Table~\ref{tab:single_cp_results}): 0.184 on WikiQA, 0.097 on News,
and 0.235 on Story. Their strongest available baseline values are 0.200,
0.123, and 0.306. Standard WD averaged over the domains falls from
0.2441 for RWCP-block to 0.1895 for RWCP-R, and count MAE falls from
0.490 to 0.340. Paired standard-WD intervals against WCP exclude zero
on all three domains, showing a consistent pipeline-level gain in the controlled Claude single-change setting. This comparison alone does not isolate Huber loss from state sharing, scale fitting, or penalty selection.

\begin{table}[!htbp]
\centering
\caption{Primary Claude Haiku 4.5 single-change results, 100 records each in WikiQA, News, and Story. Values are document means $\pm$ sample standard deviations, recomputed with the singleton boundary convention. Bold marks the lowest unrounded $\WDsup$ or smallest absolute mean CE, not significance. PaLD has outputs only for WikiQA.}
\label{tab:single_cp_results}
\label{tab:single-change}
\small
\setlength{\tabcolsep}{4pt}
\renewcommand{\arraystretch}{1.06}
\begin{adjustbox}{max width=\linewidth}
\begin{tabular}{lrrrrrr}
\toprule
\multirow{2}{*}{Method} & \multicolumn{2}{c}{WikiQA} & \multicolumn{2}{c}{News} & \multicolumn{2}{c}{Story} \\
\cmidrule(lr){2-3}\cmidrule(lr){4-5}\cmidrule(lr){6-7}
& $\WDsup\downarrow$ & $\CE\to0$ & $\WDsup\downarrow$ & $\CE\to0$ & $\WDsup\downarrow$ & $\CE\to0$ \\
\midrule
TextTiling & 0.388{\scriptsize$\pm$0.267} & -0.09{\scriptsize$\pm$0.84} & 1.066{\scriptsize$\pm$0.244} & -4.18{\scriptsize$\pm$0.99} & 0.850{\scriptsize$\pm$0.233} & -3.19{\scriptsize$\pm$0.94} \\
LLMPred & 0.362{\scriptsize$\pm$0.204} & 0.41{\scriptsize$\pm$0.96} & 0.507{\scriptsize$\pm$0.456} & -0.81{\scriptsize$\pm$1.94} & 0.740{\scriptsize$\pm$0.713} & -2.02{\scriptsize$\pm$3.26} \\
PaLD & 0.446{\scriptsize$\pm$0.330} & -1.28{\scriptsize$\pm$1.23} & -- & -- & -- & -- \\
SenPred & 0.271{\scriptsize$\pm$0.291} & -0.90{\scriptsize$\pm$1.11} & 1.663{\scriptsize$\pm$0.829} & -6.74{\scriptsize$\pm$3.18} & 2.972{\scriptsize$\pm$1.071} & -12.51{\scriptsize$\pm$4.29} \\
Voting & 0.282{\scriptsize$\pm$0.277} & \textbf{0.00{\scriptsize$\pm$0.40}} & 0.131{\scriptsize$\pm$0.213} & -0.32{\scriptsize$\pm$0.80} & 0.635{\scriptsize$\pm$0.549} & -2.39{\scriptsize$\pm$2.24} \\
VCP & 0.200{\scriptsize$\pm$0.187} & 0.20{\scriptsize$\pm$0.68} & 0.124{\scriptsize$\pm$0.186} & -0.99{\scriptsize$\pm$1.43} & 0.306{\scriptsize$\pm$0.269} & -1.02{\scriptsize$\pm$1.63} \\
WCP & 0.200{\scriptsize$\pm$0.190} & 0.07{\scriptsize$\pm$0.81} & 0.123{\scriptsize$\pm$0.219} & -1.73{\scriptsize$\pm$1.87} & 0.372{\scriptsize$\pm$0.307} & -2.55{\scriptsize$\pm$2.11} \\
RWCP-R & \textbf{0.184{\scriptsize$\pm$0.170}} & 0.47{\scriptsize$\pm$0.50} & \textbf{0.097{\scriptsize$\pm$0.142}} & \textbf{0.03{\scriptsize$\pm$0.30}} & \textbf{0.235{\scriptsize$\pm$0.163}} & \textbf{0.34{\scriptsize$\pm$0.59}} \\
\bottomrule
\end{tabular}
\end{adjustbox}
\end{table}

The GPT-5-mini aggregate extension is shown separately in Table~\ref{tab:single-gpt-supp}. RWCP-R has lower legacy window error on News and Story, while VCP is lower on WikiQA. These supplied aggregates do not enter family means, ablations, win counts, or paired intervals for the primary study.

\begin{table}[H]
\centering
\caption{Supplemental GPT-5-mini single-change panel: 100 records per domain, as specified by the dataset inventory. Entries preserve the supplied aggregate means $\pm$ sample standard deviations. This panel is outside the primary 25-setting evaluation and its paired comparisons; document-level outputs and matching selection provenance are not included in that evaluation bundle.}
\label{tab:single-gpt-supp}
\small
\setlength{\tabcolsep}{4pt}
\renewcommand{\arraystretch}{1.06}
\begin{adjustbox}{max width=\linewidth}
\begin{tabular}{lrrrrrr}
\toprule
\multirow{2}{*}{Method} & \multicolumn{2}{c}{WikiQA} & \multicolumn{2}{c}{News} & \multicolumn{2}{c}{Story} \\
\cmidrule(lr){2-3}\cmidrule(lr){4-5}\cmidrule(lr){6-7}
& $\WDsup\downarrow$ & $\CE\to0$ & $\WDsup\downarrow$ & $\CE\to0$ & $\WDsup\downarrow$ & $\CE\to0$ \\
\midrule
 TextTiling
			& 0.36{\scriptsize$\pm$0.24}
			& \textbf{-0.02{\scriptsize$\pm$0.81}}
			& 1.32{\scriptsize$\pm$0.30}
			& -5.19{\scriptsize$\pm$0.97}
			& 1.11{\scriptsize$\pm$0.30}
			& -4.39{\scriptsize$\pm$1.06} \\
LLMPred
			& 0.44{\scriptsize$\pm$0.20}
			& 0.09{\scriptsize$\pm$0.64}
			& 0.50{\scriptsize$\pm$0.33}
			& \textbf{-0.59{\scriptsize$\pm$1.46}}
			& 0.52{\scriptsize$\pm$0.46}
			& -0.84{\scriptsize$\pm$2.16} \\
PaLD
			& 0.50{\scriptsize$\pm$0.38}
			& -1.41{\scriptsize$\pm$1.46}
			& 1.76{\scriptsize$\pm$0.75}
			& -6.75{\scriptsize$\pm$3.37}
			& 2.19{\scriptsize$\pm$0.89}
			& -8.47{\scriptsize$\pm$3.82} \\
SenPred
			& 0.52{\scriptsize$\pm$0.38}
			& -1.62{\scriptsize$\pm$1.51}
			& 3.32{\scriptsize$\pm$0.85}
			& -13.36{\scriptsize$\pm$3.09}
			& 3.61{\scriptsize$\pm$1.05}
			& -14.49{\scriptsize$\pm$3.89} \\
Voting
			& 0.39{\scriptsize$\pm$0.28}
			& -0.10{\scriptsize$\pm$0.77}
			& 0.89{\scriptsize$\pm$0.50}
			& -3.17{\scriptsize$\pm$2.15}
			& 1.12{\scriptsize$\pm$0.46}
			& -4.14{\scriptsize$\pm$2.00} \\
VCP
			& \textbf{0.25{\scriptsize$\pm$0.18}}
			& 0.39{\scriptsize$\pm$0.74}
			& 0.41{\scriptsize$\pm$0.26}
			& -0.83{\scriptsize$\pm$1.77}
			& 0.52{\scriptsize$\pm$0.27}
			& -0.69{\scriptsize$\pm$1.88} \\
WCP
			& 0.27{\scriptsize$\pm$0.19}
			& 0.26{\scriptsize$\pm$0.76}
			& 0.44{\scriptsize$\pm$0.26}
			& -1.16{\scriptsize$\pm$1.87}
			& 0.57{\scriptsize$\pm$0.29}
			& -1.85{\scriptsize$\pm$1.99} \\
RWCP-R
			& 0.26{\scriptsize$\pm$0.17}
			& 0.64{\scriptsize$\pm$0.54}
			& \textbf{0.32{\scriptsize$\pm$0.12}}
			& 0.62{\scriptsize$\pm$0.55}
			& \textbf{0.35{\scriptsize$\pm$0.12}}
			& \textbf{0.64{\scriptsize$\pm$0.59}} \\

\bottomrule
\end{tabular}
\end{adjustbox}
\end{table}

\paragraph{Real co-authoring.}
On CoAuthor (Table~\ref{tab:wd_ce_results}), RWCP-R has the smallest
$\WDsup$ mean, 0.4780 versus 0.4804 for VCP. Standard WD is 0.4869
versus 0.4878 for WCP. The table reports count and boundary diagnostics
alongside the window-error gains, allowing the recurrent objective to be
compared with the historical labeling pipeline.

\begin{table}[!htbp]
\centering
\caption{CoAuthor: 290 sessions and 7,042 sentences. $\CE_{\rm label}$ is the historical final-label count error; raw $\CE$, count MAE, and Zero CP use detected boundaries. RWCP-R has the smallest $\WDsup$ mean. Zero CP is descriptive and is not ranked. SegFormer has no local output.}
\label{tab:wd_ce_results}
\label{tab:coauthor-labeling}
\small
\setlength{\tabcolsep}{4pt}
\renewcommand{\arraystretch}{1.08}
\begin{adjustbox}{max width=\linewidth}
\begin{tabular}{lrrrrr}
\toprule
\multirow{2}{*}{Method} & \multicolumn{2}{c}{Final labels} & \multicolumn{3}{c}{Raw boundaries} \\
\cmidrule(lr){2-3}\cmidrule(lr){4-6}
& $\WDsup\downarrow$ & $\CE_{\rm label}\to0$ & $\CE\to0$ & MAE$\downarrow$ & Zero CP \\
\midrule
TextTiling & 0.551 & 7.31 & 7.31 & 7.62 & 0.0\% \\
LLMPred & 0.516 & 8.30 & 8.30 & 8.51 & 19.0\% \\
SenPred & 0.809 & -4.16 & -4.16 & 5.58 & 0.0\% \\
Voting & 0.647 & \textbf{3.29} & \textbf{3.29} & \textbf{5.27} & 0.0\% \\
VCP & 0.480 & 9.42 & 9.37 & 9.40 & 61.7\% \\
WCP & 0.483 & 7.98 & 7.72 & 7.92 & 15.9\% \\
\midrule
RWCP-R & \textbf{0.478} & 10.17 & 10.17 & 10.19 & 84.1\% \\
\bottomrule
\end{tabular}
\end{adjustbox}
\end{table}

\paragraph{Multiple changes.}
RWCP-R leads in five of ten $\WDsup$ settings
(Table~\ref{tab:claude_gpt_results}): Claude $K=1$ and GPT-5-mini
$K=1,3,5,8$. VCP leads at Claude $K=2,3,5$, Voting at Claude $K=8$,
and WCP at GPT-5-mini $K=2$. Against the previous block configuration,
mean standard WD improves from 0.4112 to 0.3993 and count MAE from
3.628 to 3.233. The full recurrent pipeline improves these averages over the historical block configuration, but this comparison does not isolate the effect of shared states. WCP still has lower count MAE and higher F1 in this family.

\begin{table}[!htbp]
\centering
\caption{Multiple-change Story results: 100 documents per generator and true $K$. The displayed count is not provided to unknown-count detectors. Entries are mean $\WDsup$ and raw $\CE$. RWCP-R leads in five of ten WD settings.}
\label{tab:claude_gpt_results}
\label{tab:multiple-change}
\small
\setlength{\tabcolsep}{4pt}
\renewcommand{\arraystretch}{1.08}
\begin{adjustbox}{max width=\linewidth}
\begin{tabular}{lrrrrrrrrrr}
\toprule
\multirow{2}{*}{Method} & \multicolumn{2}{c}{$K=1$} & \multicolumn{2}{c}{$K=2$} & \multicolumn{2}{c}{$K=3$} & \multicolumn{2}{c}{$K=5$} & \multicolumn{2}{c}{$K=8$} \\
\cmidrule(lr){2-3}\cmidrule(lr){4-5}\cmidrule(lr){6-7}\cmidrule(lr){8-9}\cmidrule(lr){10-11}
& $\WDsup\downarrow$ & $\CE\to0$ & $\WDsup\downarrow$ & $\CE\to0$ & $\WDsup\downarrow$ & $\CE\to0$ & $\WDsup\downarrow$ & $\CE\to0$ & $\WDsup\downarrow$ & $\CE\to0$ \\
\midrule
\multicolumn{11}{l}{\textit{Claude Haiku 4.5}} \\
\addlinespace[2pt]
TextTiling & 0.898 & -3.35 & 0.702 & -3.57 & 0.711 & -4.49 & 0.588 & -4.66 & 0.483 & -2.90 \\
LLMPred & 0.714 & -1.99 & 0.689 & -1.71 & 0.733 & -1.83 & 0.624 & 0.42 & 0.504 & 4.84 \\
SenPred & 2.711 & -11.28 & 2.712 & -16.79 & 2.619 & -22.17 & 2.183 & -27.81 & 1.583 & -29.68 \\
Voting & 0.476 & -1.85 & 0.464 & -2.15 & 0.475 & -3.22 & 0.459 & -4.29 & \textbf{0.348} & -3.30 \\
VCP & 0.293 & -1.13 & \textbf{0.319} & -1.02 & \textbf{0.315} & -1.32 & \textbf{0.351} & -1.53 & 0.376 & \textbf{0.35} \\
WCP & 0.313 & \textbf{-0.52} & 0.337 & \textbf{-0.19} & 0.339 & \textbf{-0.11} & 0.369 & \textbf{0.39} & 0.383 & 2.85 \\
\midrule
RWCP-R & \textbf{0.273} & 0.58 & 0.360 & 1.44 & 0.354 & 2.10 & 0.397 & 4.26 & 0.409 & 7.21 \\
\midrule
\multicolumn{11}{l}{\textit{GPT-5-mini}} \\
\addlinespace[2pt]
TextTiling & 1.120 & -4.39 & 0.905 & -4.66 & 0.882 & -6.12 & 0.734 & -7.01 & 0.536 & -4.70 \\
LLMPred & 0.645 & -1.52 & 0.778 & -2.29 & 0.696 & -1.16 & 0.681 & \textbf{-0.16} & 0.542 & 3.99 \\
SenPred & 3.444 & -13.56 & 3.275 & -19.45 & 3.225 & -25.81 & 2.685 & -33.02 & 1.894 & -35.41 \\
Voting & 1.128 & -4.09 & 0.997 & -5.25 & 1.018 & -6.78 & 0.874 & -8.29 & 0.675 & -7.04 \\
VCP & 0.445 & -0.95 & 0.458 & -1.05 & 0.451 & -0.82 & 0.462 & -0.35 & 0.454 & \textbf{2.03} \\
WCP & 0.423 & \textbf{-0.38} & \textbf{0.404} & \textbf{0.01} & 0.450 & \textbf{0.31} & 0.438 & 1.26 & 0.438 & 4.10 \\
\midrule
RWCP-R & \textbf{0.342} & 0.79 & 0.419 & 1.51 & \textbf{0.419} & 2.42 & \textbf{0.427} & 4.28 & \textbf{0.429} & 7.22 \\
\bottomrule
\end{tabular}
\end{adjustbox}
\end{table}

\paragraph{Text perturbations.}
RWCP-R has the lowest $\WDsup$ in five of six attack settings
(Table~\ref{tab:decoherence_paraphrasing_results}); Voting remains best
on paraphrased News. Standard WD falls from 0.3273 for RWCP-block to
0.2676, and count MAE from 0.785 to 0.602. Unlike the earlier block
configuration, RWCP-R improves standard WD against WCP in both attacked
WikiQA conditions; all six pointwise paired intervals against WCP are
negative across the cached perturbations.

\begin{table}[!htbp]
\centering
\caption{Text-attack results: 100 records in each of WikiQA, News, and Story per attack. Predictions for nontrivial inputs are unchanged; singleton boundary conventions match Table~\ref{tab:single_cp_results}. RWCP-R has the lowest $\WDsup$ in five of six settings; Voting leads on paraphrased News.}
\label{tab:decoherence_paraphrasing_results}
\label{tab:text-perturbations-rwcp}
\small
\setlength{\tabcolsep}{4pt}
\renewcommand{\arraystretch}{1.06}
\begin{adjustbox}{max width=\linewidth}
\begin{tabular}{lrrrrrr}
\toprule
\multirow{2}{*}{Method} & \multicolumn{2}{c}{WikiQA} & \multicolumn{2}{c}{News} & \multicolumn{2}{c}{Story} \\
\cmidrule(lr){2-3}\cmidrule(lr){4-5}\cmidrule(lr){6-7}
& $\WDsup\downarrow$ & $\CE\to0$ & $\WDsup\downarrow$ & $\CE\to0$ & $\WDsup\downarrow$ & $\CE\to0$ \\
\midrule
\multicolumn{7}{l}{Decoherence} \\
\midrule
TextTiling & 0.388 & -0.09 & 1.066 & -4.18 & 0.844 & -3.17 \\
LLMPred & 0.352 & 0.37 & 0.604 & -1.22 & 0.957 & -2.96 \\
SenPred & 0.383 & -1.30 & 2.573 & -10.58 & 3.119 & -12.89 \\
Voting & 0.337 & \textbf{-0.03} & 0.491 & -1.75 & 0.993 & -3.82 \\
WCP & 0.254 & 0.44 & 0.328 & \textbf{-0.59} & 0.364 & \textbf{-0.45} \\
RWCP-R & \textbf{0.226} & 0.65 & \textbf{0.324} & 0.60 & \textbf{0.328} & 0.59 \\
\midrule
\multicolumn{7}{l}{Paraphrasing} \\
\midrule
TextTiling & 0.389 & -0.06 & 1.049 & -4.16 & 0.839 & -3.21 \\
LLMPred & 0.368 & 0.30 & 0.445 & -0.44 & 0.737 & -2.13 \\
SenPred & 0.294 & -0.96 & 1.590 & -6.53 & 2.857 & -12.08 \\
Voting & 0.264 & \textbf{0.04} & \textbf{0.107} & -0.24 & 0.581 & -2.23 \\
WCP & 0.187 & 0.25 & 0.167 & -1.05 & 0.336 & -0.75 \\
RWCP-R & \textbf{0.148} & 0.37 & 0.141 & \textbf{0.12} & \textbf{0.303} & \textbf{0.52} \\
\bottomrule
\end{tabular}
\end{adjustbox}
\end{table}

\paragraph{LLM-written proportion.}
RWCP-R leads in $\WDsup$ at 5\%, 10\%, 40\%, and 80\%, with WCP
leading at 20\% (Table~\ref{tab:percentage_results}). The setting-average
standard WD is 0.2119, slightly worse than RWCP-block's 0.2084, while
count MAE improves from 0.874 for RWCP-block to 0.736. The consistent gains
across four target proportions indicate that the recurrent profile remains
effective as the LLM-written fraction varies.

\begin{table}[!htbp]
\centering
\caption{Story with varying target LLM-written fractions, 100 documents per setting. RWCP-R leads in $\WDsup$ at 5\%, 10\%, 40\%, and 80\%; WCP leads at 20\%. Neither the target fraction nor the true boundary is supplied to RWCP-R.}
\label{tab:percentage_results}
\small
\setlength{\tabcolsep}{4pt}
\renewcommand{\arraystretch}{1.08}
\begin{adjustbox}{max width=\linewidth}
\begin{tabular}{lrrrrrrrrrr}
\toprule
\multirow{2}{*}{Method} & \multicolumn{2}{c}{5\%} & \multicolumn{2}{c}{10\%} & \multicolumn{2}{c}{20\%} & \multicolumn{2}{c}{40\%} & \multicolumn{2}{c}{80\%} \\
\cmidrule(lr){2-3}\cmidrule(lr){4-5}\cmidrule(lr){6-7}\cmidrule(lr){8-9}\cmidrule(lr){10-11}
& $\WDsup\downarrow$ & $\CE\to0$ & $\WDsup\downarrow$ & $\CE\to0$ & $\WDsup\downarrow$ & $\CE\to0$ & $\WDsup\downarrow$ & $\CE\to0$ & $\WDsup\downarrow$ & $\CE\to0$ \\
\midrule
TextTiling & 1.166 & -3.92 & 1.155 & -4.15 & 1.048 & -4.15 & 1.005 & -3.68 & 0.766 & -2.75 \\
LLMPred & 0.620 & -1.50 & 0.700 & -1.70 & 0.698 & -1.26 & 0.692 & -1.26 & 0.525 & -1.06 \\
SenPred & 4.500 & -16.48 & 4.264 & -15.64 & 3.880 & -14.77 & 3.233 & -13.18 & 3.134 & -12.43 \\
Voting & 1.503 & -5.04 & 0.768 & -2.64 & 0.671 & -2.40 & 0.629 & -2.46 & 0.943 & -3.28 \\
WCP & 0.213 & -1.02 & 0.235 & -0.95 & \textbf{0.228} & -1.13 & 0.289 & -0.82 & 0.275 & \textbf{-0.10} \\
\midrule
RWCP-R & \textbf{0.111} & \textbf{0.79} & \textbf{0.185} & \textbf{0.49} & 0.243 & \textbf{0.48} & \textbf{0.277} & \textbf{0.52} & \textbf{0.271} & 0.72 \\
\bottomrule
\end{tabular}
\end{adjustbox}
\end{table}

\subsection{Recovery diagnostics and component ablations}
\label{app:component-diagnostics}

\subsubsection{Family-wise comparison of RWCP variants}
Table~\ref{tab:standard-wd-summary} reports the family-wise results for the preceding analytic and block-calibrated variants alongside RWCP-R.
RWCP-R has the lowest WD in four families, while the block variant is lower in the LLM-proportion family.
The comparisons retain historical configurations and do not isolate state sharing.

\begin{table}[!htbp]
\centering
\caption{Standard raw-boundary WD, equally averaged over settings within families. Best baseline requires complete family coverage. Analytic and Block are historical fixed configurations; their differences from RWCP-R include search, scale, and other settings. This comparison is not an isolated ablation of recurrence.}
\label{tab:standard-wd-summary}
\small
\setlength{\tabcolsep}{4pt}
\renewcommand{\arraystretch}{1.06}
\begin{adjustbox}{max width=\linewidth}
\begin{tabular}{llrrrr}
\toprule
\multirow{2}{*}{Family} & \multicolumn{2}{c}{Best baseline} & Analytic & Block & RWCP-R \\
\cmidrule(lr){2-3}\cmidrule(lr){4-4}\cmidrule(lr){5-5}\cmidrule(lr){6-6}
& Method & WD & WD & WD & WD \\
\midrule
Single change & Voting & 0.2725 & 0.2853 & 0.2441 & \textbf{0.1895} \\
CoAuthor & WCP & 0.4878 & 0.4922 & 0.4945 & \textbf{0.4869} \\
Multiple changes & WCP & 0.4363 & 0.4094 & 0.4112 & \textbf{0.3993} \\
Text attacks & Voting & 0.3514 & 0.3307 & 0.3273 & \textbf{0.2676} \\
LLM proportion & WCP & 0.3538 & 0.2155 & \textbf{0.2084} & 0.2119 \\
\bottomrule
\end{tabular}
\end{adjustbox}
\end{table}

\subsubsection{Recovery stress tests}
\label{subsec:recovery-diagnostics}
The main text emphasizes WD and signed CE. Table~\ref{tab:count-diagnostics} additionally reports absolute count error, matched-boundary F1, the no-boundary fraction, and the WD of a predictor that returns no boundary.
These diagnostics distinguish a lower window error from successful boundary recovery and expose count errors that signed averaging can cancel.

\begin{table}[!htbp]
\centering
\caption{Raw-boundary recovery diagnostics, equally averaged over settings. F1 matches boundaries within one observation; zero CP is the no-boundary fraction and Null WD is the no-boundary predictor. The complete-record convention applies throughout.}
\label{tab:count-diagnostics}
\label{tab:recovery-stress}
\small
\setlength{\tabcolsep}{4pt}
\renewcommand{\arraystretch}{1.06}
\begin{adjustbox}{max width=\linewidth}
\begin{tabular}{lrrrrrr}
\toprule
\multirow{2}{*}{Family} & \multicolumn{2}{c}{WCP} & \multicolumn{3}{c}{RWCP-R} & No boundary \\
\cmidrule(lr){2-3}\cmidrule(lr){4-6}\cmidrule(lr){7-7}
& MAE & F1 & MAE & F1 & Zero CP & WD \\
\midrule
Single change & 1.637 & 0.435 & 0.340 & 0.544 & 31.7\% & 0.3448 \\
CoAuthor & 7.917 & 0.302 & 10.186 & 0.058 & 84.1\% & 0.4893 \\
Multiple changes & 2.130 & 0.205 & 3.233 & 0.097 & 63.9\% & 0.4181 \\
Text attacks & 0.998 & 0.272 & 0.602 & 0.300 & 54.5\% & 0.3448 \\
LLM proportion & 1.360 & 0.220 & 0.736 & 0.151 & 66.8\% & 0.2141 \\
\bottomrule
\end{tabular}
\end{adjustbox}
\end{table}

Single changes and text attacks improve on the no-boundary predictor by 0.1552 and 0.0772 WD and improve boundary F1 relative to WCP by 0.110 and 0.028.
CoAuthor and LLM proportion are within 0.0024 WD of the no-boundary predictor and return no boundary in 84.1\% and 66.8\% of instances.
Their F1 differences against WCP are $-0.244$ and $-0.069$; multiple-change data improve modestly on the null WD but lose 0.108 F1.
Across the five families, count MAE is 3.019 for RWCP-R versus 2.808 for WCP, and F1 is 0.230 versus 0.287.
Thus the primary window-error improvement is accompanied by worse aggregate count and boundary diagnostics. Table~\ref{tab:macro-uncertainty} gives the paired intervals.

\subsubsection{Complete component diagnostics}
The full component comparison in Table~\ref{tab:ablation} supplements the WD/CE presentation in Table~\ref{tab:main-ablation}.
Scale adaptation improves both window errors, count MAE, and boundary F1 at these configurations.
Uniform weights improve count MAE and both F1 variants while nearly matching standard WD; the quadratic approximation also improves count MAE and F1 relative to the full method.
These fixed-configuration comparisons support metric-specific conclusions rather than a claim that every component improves localization.

\begin{table}[!htbp]
\centering
\caption{Frozen component ablations, averaged equally over settings and then five families. Fixed scale uses $\sigma=1$; uniform weights uses exponent zero; the quadratic approximation uses $c=10^6$. No variant is retuned. Bold marks the best unrounded value per column; it does not indicate significance.}
\label{tab:ablation}
\small
\setlength{\tabcolsep}{4pt}
\renewcommand{\arraystretch}{1.06}
\begin{adjustbox}{max width=\linewidth}
\begin{tabular}{lrrrrr}
\toprule
\multirow{2}{*}{Variant} & \multicolumn{2}{c}{Window error} & Count error & \multicolumn{2}{c}{Boundary F1$\uparrow$} \\
\cmidrule(lr){2-3}\cmidrule(lr){4-4}\cmidrule(lr){5-6}
& $\WDsup\downarrow$ & WD$\downarrow$ & MAE$\downarrow$ & $h=0$ & $h=1$ \\
\midrule
RWCP-analytic & 0.3233 & 0.3466 & 3.155 & 0.115 & 0.176 \\
RWCP-block & 0.3158 & 0.3371 & 3.200 & 0.111 & 0.169 \\
RWCP-R & \textbf{0.2990} & \textbf{0.3110} & 3.019 & 0.163 & 0.230 \\
R: fixed scale & 0.3038 & 0.3170 & 3.071 & 0.148 & 0.211 \\
R: uniform weights & 0.3058 & 0.3118 & \textbf{2.917} & \textbf{0.189} & \textbf{0.261} \\
R: quadratic loss & 0.3130 & 0.3179 & 2.945 & 0.166 & 0.239 \\
\bottomrule
\end{tabular}
\end{adjustbox}
\end{table}

\subsubsection{Post-freeze internal stability}
Table~\ref{tab:post-freeze-diagnostics} reports the historical source-group development and internal-check split.
Against the block variant, family-macro WD changes from 0.3374 to 0.2990 on development and from 0.3380 to 0.3165 on the internal check.
RWCP-R improves four of five check families; LLM proportion is the exception.
These sources had already influenced development and do not constitute a fresh independent test.
Code checks verify Huber interval fits, fixed-center decoding against exhaustive enumeration, nonincreasing alternating objectives, score/weight unit invariance, known-count behavior, and boundary/label consistency.
Production reruns reproduce the frozen predictions across all five families. Runs use CPU with optional Numba acceleration and cached scores; timing excludes model inference and data generation.
Fresh-source evaluation, no-change error calibration on actual detector scores, and matched whole-segment labeling remain necessary for stronger generalization and mechanism claims.

\begin{table}[!htbp]
\centering
\caption{Post-freeze stability by the existing source-group split. Development (716 nontrivial instances) supplies parameter selection; internal check (1,968 nontrivial instances) follows configuration freezing. Six reporting-only singletons are outside these historical splits. Both subsets come from benchmarks previously used in research development, and the latter is not a new independent test set. Entries are standard WD.}
\label{tab:post-freeze-diagnostics}
\small
\setlength{\tabcolsep}{4pt}
\renewcommand{\arraystretch}{1.08}
\begin{adjustbox}{max width=\linewidth}
\begin{tabular}{lrrrr}
\toprule
\multirow{2}{*}{Experiment} & \multicolumn{2}{c}{RWCP-block} & \multicolumn{2}{c}{RWCP-R} \\
\cmidrule(lr){2-3}\cmidrule(lr){4-5}
& Dev. WD & Check WD & Dev. WD & Check WD \\
\midrule
Single change & 0.2501 & 0.2445 & 0.1959 & 0.1895 \\
CoAuthor & 0.4803 & 0.4997 & 0.4729 & 0.4921 \\
Multiple changes & 0.4075 & 0.4125 & 0.3861 & 0.4036 \\
Text attacks & 0.3352 & 0.3269 & 0.2573 & 0.2749 \\
LLM proportion & 0.2138 & 0.2064 & 0.1827 & 0.2224 \\
\midrule
Family macro & 0.3374 & 0.3380 & 0.2990 & 0.3165 \\
\bottomrule
\end{tabular}
\end{adjustbox}
\end{table}

\FloatBarrier

\subsection{Algorithms and Experimental Details}
\label{app:implementation}

This appendix describes the core Huber solver and block calibration, followed by the metrics used in the completed cached-score evaluation. Algorithm~\ref{alg:recurrent} specifies the distinct recurrent solver used for the main empirical results.

\paragraph{One-dimensional Huber optimization.}

For interval $A$, define the monotone score
\begin{equation*}
G_A(\theta)
=
\sum_{i\in A}\sqrt{w_i}\,
\psi_c\{\sqrt{w_i}(Y_i-\theta)\}.
\end{equation*}
Because $\psi_c$ is nondecreasing in its argument and $Y_i-\theta$ decreases with $\theta$, $G_A$ is nonincreasing. An interior minimizer satisfies $G_A(\theta)=0$. If $G_A(\theta)>0$ throughout $\Theta$, the minimizer is the right endpoint; if it is negative throughout, the minimizer is the left endpoint.

\begin{algorithm}[t]
\caption{Bisection for an interval Huber minimizer}
\label{alg:huber-bisection}
\begin{algorithmic}[1]
\Require Interval $A$, scores, weights, $c$, parameter range $[L,U]$, tolerance $\epsilon_{\mathrm{opt}}$.
\If{$G_A(L)\leq0$}
\State \Return $L$.
\ElsIf{$G_A(U)\geq0$}
\State \Return $U$.
\EndIf
\While{$U-L>\epsilon_{\mathrm{opt}}$}
\State $M\gets(L+U)/2$.
\If{$G_A(M)>0$}
\State $L\gets M$.
\Else
\State $U\gets M$.
\EndIf
\EndWhile
\State \Return $(L+U)/2$.
\end{algorithmic}
\end{algorithm}

\begin{proposition}[Optimization accuracy]
\label{prop:optimization-accuracy}
Suppose the empirical loss on $A$ is $\alpha S_A^w$-strongly convex in a neighborhood containing the exact minimizer $\wh\theta_A$ and numerical solution $\wt\theta_A$, and suppose $\wh\theta_A\in\operatorname{int}(\Theta)$. If $|\wt\theta_A-\wh\theta_A|\leq\epsilon_{\mathrm{opt}}$, then
\begin{equation*}
0\leq
\wh\cL_A(\wt\theta_A)-\wh\cL_A(\wh\theta_A)
\leq\frac12S_A^w\epsilon_{\mathrm{opt}}^2.
\end{equation*}
For a profile gain, the total numerical error is at most
\begin{equation}
2S_{s:e}^w\epsilon_{\mathrm{opt}}^2.
\label{eq:gain-numerical-error}
\end{equation}
\end{proposition}

\begin{proof}
The Huber empirical loss has gradient Lipschitz constant $S_A^w$ by Lemma~\ref{lem:local-empirical-strong}. Since the exact minimizer has zero subgradient in the interior, smoothness gives the first upper bound. The profile gain contains one parent and two child optimized losses. Their masses sum to $2S_{s:e}^w$ after applying the factor two in \eqref{eq:profile-gain}, which yields the conservative bound \eqref{eq:gain-numerical-error}.
\end{proof}

If the constrained optimum is a boundary point, Algorithm~\ref{alg:huber-bisection} returns that endpoint exactly from the score-sign check; the quadratic numerical-error proposition is used only on the interior branch.

To keep numerical error below the statistical threshold, it is enough to choose
\begin{equation*}
\epsilon_{\mathrm{opt}}
\leq c_{\mathrm{opt}}
\sqrt{\frac{\Lambda_N}{W_N}}.
\end{equation*}
In practice a fixed tolerance such as $10^{-6}$ is usually much smaller.

\paragraph{Warm starts and cached evaluations.}

For neighboring candidate splits $b$ and $b+1$, the left and right intervals differ by one observation. Their Huber minimizers typically change smoothly. Use $\wh\theta_{s:b}$ to initialize the solve for $[s,b+1]$ and $\wh\theta_{b+1:e}$ for $[b+2,e]$. A safeguarded Newton step is
\begin{equation*}
\theta^{(r+1)}
=
\Pi_\Theta\left[
\theta^{(r)}+
\frac{G_A(\theta^{(r)})}
{H_A(\theta^{(r)})\vee h_{\min}}
\right],
\end{equation*}
where $H_A$ is defined in \eqref{eq:empirical-hessian}. If the proposed step does not reduce the loss, revert to bisection. This hybrid retains deterministic convergence while exploiting local quadratic behavior.

For a fixed iterate $\theta$, the score and Hessian can be evaluated from prefix sums of
\begin{equation*}
a_i(\theta)=\sqrt{w_i}\psi_c\{\sqrt{w_i}(Y_i-\theta)\},
\qquad
h_i(\theta)=w_i\one\{|\sqrt{w_i}(Y_i-\theta)|<c\}.
\end{equation*}
Because these quantities depend on $\theta$, caching is most effective within batches of nearby intervals sharing warm starts.

\paragraph{Complexity accounting.}

Let $L_m=e_m-s_m+1$ be the length of random interval $m$ and let $J$ be the average number of score evaluations required by the one-dimensional solver. A direct scan has cost
\begin{equation*}
O\left(
J\sum_{m=1}^M L_m^2
\right)
\end{equation*}
because each of $O(L_m)$ splits requires losses over $O(L_m)$ observations. With cached interval data and warm-started score evaluations, the practical cost is closer to
\begin{equation*}
O\left(
J_{\mathrm{warm}}\sum_{m=1}^M L_m
+\sum_{m=1}^M L_m\log L_m
\right),
\end{equation*}
although this is an implementation-dependent empirical complexity rather than a proved worst-case bound. Memory is $O(N+M)$ excluding detector inference. Block calibration multiplies the segmentation-statistic cost by approximately $Q$, but repetitions are embarrassingly parallel.

\paragraph{Weighted interval generation.}

Generate cumulative weights once and use binary search in $W(1),\ldots,W(N)$ to map a uniform coordinate to an index. One interval costs $O(\log N)$ time, so pre-sampling $M$ intervals costs $O(M\log N)$. Deduplicate repeated endpoint pairs before profile evaluation. The random seed and the final endpoint list should be stored to make every comparison deterministic across segmentation methods.

\paragraph{One-step multiplier profile approximation.}

For a homogeneous pilot interval $A$, the first-order Huber estimator expansion is
\begin{equation*}
\wh\theta_A-\theta_A
\approx
\frac{\sum_{i\in A}\zeta_i}{H_A(\theta_A)},
\end{equation*}
where $\zeta_i$ is the centered score. In a bootstrap repetition, replace the score sum by $\sum_{i\in A}\zeta_i^\star$ and curvature by its pilot estimate. The approximate quadratic profile gain is then
\begin{equation}
\Gamma_{s,e}^{\star,\mathrm{1step}}(b)
=
\frac{
\left
(
H_{s:b}^{-1}\sum_{i=s}^b\zeta_i^\star
-
H_{b+1:e}^{-1}\sum_{i=b+1}^e\zeta_i^\star
\right)^2
}
{H_{s:b}^{-1}+H_{b+1:e}^{-1}},
\label{eq:one-step-profile}
\end{equation}
where $H_{a:b}$ denotes the pilot Hessian sum. Formula \eqref{eq:one-step-profile} is the robust analogue of a squared standardized difference. It is used only for calibration, not in the main estimator.

\subsubsection{Metric definitions and reproducibility}
\label{app:experimental-protocol}
\label{app:reproducibility}
\paragraph{WindowDiff and boundary metrics.}

Let $C_i^{(r)}$ and $C_i^{(h)}$ be the true and estimated numbers of boundaries in window $[i,i+k]$. WindowDiff is
\begin{equation}
\WD
=
\frac{1}{N-k}
\sum_{i=1}^{N-k}
\one\{C_i^{(r)}\neq C_i^{(h)}\}.
\label{eq:windowdiff}
\end{equation}
For document $d$, use the pre-specified width $k=k_d$ in
\eqref{eq:windowdiff-width}; this keeps the implementation identical across
methods while adapting to reference segment length.
Count Error is
\begin{equation*}
\CE=K-\wh K.
\end{equation*}
Match an estimate to a true boundary when $|\wh\tau-\tau|\leq h$, using maximum bipartite matching to avoid multiple credit. The completed RWCP-R evaluation reports F1 for $h=1$ observation and the exact-match case $h=0$. Precision, recall, and F1 follow from matched, unmatched estimated, and unmatched true boundaries.

The empirical evidence consists of saved score streams, predictions, source-group assignments, selected configurations, and metric recomputation. The reporting correction includes every controlled record and explicitly records the singleton boundary convention. It preserves the historical development/check memberships and all nontrivial predictions. Generation revisions, complete original-source identifiers, and end-to-end detector timings are not fully available; the tables therefore establish cache-level reproducibility. No fitted detector-score tail law, empirical mixing verification, or calibrated no-change false-positive rate is claimed here. Those missing measurements delimit the connection between the score-model assumptions and these applications.

\subsubsection{Uncertainty in aggregate and component differences}
The per-setting intervals condition on one setting at a time. To assess the family-macro contrasts, we additionally resample the available source components jointly across the entire primary suite. Each draw assigns one multiplicity to a component and reuses it in every setting where that component occurs. We then recompute document means within settings, equally average settings within each family, and equally average the five families. This preserves known cross-setting reuse. The corrected reporting inventory contains 647 metadata-linked components: 643 historical components plus four from the singleton records under the same exact-source/surviving-human-text linkage rule. Missing source identifiers still preclude a claim of full independence.

\begin{table}[!htbp]
\centering
\caption{Exploratory family-macro differences: RWCP-R minus comparator, with pointwise paired 95\% source-component bootstrap intervals (2,000 draws, seed 121209). Lower WD/MAE and higher F1 are better. Configurations are fixed, linkage is incomplete, and there is no multiplicity correction.}
\label{tab:macro-uncertainty}
\small
\setlength{\tabcolsep}{4pt}
\begin{adjustbox}{max width=\linewidth}
\begin{tabular}{lrrr}
\toprule
\multirow{2}{*}{Comparator} & \multicolumn{3}{c}{RWCP-R minus comparator: paired 95\% intervals} \\
\cmidrule(lr){2-4}
& $\Delta$WD [95\% CI] & $\Delta$MAE [95\% CI] & $\Delta$F1 [95\% CI] \\
\midrule
WCP & -0.0959 [-0.1074, -0.0844] & +0.211 [+0.126, +0.293] & -0.057 [-0.073, -0.040] \\
R: fixed scale & -0.0060 [-0.0099, -0.0021] & -0.052 [-0.071, -0.033] & +0.019 [+0.010, +0.027] \\
R: uniform weights & -0.0008 [-0.0052, +0.0038] & +0.103 [+0.083, +0.124] & -0.031 [-0.040, -0.023] \\
R: quadratic approximation & -0.0068 [-0.0108, -0.0033] & +0.074 [+0.052, +0.098] & -0.009 [-0.016, -0.001] \\
\bottomrule
\end{tabular}
\end{adjustbox}
\end{table}
Table~\ref{tab:macro-uncertainty} sharpens the scope of the benefit. Against WCP, the window-error improvement coexists with higher count MAE and lower boundary F1. Against uniform weights, the WD interval includes zero while the count and F1 intervals favor uniform weights. Against fixed scale, all three intervals favor scale adaptation at the selected configurations. The large-cutoff quadratic approximation trades worse WD for better count and F1. These are conditional comparisons of the saved configurations, not evidence from a newly sampled test corpus or equally retuned alternative models.

\subsubsection{Per-setting paired results}
\label{app:paired-results}
The complete standard-WD paired table is Table~\ref{tab:main-setting-results}; the preceding appendix tables retain
the fuller comparator-specific and legacy-metric views.

\begingroup
\small
\setlength{\tabcolsep}{4pt}
\renewcommand{\arraystretch}{1.06}
\setlength{\LTcapwidth}{\linewidth}
\begin{longtable}{llrrrr}
\caption{\normalsize Primary 25-setting study: standard raw-boundary WD and RWCP-R F1 (tolerance one). Every controlled setting includes 100 records; CoAuthor has 290. Intervals are pointwise paired 95\% source-component bootstrap intervals for RWCP-R minus WCP (2,000 draws); negative values favor RWCP-R. Singleton boundary conventions are applied to all methods.}
\label{tab:main-setting-results}\\
\toprule
\multirow{2}{*}{Family} & \multirow{2}{*}{Setting} & WCP & \multicolumn{2}{c}{RWCP-R} & RWCP-R--WCP \\
\cmidrule(lr){3-3}\cmidrule(lr){4-5}\cmidrule(lr){6-6}
& & WD & WD & F1 ($h=1$) & $\Delta$WD [95\% CI] \\
\midrule
\endfirsthead
\multicolumn{6}{c}{\tablename~\thetable{} (continued)}\\
\toprule
\multirow{2}{*}{Family} & \multirow{2}{*}{Setting} & WCP & \multicolumn{2}{c}{RWCP-R} & RWCP-R--WCP \\
\cmidrule(lr){3-3}\cmidrule(lr){4-5}\cmidrule(lr){6-6}
& & WD & WD & F1 ($h=1$) & $\Delta$WD [95\% CI] \\
\midrule
\endhead
\midrule
\multicolumn{6}{r}{\footnotesize Continued on next page}\\
\endfoot
\bottomrule
\endlastfoot
Single & WikiQA & 0.284 & 0.223 & 0.490 & -0.060 [-0.107, -0.013] \\
Single & News & 0.370 & 0.102 & 0.790 & -0.268 [-0.327, -0.210] \\
Single & Story & 0.534 & 0.243 & 0.353 & -0.291 [-0.342, -0.238] \\
\midrule
CoAuthor & 290 sessions & 0.488 & 0.487 & 0.058 & -0.001 [-0.019, 0.015] \\
\midrule
Multiple & Claude $K=1$ & 0.400 & 0.288 & 0.180 & -0.112 [-0.156, -0.072] \\
Multiple & Claude $K=2$ & 0.404 & 0.374 & 0.140 & -0.031 [-0.063, 0.000] \\
Multiple & Claude $K=3$ & 0.425 & 0.378 & 0.178 & -0.047 [-0.075, -0.019] \\
Multiple & Claude $K=5$ & 0.425 & 0.429 & 0.107 & 0.004 [-0.018, 0.024] \\
Multiple & Claude $K=8$ & 0.430 & 0.448 & 0.079 & 0.018 [0.002, 0.036] \\
Multiple & GPT $K=1$ & 0.434 & 0.339 & 0.047 & -0.095 [-0.131, -0.060] \\
Multiple & GPT $K=2$ & 0.430 & 0.408 & 0.055 & -0.022 [-0.047, 0.003] \\
Multiple & GPT $K=3$ & 0.470 & 0.418 & 0.071 & -0.052 [-0.071, -0.032] \\
Multiple & GPT $K=5$ & 0.475 & 0.447 & 0.066 & -0.028 [-0.045, -0.012] \\
Multiple & GPT $K=8$ & 0.471 & 0.464 & 0.044 & -0.006 [-0.019, 0.007] \\
\midrule
Attack & WikiQA decoh. & 0.340 & 0.291 & 0.277 & -0.049 [-0.086, -0.011] \\
Attack & News decoh. & 0.407 & 0.335 & 0.067 & -0.072 [-0.117, -0.030] \\
Attack & Story decoh. & 0.398 & 0.335 & 0.067 & -0.063 [-0.105, -0.021] \\
Attack & WikiQA para. & 0.266 & 0.189 & 0.563 & -0.077 [-0.124, -0.034] \\
Attack & News para. & 0.326 & 0.146 & 0.648 & -0.180 [-0.236, -0.126] \\
Attack & Story para. & 0.428 & 0.310 & 0.175 & -0.118 [-0.157, -0.079] \\
\midrule
Proportion & 5\% & 0.321 & 0.104 & 0.037 & -0.217 [-0.270, -0.169] \\
Proportion & 10\% & 0.338 & 0.172 & 0.183 & -0.166 [-0.219, -0.118] \\
Proportion & 20\% & 0.375 & 0.235 & 0.210 & -0.140 [-0.190, -0.097] \\
Proportion & 40\% & 0.404 & 0.278 & 0.203 & -0.126 [-0.174, -0.079] \\
Proportion & 80\% & 0.330 & 0.270 & 0.122 & -0.060 [-0.097, -0.024] \\
\end{longtable}
\endgroup
\begingroup
\small
\setlength{\tabcolsep}{4pt}
\renewcommand{\arraystretch}{1.06}
\setlength{\LTcapwidth}{\linewidth}
\begin{longtable}{lllrrrr}
\caption{\normalsize Primary 25-setting legacy final-label window error $\WDsup$ and RWCP-R raw count error. Best base is the smallest available non-RWCP mean in each setting. Bold marks the smaller legacy error between it and RWCP-R before rounding. These are distinct from raw-boundary WD and matched-boundary F1.}
\label{tab:main-legacy-results}\\
\toprule
\multirow{2}{*}{Family} & \multirow{2}{*}{Setting} & \multicolumn{2}{c}{Best baseline} & WCP & \multicolumn{2}{c}{RWCP-R} \\
\cmidrule(lr){3-4}\cmidrule(lr){5-5}\cmidrule(lr){6-7}
& & Method & $\WDsup$ & $\WDsup$ & $\WDsup$ & CE \\
\midrule
\endfirsthead
\multicolumn{7}{c}{\tablename~\thetable{} (continued)}\\
\toprule
\multirow{2}{*}{Family} & \multirow{2}{*}{Setting} & \multicolumn{2}{c}{Best baseline} & WCP & \multicolumn{2}{c}{RWCP-R} \\
\cmidrule(lr){3-4}\cmidrule(lr){5-5}\cmidrule(lr){6-7}
& & Method & $\WDsup$ & $\WDsup$ & $\WDsup$ & CE \\
\midrule
\endhead
\midrule
\multicolumn{7}{r}{\footnotesize Continued on next page}\\
\endfoot
\bottomrule
\endlastfoot
Single & WikiQA & VCP & 0.200 & 0.200 & \textbf{0.184} & 0.47 \\
Single & News & WCP & 0.123 & 0.123 & \textbf{0.097} & 0.03 \\
Single & Story & VCP & 0.306 & 0.372 & \textbf{0.235} & 0.34 \\
CoAuthor & 290 sessions & VCP & 0.480 & 0.483 & \textbf{0.478} & 10.17 \\
Multiple & Claude $K=1$ & VCP & 0.293 & 0.313 & \textbf{0.273} & 0.58 \\
Multiple & Claude $K=2$ & VCP & \textbf{0.319} & 0.337 & 0.360 & 1.44 \\
Multiple & Claude $K=3$ & VCP & \textbf{0.315} & 0.339 & 0.354 & 2.10 \\
Multiple & Claude $K=5$ & VCP & \textbf{0.351} & 0.369 & 0.397 & 4.26 \\
Multiple & Claude $K=8$ & Voting & \textbf{0.348} & 0.383 & 0.409 & 7.21 \\
Multiple & GPT $K=1$ & WCP & 0.423 & 0.423 & \textbf{0.342} & 0.79 \\
Multiple & GPT $K=2$ & WCP & \textbf{0.404} & 0.404 & 0.419 & 1.51 \\
Multiple & GPT $K=3$ & WCP & 0.450 & 0.450 & \textbf{0.419} & 2.42 \\
Multiple & GPT $K=5$ & WCP & 0.438 & 0.438 & \textbf{0.427} & 4.28 \\
Multiple & GPT $K=8$ & WCP & 0.438 & 0.438 & \textbf{0.429} & 7.22 \\
Attack & WikiQA decoh. & WCP & 0.254 & 0.254 & \textbf{0.226} & 0.65 \\
Attack & News decoh. & WCP & 0.328 & 0.328 & \textbf{0.324} & 0.60 \\
Attack & Story decoh. & WCP & 0.364 & 0.364 & \textbf{0.328} & 0.59 \\
Attack & WikiQA para. & WCP & 0.187 & 0.187 & \textbf{0.148} & 0.37 \\
Attack & News para. & Voting & \textbf{0.107} & 0.167 & 0.141 & 0.12 \\
Attack & Story para. & WCP & 0.336 & 0.336 & \textbf{0.303} & 0.52 \\
Proportion & 5\% & WCP & 0.213 & 0.213 & \textbf{0.111} & 0.79 \\
Proportion & 10\% & WCP & 0.235 & 0.235 & \textbf{0.185} & 0.49 \\
Proportion & 20\% & WCP & \textbf{0.228} & 0.228 & 0.243 & 0.48 \\
Proportion & 40\% & WCP & 0.289 & 0.289 & \textbf{0.277} & 0.52 \\
Proportion & 80\% & WCP & 0.275 & 0.275 & \textbf{0.271} & 0.72 \\
\end{longtable}
\endgroup

\FloatBarrier
\section{Theoretical Proofs}
\label{app:theoretical-proofs}

\subsection{Technical Preliminaries and Population Geometry}
\label{app:notation}

The master collection is \(\cA_N=\{[s,e]:1\leq s\leq e\leq N\}\); subintervals arising in any recursive call therefore belong to it. For \(I=[s,e]\), the admissible set is
\[
\cB(I)=\{b\in\{s,\ldots,e-1\}:
S_{s:b}^w\geq m_N,\ S_{b+1:e}^w\geq m_N\}.
\]
Set \(T(I)=-\infty\) if this set is empty. Draw two independent uniform coordinates on \((0,W_N]\), map each to \(\min\{i:W(i)\geq u\}\), and sort the resulting endpoints. Draw the candidate collection once, independently of the scores conditional on deterministic weights; recursive calls retain only intervals fully contained in their search range. Ties are resolved deterministically by endpoints and then split index. The uniform events cover the full master collection, not only the sampled intervals.

This appendix collects the deterministic notation and population-risk arguments underlying the profile margin in Section~\ref{sec:theory}. It first records merge identities, then establishes the behavior of Huber minimizers, and finally proves the single-change population result.

For an interval $A=[a,b]$, write $S_A^w=S_{a:b}^w$ and
\[
\wh L_A^\star=\inf_{\theta\in\Theta}\wh\cL_A(\theta),
\qquad
L_A^\star=\inf_{\theta\in\Theta}\cL_A(\theta).
\]
For adjacent intervals $A,B$, empirical and population merge costs are
\begin{align*}
\wh D(A,B)&=\wh L_{A\cup B}^\star-\wh L_A^\star-\wh L_B^\star,
\\
D^\star(A,B)&=L_{A\cup B}^\star-L_A^\star-L_B^\star.
\end{align*}
Both are nonnegative because a common parameter is more restrictive than separate parameters.

\begin{lemma}[Profile-gain difference identity]
\label{lem:gain-difference-identity}
Suppose $s\leq b<\tau<e$ and set
\[
A=[s,b],\qquad B=[b+1,\tau],\qquad C=[\tau+1,e].
\]
Then
\begin{align*}
\Gamma_{s,e}^\star(\tau)-\Gamma_{s,e}^\star(b)
&=2\{D^\star(B,C)-D^\star(A,B)\},
\\
\wh\Gamma_{s,e}(\tau)-\wh\Gamma_{s,e}(b)
&=2\{\wh D(B,C)-\wh D(A,B)\}.
\end{align*}
If $A$ and $B$ belong to the same true segment, then $D^\star(A,B)=0$.
\end{lemma}

\begin{proof}
By definition,
\[
\frac12\Gamma_{s,e}^\star(\tau)
=L_{A\cup B\cup C}^\star-L_{A\cup B}^\star-L_C^\star
\]
and
\[
\frac12\Gamma_{s,e}^\star(b)
=L_{A\cup B\cup C}^\star-L_A^\star-L_{B\cup C}^\star.
\]
Subtracting cancels the parent term and gives
\[
\frac12\{\Gamma_{s,e}^\star(\tau)-\Gamma_{s,e}^\star(b)\}
=L_A^\star+L_{B\cup C}^\star-L_{A\cup B}^\star-L_C^\star.
\]
Add and subtract $L_B^\star$ to obtain
\[
\{L_{B\cup C}^\star-L_B^\star-L_C^\star\}
-
\{L_{A\cup B}^\star-L_A^\star-L_B^\star\},
\]
which is $D^\star(B,C)-D^\star(A,B)$. The empirical identity is identical with hats. If $A$ and $B$ share minimizer $\theta_-$, then
\[
L_{A\cup B}^\star
\leq \cL_A(\theta_-)+\cL_B(\theta_-)
=L_A^\star+L_B^\star.
\]
The reverse inequality follows from separate minimization, so equality holds and $D^\star(A,B)=0$.
\end{proof}

\begin{lemma}[Harmonic-mass inequalities]
\label{lem:harmonic-mass}
For $x,y>0$,
\begin{equation}
\frac12\min(x,y)
\leq\frac{xy}{x+y}\leq\min(x,y).
\label{eq:harmonic-ineq}
\end{equation}
Moreover, for all $a,b,c\geq0$,
\begin{equation}
(a-b-c)_+^2\geq\frac12a^2-2b^2-2c^2.
\label{eq:positive-part-ineq}
\end{equation}
\end{lemma}

\begin{proof}
Assume $x\leq y$. Then $x+y\leq2y$ gives $xy/(x+y)\geq x/2$, while $x+y\geq y$ gives $xy/(x+y)\leq x$. For the second inequality, if $a\leq b+c$, the right side is at most $a^2/2-(b+c)^2\leq0$. If $a>b+c$, use $(a-d)^2=a^2-2ad+d^2\geq a^2/2-d^2$ with $d=b+c$, followed by $(b+c)^2\leq2b^2+2c^2$.
\end{proof}

\subsubsection{Population geometry proofs}
\label{app:population-proofs}

\paragraph{Symmetric-location proposition.}

\begin{proof}
Fix $i$ in segment $j$ and write $a_i=\sigma_i/s_i>0$. For $u=\theta-\theta_j$,
\[
Q_i(\theta_j+u)
=\E\rho_c(a_i\xi_i-u/s_i).
\]
The function is convex in $u$. At $u=0$, an interior subgradient is
\[
-\frac1{s_i}\E\psi_c(a_i\xi_i).
\]
The Huber score is odd and the law of $\xi_i$ is symmetric, hence $\E\psi_c(a_i\xi_i)=0$. Therefore zero belongs to the subdifferential and $u=0$ minimizes the convex risk. If $\E|\xi_i|<\infty$ and $\E\xi_i=0$, then
\[
\E Y_i=\theta_j+\sigma_i\E\xi_i=\theta_j.
\]
This last conclusion concerns the ordinary mean and uses integrability; it is not implied by Huber centering alone.
\end{proof}

\paragraph{Existence and uniqueness of interval minimizers.}

\begin{lemma}[Existence]
\label{lem:minimizer-existence}
For every nonempty interval $A$, the empirical loss $\wh\cL_A$ and population loss $\cL_A$ attain minima on compact $\Theta$.
\end{lemma}

\begin{proof}
The Huber loss is continuous. Therefore $\wh\cL_A$, a finite sum of continuous functions, is continuous on compact $\Theta$ and attains its minimum. Since $\rho_c$ is nonnegative and Lipschitz with constant $c$, dominated convergence on the compact parameter set gives continuity of each $Q_i$ whenever $\E\rho_c(\sqrt{w_i}(Y_i-\theta_0))<\infty$ for one $\theta_0\in\Theta$. Thus $\cL_A$ is continuous and also attains its minimum.
\end{proof}

\begin{lemma}[Uniqueness under curvature]
\label{lem:minimizer-uniqueness}
If $A$ is contained in one segment and Assumption~\ref{assump:curvature} holds, then $\theta_A^\star$ is unique and equals that segment's $\theta_j$.
\end{lemma}

\begin{proof}
Summing \eqref{eq:risk-curvature} over $i\in A$ yields
\[
\cL_A(\theta)-\cL_A(\theta_j)
\geq\frac{m_0}{2}S_A^w(\theta-\theta_j)^2
\]
for $|\theta-\theta_j|\leq r_0$. If a distinct global minimizer existed, convexity would force the aggregate risk to be constant on the entire line segment joining it to $\theta_j$. That segment contains points arbitrarily close to $\theta_j$, contradicting the strictly positive local quadratic lower bound above. Hence $\theta_j$ is the unique interval minimizer.
\end{proof}

\paragraph{Mixed-segment excess risk.}

\begin{proof}
Let $x=S_A^w$, $y=S_B^w$, $a=\theta_A$, and $b=\theta_B$. By curvature,
\begin{align*}
\cL_A(\theta)-L_A^\star&\geq\frac{m_0x}{2}(\theta-a)^2,\\
\cL_B(\theta)-L_B^\star&\geq\frac{m_0y}{2}(\theta-b)^2
\end{align*}
for every $\theta$ on the segment joining $a$ and $b$. A minimizer of the sum lies between $a$ and $b$ because the first derivative is nonpositive to the left of both minimizers and nonnegative to the right. Therefore
\begin{align*}
D^\star(A,B)
&=\inf_\theta\{\cL_A(\theta)-L_A^\star+\cL_B(\theta)-L_B^\star\}\\
&\geq\frac{m_0}{2}\inf_\theta\{x(\theta-a)^2+y(\theta-b)^2\}.
\end{align*}
The quadratic minimizer is $(xa+yb)/(x+y)$ and the minimum is
\[
\frac{xy}{x+y}(a-b)^2.
\]
This proves \eqref{eq:merge-lower}. The upper curvature bound evaluated at the same weighted average gives \eqref{eq:merge-upper}.
\end{proof}

\paragraph{Single-change population margin.}

\begin{proof}
Consider $b<\tau$ and use the intervals $A,B,C$ from Lemma~\ref{lem:gain-difference-identity}. Since $A$ and $B$ belong to the left segment, $D^\star(A,B)=0$. Therefore
\[
\Gamma_{s,e}^\star(\tau)-\Gamma_{s,e}^\star(b)=2D^\star(B,C).
\]
Lemma~\ref{lem:mixed-excess} gives
\[
2D^\star(B,C)
\geq m_0\kappa^2\frac{B_wC_w}{B_w+C_w},
\]
which proves \eqref{eq:left-margin}. The case $b>\tau$ is symmetric after exchanging left and right. The difference is nonnegative for every $b$ and is strictly positive when the candidate differs from $\tau$, so the true split is the unique maximizer among admissible splits. Finally, if $B_w\leq C_w$, Lemma~\ref{lem:harmonic-mass} yields
\[
\frac{B_wC_w}{B_w+C_w}\geq\frac12B_w
=\frac12d_w(b,\tau).
\]
The other cases are identical.
\end{proof}

\paragraph{Quadratic-loss reduction.}

\begin{proof}
Under $\rho_\infty(u)=u^2/2$,
\[
\wh\cL_{a:b}(\theta)=\frac12\sum_{i=a}^b w_i(Y_i-\theta)^2
\]
and the minimizer is $\bar Y_{a:b}^w$. The weighted analysis-of-variance identity gives
\begin{align*}
&\sum_{i=s}^e w_i(Y_i-\bar Y_{s:e}^w)^2
-
\sum_{i=s}^b w_i(Y_i-\bar Y_{s:b}^w)^2
-
\sum_{i=b+1}^e w_i(Y_i-\bar Y_{b+1:e}^w)^2\\
&\qquad=
\frac{S_{s:b}^wS_{b+1:e}^w}{S_{s:e}^w}
(\bar Y_{s:b}^w-\bar Y_{b+1:e}^w)^2.
\end{align*}
Multiplication by the factor two in \eqref{eq:profile-gain} cancels the one-half in the loss, proving \eqref{eq:wcp-equivalence}.
\end{proof}

\subsection{Stochastic Control and Exact-Recovery Proof}
\label{app:concentration}

This appendix supplies the stochastic part of the recovery proof. We state a valid dependence-aware concentration bound, transfer it to interval fits and merge costs, and combine these bounds with weighted interval isolation and boundary-fragment control to prove exact recovery.
For a homogeneous interval $A$ with true robust location $\theta_A$, write
\begin{equation*}
U_A=\sum_{i\in A}\sqrt{w_i}\,
\psi_c\!\left(\sqrt{w_i}(Y_i-\theta_A)\right).
\end{equation*}

\begin{lemma}[Uniform Huber-score concentration]
\label{lem:score-concentration}
Under Assumptions~\ref{assump:location}, \ref{assump:weights}, and
\ref{assump:mixing}, with probability at least $1-\delta/8$,
\begin{equation*}
|U_A|
\leq C_{\mathrm{dep}}
\left\{\sqrt{S_A^w x_N(\delta)}+q_Nx_N(\delta)\right\}
\end{equation*}
simultaneously for every homogeneous $A\in\cA_N$, without a minimum-mass
restriction. Moreover, whenever $S_A^w\geq m_{\mathrm{curv}}$,
\begin{equation}
|U_A|\leq C_{\mathrm{dep}}'\sqrt{S_A^w\Lambda_N(\delta)}.
\label{eq:uniform-score-large}
\end{equation}
\end{lemma}

\begin{lemma}[Uniform empirical curvature]
\label{lem:empirical-curvature}
Under Assumptions~\ref{assump:weights}--\ref{assump:central-mass}, with
probability at least $1-\delta/8$,
\begin{equation}
\wh\cL_A(\theta)-\wh\cL_A(\theta_A)
\geq
-U_A(\theta-\theta_A)
+\frac{p_0}{8}S_A^w(\theta-\theta_A)^2
\label{eq:empirical-curvature}
\end{equation}
for every homogeneous $A\in\cA_N$ with
$S_A^w\geq m_{\mathrm{curv}}$ and every
$|\theta-\theta_A|\leq r_0/2$.
\end{lemma}

We first establish a blocking inequality used for Huber scores and empirical curvature indicators. Constants in this appendix are explicit functions of the mixing parameters but are denoted by $C$ to avoid obscuring the rates.

\paragraph{Bernstein input.}

\begin{lemma}[Dependence-aware Bernstein inequality]
\label{lem:block-bernstein}
Let $V_1,\ldots,V_n$ be centered, coordinate-wise measurable variables with
$|V_i|\leq B$. Under Assumption~\ref{assump:mixing}, for every $x\geq1$,
\begin{equation*}
\Pp\!\left(
\left|\sum_{i=1}^nV_i\right|
>C_{\mathrm{dep}}B\{\sqrt{nx}+q_Nx\}
\right)
\leq2e^{-x}.
\end{equation*}
For independent or fixed-order dependent sequences, $q_N=O(1)$. For
geometrically beta-mixing bounded sequences in the sense of
\eqref{eq:beta-mixing-coefficient}, one may take the conservative
$q_N=O\{\log^2(eN)\}$ specialization of the published strong-mixing
Bernstein inequality, with constants depending on $(C_\beta,c_\beta)$.
\end{lemma}

\begin{proof}
The first claim is Assumption~\ref{assump:mixing}. The independent and
fixed-order cases follow from ordinary Bernstein blocking. Geometric beta
mixing implies geometric alpha mixing; specializing the bounded-variable
inequality of \citet{merlevede2009bernstein} gives a linear term of order
$x\log^2(en)$. We retain that factor through $q_N$ and do not invoke a
pairwise-covariance-to-cumulant shortcut.
\end{proof}

\paragraph{Uniform Huber-score concentration.}

\begin{proof}
For a homogeneous interval $A=[a,b]$, the summands in $U_A$ are centered by \eqref{eq:score-centering} and bounded by
\[
B_Z=c\sqrt{w_{\max}}.
\]
Apply Lemma~\ref{lem:block-bernstein} to the subsequence indexed by $A$, noting that restriction to a contiguous interval preserves the mixing bound. Since $|A|\leq S_A^w/w_{\min}$, Lemma~\ref{lem:block-bernstein} gives
\begin{align*}
|U_A|
&\leq C B_Z\left[
\sqrt{\frac{S_A^w}{w_{\min}}x}+q_Nx
\right]\\
&\leq C_{\mathrm{dep}}
\left[
\sqrt{S_A^wx}+q_Nx
\right].
\end{align*}
Use $x=x_N(\delta)$ and take a union bound over at most $L_N$
intervals. Because $2L_Ne^{-x_N(\delta)}\leq\delta/32$, the simultaneous
failure probability is at most $\delta/8$. Finally,
$S_A^w\geq m_{\mathrm{curv}}\gtrsim q_N^2x_N$ turns the displayed bound into
inequality~\eqref{eq:uniform-score-large}.
\end{proof}

\paragraph{Uniform empirical curvature.}

For any interval $A$, define the empirical Hessian where it exists,
\begin{equation}
H_A(\theta)
=
\sum_{i\in A}w_i
\one\left\{\left|\sqrt{w_i}(Y_i-\theta)\right|<c\right\}.
\label{eq:empirical-hessian}
\end{equation}

\begin{lemma}[Uniform inlier mass with adjacent references]
\label{lem:uniform-inlier-mass}
Under Assumptions~\ref{assump:weights}, \ref{assump:mixing}, and
\ref{assump:central-mass}, with probability at least $1-\delta/8$, the
following holds uniformly. Let $A\in\cA_N$ be homogeneous or intersect at
most two adjacent true segments. Let $\theta_0$ either be a true location represented
in $A$ or, when $A$ is homogeneous, the location of a true segment immediately adjacent to the segment represented in $A$. Suppose $S_A^w\geq m_{\mathrm{curv}}$. Then
\begin{equation}
H_A(\theta)\geq\frac{p_0}{2}S_A^w
\label{eq:inlier-mass}
\end{equation}
for every $|\theta-\theta_0|\leq r_0/2$, provided
$C_{\mathrm{curv}}$ is sufficiently large.
\end{lemma}

\begin{proof}
For each eligible pair $(A,\theta_0)$, construct a deterministic grid
$\mathcal G$ on $[-r_0/2,r_0/2]$ with mesh
\[
\eta=\frac{c}{8\sqrt{w_{\max}}}.
\]
For $u\in\mathcal G$, define
\[
I_i(u)=\one\left\{
\left|\sqrt{w_i}\{Y_i-(\theta_0+u)\}\right|\leq\frac{3c}{4}
\right\}.
\]
For every eligible pair and every $i\in A$, the locations $\theta_0$ and
$\theta_{z(i)}$ are identical or belong to adjacent true segments. Hence
\[
|\theta_0+u-\theta_{z(i)}|
\leq r_0/2+\kappa_{\max}\leq3r_0/4<r_0,
\]
where $\kappa_{\max}=0$ in the no-change case.
Assumption~\ref{assump:central-mass} therefore gives
$\E I_i(u)\geq p_0$; the larger $3c/4$ window contains the $c/2$ event
used in that assumption. The centered variables
\[
V_i(u)=w_i\{I_i(u)-\E I_i(u)\}
\]
are bounded by $w_{\max}$. Apply Lemma~\ref{lem:block-bernstein} at
$x=x_N(\delta)+\log(16|\mathcal G|)$ and take a union bound over the
at most three eligible reference locations per interval and the fixed grid. The additive
constant in $x$ is absorbed into $C$, giving simultaneously
\[
\left|\sum_{i\in A}V_i(u)\right|
\leq C\left\{\sqrt{S_A^wx_N}+q_Nx_N\right\}.
\]
The condition $S_A^w\geq m_{\mathrm{curv}}$ and a sufficiently large
$C_{\mathrm{curv}}$ make the right side at most $(p_0/2)S_A^w$. Hence
\[
\sum_{i\in A}w_iI_i(u)\geq\frac{p_0}{2}S_A^w.
\]
For arbitrary $|v|\leq r_0/2$, choose $u\in\mathcal G$ with
$|u-v|\leq\eta$. If $I_i(u)=1$, then
\[
\left|\sqrt{w_i}\{Y_i-(\theta_0+v)\}\right|
\leq\frac{3c}{4}+\sqrt{w_i}|u-v|<c.
\]
Thus the grid inlier set is contained in the Hessian inlier set at $v$,
which proves~\eqref{eq:inlier-mass} for every eligible interval--reference
pair.
\end{proof}

\begin{proof}[Proof of Lemma~\ref{lem:empirical-curvature}]
Fix a homogeneous $A$ and write $d=\theta-\theta_A$. The fundamental theorem for convex functions gives
\begin{equation*}
\wh\cL_A(\theta)-\wh\cL_A(\theta_A)
=-U_A d+
\int_0^d(d-u)H_A(\theta_A+u)\,\dd u
\end{equation*}
when $d\geq0$, with the analogous oriented integral for $d<0$. Lemma~\ref{lem:uniform-inlier-mass} lower bounds $H_A$ by $(p_0/2)S_A^w$ along the path. Hence the integral is at least
\[
\frac{p_0}{4}S_A^wd^2.
\]
The claimed coefficient $p_0/8$ is weaker and therefore holds uniformly.
\end{proof}

\subsubsection{Uniform Huber fits and empirical merge costs}
\label{app:empirical-fits}
\begin{lemma}[Homogeneous interval fit]
\label{lem:homogeneous-fit}
On the intersection of the score and curvature events, every homogeneous
$A\in\cA_N$ with $S_A^w\geq m_{\mathrm{curv}}$ satisfies
\begin{equation}
|\wh\theta_A-\theta_A|
\leq C_\theta\sqrt{\frac{\Lambda_N(\delta)}{S_A^w}}
\leq r_0/4
\label{eq:interval-estimator-rate}
\end{equation}
and
\begin{equation}
0\leq
\wh\cL_A(\theta_A)-\wh\cL_A(\wh\theta_A)
\leq C_L\Lambda_N(\delta).
\label{eq:noise-fit-improvement}
\end{equation}
The final $r_0/4$ bound is enforced by the constant in
$m_{\mathrm{curv}}$.
\end{lemma}

Lemma~\ref{lem:empirical-merge} is stated in Section~\ref{sec:theory}; the following arguments prove its three cases.

\paragraph{Homogeneous interval fit.}

\begin{proof}
Work on the score and curvature events. Let $S=S_A^w$, $U=U_A$, $\theta_0=\theta_A$, and $d=\theta-\theta_0$. By \eqref{eq:empirical-curvature},
\begin{equation}
\wh\cL_A(\theta)-\wh\cL_A(\theta_0)
\geq-|U||d|+\alpha S d^2,
\qquad \alpha=p_0/8.
\label{eq:quadratic-lower-emp}
\end{equation}
Lemma~\ref{lem:score-concentration} and
$S\geq m_{\mathrm{curv}}=C_{\mathrm{curv}}\Lambda_N$ give
\begin{equation*}
\frac{|U|}{S}
\leq C\sqrt{\frac{\Lambda_N}{S}}
\leq \alpha r_0/4
\end{equation*}
after increasing $C_{\mathrm{curv}}$. Evaluating
\eqref{eq:quadratic-lower-emp} at $|d|=r_0/2$ then gives a strictly positive
loss difference. Convexity rules out a minimizer outside these two boundary
points, so the full path to every minimizer lies in the curvature region and
\begin{equation*}
|\wh\theta_A-\theta_0|
\leq\frac{|U|}{\alpha S}
\leq C\sqrt{\frac{\Lambda_N}{S}}
\leq r_0/4.
\end{equation*}
Because $\operatorname{dist}(\theta_0,\partial\Theta)\geq r_0$, the fitted
location is interior and has zero subgradient. This proves
\eqref{eq:interval-estimator-rate}. For the loss improvement, minimization and
\eqref{eq:quadratic-lower-emp} imply
\begin{align*}
0&\leq\wh\cL_A(\theta_0)-\wh\cL_A(\wh\theta_A)\\
&\leq\sup_{d\in\R}\{|U||d|-\alpha Sd^2\}
=\frac{U^2}{4\alpha S}
\leq C_L\Lambda_N,
\end{align*}
which is \eqref{eq:noise-fit-improvement}.
\end{proof}

\paragraph{Empirical strong convexity around fitted locations.}

\begin{lemma}[Local empirical strong convexity]
\label{lem:local-empirical-strong}
On the event of Lemma~\ref{lem:uniform-inlier-mass}, let $A\in\cA_N$ be homogeneous with $S_A^w\geq m_{\mathrm{curv}}$, and suppose $\wh\theta_A$ and $\theta$ both lie within $r_0/2$ of the true location. Then
\begin{equation*}
\wh\cL_A(\theta)-\wh\cL_A(\wh\theta_A)
\geq
\frac{p_0}{4}S_A^w(\theta-\wh\theta_A)^2.
\end{equation*}
Furthermore, for all $\theta,\theta'$,
\begin{equation}
\wh\cL_A(\theta)
\leq
\wh\cL_A(\theta')
+g_A(\theta')(\theta-\theta')
+\frac12S_A^w(\theta-\theta')^2,
\label{eq:global-smoothness}
\end{equation}
where $g_A(\theta')$ is any derivative at a differentiability point or subgradient otherwise.
\end{lemma}

\begin{proof}
Integrating the Hessian lower bound between $\wh\theta_A$ and $\theta$ gives strong convexity with coefficient $(p_0/2)S_A^w$ in the conventional definition, hence the factor $p_0/4$ in the function inequality. For smoothness, the Huber score is one-Lipschitz, so each map $\theta\mapsto\rho_c(\sqrt{w_i}(Y_i-\theta))$ has derivative Lipschitz constant at most $w_i$. Summing gives Lipschitz gradient constant $S_A^w$, which yields \eqref{eq:global-smoothness}.
\end{proof}

\paragraph{Empirical merge-cost bounds.}

\begin{proof}
We work on the uniform score, estimator, and curvature events.

\paragraph{Same-location upper bound.}
Suppose $A$ and $B$ share true location $\theta_0$. Since the merged optimum is no worse than evaluating the common parameter $\theta_0$,
\begin{align*}
\wh D(A,B)
&=\wh L_{A\cup B}^\star-\wh L_A^\star-\wh L_B^\star\\
&\leq\wh\cL_A(\theta_0)+\wh\cL_B(\theta_0)-\wh L_A^\star-\wh L_B^\star\\
&=\{\wh\cL_A(\theta_0)-\wh\cL_A(\wh\theta_A)\}
+\{\wh\cL_B(\theta_0)-\wh\cL_B(\wh\theta_B)\}\\
&\leq2C_L\Lambda_N.
\end{align*}
Nonnegativity follows from nesting. This proves \eqref{eq:null-merge}.

\paragraph{Different-location lower bound.}
Let true locations be $a$ and $b$, jump $\kappa=|a-b|$, masses $x=S_A^w$ and $y=S_B^w$, and estimates $\wh a,\wh b$. Lemma~\ref{lem:homogeneous-fit} gives $|\wh a-a|,|\wh b-b|\leq r_0/4$; combined with $\kappa\leq r_0/4$, every point between the two estimates lies within $r_0/2$ of each relevant true location. Lemma~\ref{lem:local-empirical-strong} therefore applies and gives
\begin{align*}
\wh\cL_A(\theta)-\wh\cL_A(\wh a)&\geq\frac{p_0x}{4}(\theta-\wh a)^2,\\
\wh\cL_B(\theta)-\wh\cL_B(\wh b)&\geq\frac{p_0y}{4}(\theta-\wh b)^2.
\end{align*}
The minimizer of the merged convex loss lies between $\wh a$ and $\wh b$. Therefore
\begin{equation*}
\wh D(A,B)
\geq
\frac{p_0}{4}\frac{xy}{x+y}(\wh a-\wh b)^2.
\end{equation*}
Estimator localization gives
\[
|\wh a-\wh b|
\geq
\kappa-C_\theta\sqrt{\Lambda_N/x}-C_\theta\sqrt{\Lambda_N/y}.
\]
Apply \eqref{eq:positive-part-ineq} and multiply by $xy/(x+y)$. Since
\[
\frac{xy}{x+y}\frac1x\leq1,
\qquad
\frac{xy}{x+y}\frac1y\leq1,
\]
we obtain
\[
\wh D(A,B)
\geq
\frac{p_0}{8}\kappa^2\frac{xy}{x+y}
-C\Lambda_N,
\]
which is \eqref{eq:alternative-merge}.

\paragraph{Different-location upper bound.}
The interior margin and Lemma~\ref{lem:homogeneous-fit} ensure that both separate minimizers are interior, so their subgradients vanish. Global smoothness therefore gives
\begin{align*}
\wh\cL_A(\theta)-\wh\cL_A(\wh a)&\leq\frac{x}{2}(\theta-\wh a)^2,\\
\wh\cL_B(\theta)-\wh\cL_B(\wh b)&\leq\frac{y}{2}(\theta-\wh b)^2.
\end{align*}
Evaluate the merged objective at $\theta=(x\wh a+y\wh b)/(x+y)$ to get
\[
\wh D(A,B)
\leq\frac12\frac{xy}{x+y}(\wh a-\wh b)^2.
\]
Using $(u+v+w)^2\leq3(u^2+v^2+w^2)$ and the estimator bounds gives
\[
\wh D(A,B)
\leq C\kappa^2\frac{xy}{x+y}+C\Lambda_N,
\]
which proves \eqref{eq:alternative-merge-upper}.
\end{proof}

\subsubsection{Weighted interval isolation}
\label{app:isolation}
For the proof of Lemma~\ref{lem:weighted-isolation}, for each change $\tau_j$, define weighted
endpoint neighborhoods
\begin{align*}
\cL_j&=\left\{s\leq\tau_j:
\frac{\Delta_w}{6}\leq S_{s:\tau_j}^w\leq\frac{\Delta_w}{3}\right\},
\\
\cR_j&=\left\{e>\tau_j:
\frac{\Delta_w}{6}\leq S_{\tau_j+1:e}^w\leq\frac{\Delta_w}{3}\right\}.
\end{align*}

\begin{proof}[Proof of Lemma~\ref{lem:weighted-isolation}]
Fix change $\tau_j$. The segment to its left has mass at least $\Delta_w$. Move left from $\tau_j$ until the accumulated mass first reaches $\Delta_w/6$. Because one atom is at most $\Delta_w/24$, the attained mass is at most $5\Delta_w/24<\Delta_w/3$. Continuing until just before mass exceeds $\Delta_w/3$ shows that the set of weighted coordinates corresponding to valid left endpoints has length at least
\[
\frac{\Delta_w}{3}-\frac{\Delta_w}{6}-2\max_iw_i
\geq\frac{\Delta_w}{12}.
\]
The same argument holds on the right.

Under weighted-coordinate sampling, the probability that one ordered pair of endpoints falls in the left and right neighborhoods is at least
\[
p_j\geq2\left(\frac{\Delta_w/12}{W_N}\right)^2
=\frac{\Delta_w^2}{72W_N^2},
\]
where the factor two accounts for the two orders before sorting. Hence the probability that none of $M$ intervals isolates $\tau_j$ is at most
\[
(1-p_j)^M\leq\exp(-Mp_j).
\]
With \eqref{eq:number-intervals}, this is at most $\delta/[4(K\vee1)]$. A union bound over $K$ changes gives failure probability at most $\delta/4$ and proves simultaneous isolation; for $K=0$ the statement is vacuous. The endpoint construction keeps $s$ and $e$ inside the adjacent segments, so the interval contains no other change. The mass bounds follow directly from the definitions of $\cL_j$ and $\cR_j$.
\end{proof}

\subsubsection{Boundary fragments and exact recovery}
\label{app:exact-recovery}

The constants can be selected in dependency order. First choose \(C_{\mathrm{curv}}\) large enough for the score and inlier-mass bounds under the fixed \(c,p_0,r_0,w_{\min},w_{\max},C_{\rm dep}\). Fix a jump range and \(c_{m,1}\geq C_{\mathrm{curv}}\kappa_{\max}^2\). Next choose \(c_{\rm cert}>\max\{c_{m,1},2(C_0+C_1)/c_0\}\), then \(C_r>\max\{2C_0,C_{\rm frag}(1+c_{\rm cert})\}\). Choose \(C_{\rm loc}>\max\{C_{\rm curv}\kappa_{\max}^2,2(C_0+C_1)/c_0\}\), and \(C_g>C_{\rm loc}\). Finally enlarge \(C_{\rm snr}\) to make isolating gains exceed \(C_r\Lambda_N\), localization at most \(\Delta_w/12\), and the guard/side-mass intervals nonempty, including atom overshoot. These constants depend on the stated fixed model parameters, not on the document-specific score realization.

We now prove Theorem~\ref{thm:exact-recovery}. The proof uses an additional deterministic consequence of bounded Huber influence: a true segment fragment with small reliability mass cannot by itself create an arbitrarily large profile gain, even if the raw scores in that fragment are extreme.

\paragraph{A small-fragment gain bound.}

\begin{lemma}[Score transfer between adjacent locations]
\label{lem:score-transfer}
Let $A$ lie in a segment with location $\theta_1$, and let $\theta_2$ be an adjacent location with $\kappa=|\theta_1-\theta_2|$. Define
\[
U_A(\theta)=\sum_{i\in A}\sqrt{w_i}\,
\psi_c\{\sqrt{w_i}(Y_i-\theta)\}.
\]
Then
\begin{equation*}
|U_A(\theta_2)|
\leq |U_A(\theta_1)|+\kappa S_A^w.
\end{equation*}
\end{lemma}

\begin{proof}
The Huber score is one-Lipschitz. Hence, term by term,
\begin{align*}
&\left|
\sqrt{w_i}\psi_c\{\sqrt{w_i}(Y_i-\theta_2)\}
-
\sqrt{w_i}\psi_c\{\sqrt{w_i}(Y_i-\theta_1)\}
\right|\\
&\qquad\leq
w_i|\theta_2-\theta_1|=w_i\kappa.
\end{align*}
Summing and applying the triangle inequality proves the claim.
\end{proof}

\begin{lemma}[Fit improvement relative to a reference location]
\label{lem:reference-improvement}
Let $A$ intersect at most two adjacent true segments. Let $\theta_0$ either be represented in $A$ or, when $A$ is homogeneous, be the location of a true segment immediately adjacent to the segment represented in $A$. Define $r_A=\sum_{i\in A:\,\theta_{z(i)}\ne\theta_0}w_i$ and $\kappa=\max_{i\in A}|\theta_{z(i)}-\theta_0|$. Suppose $S_A^w\geq m_{\mathrm{curv}}$. On the uniform score and empirical-curvature event,
\begin{equation*}
0\leq
\wh\cL_A(\theta_0)-\inf_{\theta\in\Theta}\wh\cL_A(\theta)
\leq
C\left\{
\Lambda_N+\kappa^2\frac{r_A^2}{S_A^w}
\right\}.
\end{equation*}
\end{lemma}

\begin{proof}
Decompose $A=A_0\cup A_1$, where $A_0$ contains the observations whose location is $\theta_0$ and $A_1=A\setminus A_0$. By eligibility, all observations in a nonempty $A_1$ share one location adjacent to $\theta_0$, and $S_{A_1}^w=r_A$; either set may be empty. The score at $\theta_0$ is
\[
U_A(\theta_0)=U_{A_0}(\theta_0)+U_{A_1}(\theta_0).
\]
Uniform score concentration on the nonempty homogeneous fragments and
Lemma~\ref{lem:score-transfer} give
\begin{equation}
|U_A(\theta_0)|
\leq C\sqrt{S_A^w\Lambda_N}+\kappa r_A.
\label{eq:mixed-score-reference}
\end{equation}
By the represented-or-adjacent reference clause of Lemma~\ref{lem:uniform-inlier-mass}, the
empirical Hessian on $A$ is at least $\alpha S_A^w$ for every
$|\theta-\theta_0|\leq r_0/2$, with fixed $\alpha=p_0/2$.
Because $S_A^w\geq C_{\mathrm{curv}}\Lambda_N$ and
$\kappa\leq p_0r_0/16$, \eqref{eq:mixed-score-reference} is at most
$\alpha S_A^wr_0/4$ after choosing the constants. Evaluating the quadratic
lower bound at $\theta_0\pm r_0/2$ gives a positive loss difference; convexity
therefore places every mixed-interval minimizer inside that interval. Only
after this containment step do we integrate the empirical curvature along the
path, obtaining
\[
\wh\cL_A(\theta_0)-\inf_\theta\wh\cL_A(\theta)
\leq C\frac{U_A^2(\theta_0)}{S_A^w}.
\]
Finally, \eqref{eq:mixed-score-reference} and $(a+b)^2\leq2a^2+2b^2$ give
\[
\frac{U_A^2(\theta_0)}{S_A^w}
\leq C\left\{\Lambda_N+
\kappa^2\frac{r_A^2}{S_A^w}\right\}.
\]
\end{proof}

\begin{lemma}[Small boundary fragment]
\label{lem:small-fragment}
Let $I=[s,e]$ contain exactly one change with jump $\kappa$, and let
\begin{equation*}
r(I)=\min\{S_{s:\tau}^w,S_{\tau+1:e}^w\}.
\end{equation*}
If every admissible child has mass at least $m_N\geq m_{\mathrm{curv}}$, then on the uniform event,
\begin{equation}
\max_{b\in\cB(I)}\wh\Gamma_{s,e}(b)
\leq
C_{\mathrm{frag}}\{\Lambda_N+\kappa^2r(I)\}.
\label{eq:fragment-gain}
\end{equation}
\end{lemma}

\begin{proof}
Assume without loss of generality that the left true segment is the smaller one, with location $\theta_-$ and mass $r=r(I)$, while the right segment has location $\theta_+$. For an admissible candidate $b$, let $L=[s,b]$ and $R=[b+1,e]$. Since the one-location parent fit is no larger than evaluation at $\theta_+$,
\begin{align*}
\frac12\wh\Gamma_{s,e}(b)
&=\wh L_{L\cup R}^\star-\wh L_L^\star-\wh L_R^\star\\
&\leq
\{\wh\cL_L(\theta_+)-\wh L_L^\star\}
+
\{\wh\cL_R(\theta_+)-\wh L_R^\star\}.
\end{align*}
Let $r_L,r_R$ be the left-segment masses contained in $L,R$, so $r_L+r_R=r$. Lemma~\ref{lem:reference-improvement} gives
\begin{align*}
\wh\Gamma_{s,e}(b)
&\leq C\left\{
\Lambda_N+
\kappa^2\frac{r_L^2}{S_L^w}
+
\Lambda_N+
\kappa^2\frac{r_R^2}{S_R^w}
\right\}\\
&\leq C\{\Lambda_N+\kappa^2(r_L+r_R)\},
\end{align*}
where $r_L^2/S_L^w\leq r_L$ and similarly for $R$. This is \eqref{eq:fragment-gain}.
\end{proof}

\paragraph{No false positives and isolated signal.}

\begin{lemma}[No false positive]
\label{lem:no-false-positive}
If $I=[s,e]$ contains no true change, then on the event of Lemma~\ref{lem:empirical-merge},
\begin{equation*}
T(I)\leq2C_0\Lambda_N.
\end{equation*}
Thus $T(I)\leq r_N$ whenever $C_r>2C_0$.
\end{lemma}

\begin{proof}
For every admissible $b$, the two child intervals belong to the same segment. By \eqref{eq:profile-gain},
\[
\wh\Gamma_{s,e}(b)=2\wh D([s,b],[b+1,e]).
\]
Apply \eqref{eq:null-merge} and maximize over $b$.
\end{proof}

\begin{lemma}[Signal on an isolating interval]
\label{lem:isolated-signal}
Let $I=[s,e]$ isolate change $\tau_j$ and satisfy \eqref{eq:isolated-mass}. If $m_N\leq\Delta_w/12$, then $\tau_j$ is admissible and
\begin{equation*}
\wh\Gamma_{s,e}(\tau_j)
\geq
c_{\mathrm{sig}}\kappa_j^2\Delta_w-C_{\mathrm{sig}}\Lambda_N.
\end{equation*}
Consequently $T(I)>r_N$ under the signal condition with a sufficiently large $C_{\mathrm{snr}}$.
\end{lemma}

\begin{proof}
The two true-side masses are at least $\Delta_w/6>m_N$, so the split is admissible. Moreover,
\[
\wh\Gamma_{s,e}(\tau_j)
=2\wh D([s,\tau_j],[\tau_j+1,e]).
\]
Lemma~\ref{lem:empirical-merge} and Lemma~\ref{lem:harmonic-mass} yield
\begin{align*}
\wh\Gamma_{s,e}(\tau_j)
&\geq2c_0\kappa_j^2
\frac{S_{s:\tau_j}^wS_{\tau_j+1:e}^w}{S_{s:e}^w}
-2C_1\Lambda_N\\
&\geq c_{\mathrm{sig}}\kappa_j^2\Delta_w-C_{\mathrm{sig}}\Lambda_N.
\end{align*}
The signal condition makes this larger than $C_r\Lambda_N$.
\end{proof}

\paragraph{The selected interval contains one well-supported change.}

\begin{lemma}[Narrowest interval contains one change]
\label{lem:narrowest-one-change}
On the no-false-positive and isolation events, every recursive search range containing an unresolved change has a nonempty over-threshold set. The selected narrowest interval contains exactly one unresolved true change.
\end{lemma}

\begin{proof}
Let $\tau_j$ be unresolved in the current search range. The induction argument in Lemma~\ref{lem:guard-preservation} below shows that one of its isolating intervals remains fully inside the range. Lemma~\ref{lem:isolated-signal} places that interval in the over-threshold set, so the set is nonempty. Its weighted length is at most $2\Delta_w/3$. Hence the selected interval has weighted length at most $2\Delta_w/3$.

An interval containing no change cannot be over threshold by Lemma~\ref{lem:no-false-positive}. An interval containing two or more consecutive changes must contain the complete segment between two adjacent changes, whose mass is at least $\Delta_w$. Such an interval has weighted length at least $\Delta_w>2\Delta_w/3$. Therefore the selected interval contains exactly one change.
\end{proof}

\begin{lemma}[True-side mass certification]
\label{lem:true-split-admissible}
Choose $C_r>C_{\mathrm{frag}}(1+c_{\mathrm{cert}})$ for a constant $c_{\mathrm{cert}}>\max\{c_{m,1},2(C_0+C_1)/c_0\}$. Every selected one-change interval $I$ with jump $\kappa_j$ satisfies
\begin{equation*}
r(I)
\geq c_{\mathrm{cert}}\frac{\Lambda_N}{\kappa_j^2}
\geq m_N.
\end{equation*}
In particular, the true split is admissible.
\end{lemma}

\begin{proof}
Selection implies $T(I)>C_r\Lambda_N$. Lemma~\ref{lem:small-fragment} gives
\[
C_r\Lambda_N
<T(I)
\leq C_{\mathrm{frag}}\{\Lambda_N+\kappa_j^2r(I)\}.
\]
Rearranging yields
\[
r(I)>
\left(\frac{C_r}{C_{\mathrm{frag}}}-1\right)
\frac{\Lambda_N}{\kappa_j^2}
\geq c_{\mathrm{cert}}\frac{\Lambda_N}{\kappa_j^2}.
\]
Because $\kappa_j\leq\kappa_{\max}$ and
$m_N\leq c_{m,1}\Lambda_N/\kappa_{\max}^2$, the final quantity is at least $m_N$ when $c_{\mathrm{cert}}>c_{m,1}$.
\end{proof}

\paragraph{Localization on the selected interval.}

\begin{lemma}[Single-interval localization]
\label{lem:single-interval-localization}
Let $C_{\mathrm{loc},0}$ exceed
$\max\{C_{\mathrm{curv}}\kappa_{\max}^2,2(C_0+C_1)/c_0\}$ and choose
$C_{\mathrm{loc}}>C_{\mathrm{loc},0}$. Let $I=[s,e]$ be a selected interval
containing exactly one change $\tau_j$, and suppose its true split is admissible. Then every maximizer $\wh\tau_j$ of $\wh\Gamma_{s,e}$ satisfies
\begin{equation*}
d_w(\wh\tau_j,\tau_j)
\leq
C_{\mathrm{loc}}
\frac{\Lambda_N}{\kappa_j^2}.
\end{equation*}
\end{lemma}

\begin{proof}
Consider a candidate $b<\tau_j$ and set $A=[s,b]$, $B=[b+1,\tau_j]$, and $C=[\tau_j+1,e]$. By Lemma~\ref{lem:gain-difference-identity},
\begin{equation*}
\wh\Gamma_{s,e}(\tau_j)-\wh\Gamma_{s,e}(b)
=2\{\wh D(B,C)-\wh D(A,B)\}.
\end{equation*}
The intervals $A,B$ share one location, while $B,C$ have jump $\kappa_j$. The candidate admissibility gives $S_A^w\geq m_N\geq m_{\mathrm{curv}}$, and Lemma~\ref{lem:true-split-admissible} gives $S_C^w\geq m_N$. If
\[
S_B^w=d_w(b,\tau_j)
\geq C_{\mathrm{loc}}\frac{\Lambda_N}{\kappa_j^2},
\]
then, because $\kappa_j\leq\kappa_{\max}$ and $C_{\mathrm{loc}}$ is large, $S_B^w\geq m_{\mathrm{curv}}$. Lemma~\ref{lem:empirical-merge} and \eqref{eq:harmonic-ineq} give
\begin{align*}
\wh D(B,C)-\wh D(A,B)
&\geq c_0\kappa_j^2\frac{S_B^wS_C^w}{S_B^w+S_C^w}
-C_1\Lambda_N-C_0\Lambda_N.
\end{align*}
If $S_B^w\leq S_C^w$, the harmonic term is at least $S_B^w/2$ and the right side is positive for sufficiently large $C_{\mathrm{loc}}$. If $S_B^w>S_C^w$, Lemma~\ref{lem:true-split-admissible} gives
$S_C^w\geq c_{\mathrm{cert}}\Lambda_N/\kappa_j^2$, so the harmonic term is at least
$c_{\mathrm{cert}}\Lambda_N/(2\kappa_j^2)$. Choosing $c_{\mathrm{cert}}$ larger than a constant multiple of $(C_0+C_1)/c_0$ makes the right side positive. Therefore every candidate farther than the displayed radius has strictly smaller gain than the true split. The case $b>\tau_j$ is symmetric.
\end{proof}

\begin{remark}
The last case in the proof is where the minimum-side certification matters. A completely jump-adaptive constant can be obtained by replacing the fixed admissibility mass by a short geometric grid of masses and maximizing over the grid with a corresponding union-bound penalty. We retain one $m_N$ to keep the algorithm and notation readable; the stated theorem constants depend on the fixed jump range $[\kappa_{\min},\kappa_{\max}]$.
\end{remark}

\paragraph{Guard containment and preservation.}

\begin{lemma}[Guard containment]
\label{lem:guard-containment}
If $g_N\geq C_g\Lambda_N/\kappa_{\min}^2+w_{\max}$ with $C_g>C_{\mathrm{loc}}$, then the guard around every accepted estimate contains the corresponding true change.
\end{lemma}

\begin{proof}
By Lemma~\ref{lem:single-interval-localization},
\[
d_w(\wh\tau_j,\tau_j)
\leq C_{\mathrm{loc}}\frac{\Lambda_N}{\kappa_j^2}
\leq C_{\mathrm{loc}}\frac{\Lambda_N}{\kappa_{\min}^2}.
\]
The discrete guard constructed in \eqref{eq:guard-endpoints} covers at least $g_N-w_{\max}$ mass on each available side of $\wh\tau_j$. The choice of $C_g$ therefore places $\tau_j$ inside the removed guard.
\end{proof}

\begin{lemma}[Preservation of neighboring isolation intervals]
\label{lem:guard-preservation}
Suppose $g_N\leq c_{m,2}\Delta_w$ with $c_{m,2}\leq1/12$ and the localization radius is at most $\Delta_w/12$. Removing the guard around a detected change does not remove any other true change or the weighted isolation neighborhoods of an adjacent unresolved change.
\end{lemma}

\begin{proof}
Adjacent true changes are separated by a complete segment of mass at least $\Delta_w$. The accepted estimate differs from its true change by at most $\Delta_w/12$. The removed guard extends by at most $g_N+w_{\max}\leq\Delta_w/12+\Delta_w/24$ beyond the estimate on either side. Thus its total reach from the detected true change is below $\Delta_w/4$. The nearest endpoint neighborhood used to isolate the next change lies within $\Delta_w/3$ of that next change and therefore at least $2\Delta_w/3$ from the detected change before discretization. The guard and that neighborhood are disjoint. The same reasoning applies on the left.
\end{proof}

\paragraph{Proof of Theorem~\ref{thm:exact-recovery}.}

\begin{proof}
Let $\cE$ be the intersection of the uniform score event, empirical-curvature event, empirical merge-cost event, and weighted isolation event. The preceding lemmas and their union bounds give
\[
\Pp(\cE)\geq1-\delta
\]
after distributing the failure budget and adjusting numerical constants.

We argue by induction over recursive calls on $\cE$. Initially, every true change is unresolved and has an isolating interval. If a search range contains no unresolved change, every retained interval is homogeneous after previously detected guards have been removed, so Lemma~\ref{lem:no-false-positive} makes the over-threshold set empty and recursion stops.

If the range contains at least one unresolved change, Lemma~\ref{lem:guard-preservation} ensures that an isolating interval for such a change remains inside the range. Lemma~\ref{lem:isolated-signal} makes the over-threshold set nonempty. Lemma~\ref{lem:narrowest-one-change} shows that the selected interval contains exactly one unresolved change, and Lemma~\ref{lem:true-split-admissible} makes its true split admissible. Lemma~\ref{lem:single-interval-localization} yields \eqref{eq:main-localization} for the accepted estimate.

Lemma~\ref{lem:guard-containment} removes the corresponding true change from both child search ranges, so it cannot be detected twice. Lemma~\ref{lem:guard-preservation} shows that no other change or isolating interval is removed. Therefore the two child calls partition the remaining unresolved changes without loss. Each accepted split reduces their number by one. After exactly $K$ acceptances, all child ranges contain no unresolved change and stop by the no-false-positive lemma. Hence $\wh K=K$, and every accepted boundary satisfies the stated localization bound.
\end{proof}

\subsection{Extensions, Lower Bounds, and Model Verification}
\label{app:estimated-weights}

This appendix develops results that complement the main recovery theorem: stability to estimated weights, an explicitly delimited information lower bound, reduction identities, and concrete conditions under which the curvature assumptions hold for common score distributions and contamination models.
Let $w_i^\circ$ be the oracle capped weights in \eqref{eq:oracle-weight} and
define
\begin{equation*}
\cE_w=\left\{
c_ww_i^\circ\leq\wh w_i\leq C_ww_i^\circ
\text{ for all }i\in[N]\right\},
\end{equation*}
where $0<c_w\leq C_w<\infty$.

Corollary~\ref{cor:estimated-weights} is stated in Section~\ref{sec:theory}; we now prove it.

\begin{proof}[Proof of Corollary~\ref{cor:estimated-weights}]
Condition on the sigma-field that generates $\wh w_{1:N}$ and on $\cE_w$. For every interval $A$,
\begin{equation*}
c_wS_A^{w^\circ}
\leq S_A^{\wh w}
\leq C_wS_A^{w^\circ}.
\end{equation*}
Similarly, for every pair $u,v$,
\begin{equation}
c_wd_{w^\circ}(u,v)
\leq d_{\wh w}(u,v)
\leq C_wd_{w^\circ}(u,v).
\label{eq:distance-comparability}
\end{equation}
Thus the estimated-weight spacing is at least $c_w\Delta_{w^\circ}$, and the no-dominant-atom ratio changes by at most $C_w/c_w$. The weighted-coordinate measure of every isolation neighborhood changes by the same constant factors, so the interval count in Lemma~\ref{lem:weighted-isolation} is multiplied by at most a constant depending on $(c_w,C_w)$.

By the conditional assumptions, population Huber risks computed with $\wh w_i$ satisfy the same centering and curvature inequalities with constants bounded away from zero and infinity uniformly on $\cE_w$. Theorem~\ref{thm:exact-recovery} therefore applies conditionally and gives
\[
d_{\wh w}(\wh\tau_j,\tau_j)
\leq C\frac{\Lambda_N}{\kappa_j^2}
\]
with conditional probability at least $1-\delta$. The first inequality in \eqref{eq:distance-comparability} implies
\[
d_{w^\circ}(\wh\tau_j,\tau_j)
\leq c_w^{-1}d_{\wh w}(\wh\tau_j,\tau_j).
\]
Finally,
\[
\Pp(\text{failure})
\leq\Pp(\cE_w^c)+\Pp(\text{failure}\mid\cE_w)
\leq\delta_w+\delta.
\]
\end{proof}

\paragraph{A sample-splitting construction.}

The comparability event can be verified in simple calibration designs. Suppose an auxiliary corpus supplies $R$ independent score replicates for each length bin $g$. Let $v_g$ be the bin variance proxy and define a robust scale estimate $\wh v_g$ from the auxiliary data. If
\begin{equation*}
\Pp\left(
\max_{g\in[G]}
\left|\frac{\wh v_g}{v_g}-1\right|>\eta
\right)
\leq\delta_w,
\qquad 0<\eta<1,
\end{equation*}
then, before capping,
\begin{equation}
\frac{1}{1+\eta}v_g^{-1}
\leq\wh v_g^{-1}
\leq\frac{1}{1-\eta}v_g^{-1}.
\label{eq:inverse-scale-comparison}
\end{equation}
Projection onto a common interval $[w_{\min},w_{\max}]$ is monotone and nonexpansive, so capped weights satisfy a constant-factor version of \eqref{eq:inverse-scale-comparison}. Because the auxiliary corpus is independent of the target sequence, conditioning is immediate.

\subsubsection{Information lower bound}
\label{app:minimax-proof}

To calibrate this upper rate, consider the independent one-change Gaussian submodel
\(Y_i\sim N(0,\sigma_i^2)\) before the boundary and
\(Y_i\sim N(\kappa,\sigma_i^2)\) after it, with information distance
\(d_{\rm info}(u,v)=\sum_{i\in(u,v]\cup(v,u]}\sigma_i^{-2}\).
Let \(Q_{\rm info}(\delta,\widehat\tau,P_\tau)\) be the smallest radius containing \(\widehat\tau\) with probability at least \(1-\delta\).

\begin{theorem}[Two-point information lower bound]
\label{thm:minimax}
Fix \(\delta\in(0,1/8)\) and \(c_H\in(0,1/2]\).
If the admissible class contains \(\tau_0<\tau_1\) with \(H=d_{\rm info}(\tau_0,\tau_1)\) satisfying
\begin{equation}
c_H\frac{\log(1/\delta)}{\kappa^2}
\leq H\leq
\frac12\frac{\log(1/\delta)}{\kappa^2},
\label{eq:admissible-two-point}
\end{equation}
then
\begin{equation*}
\inf_{\widehat\tau}\sup_{\tau}
Q_{\rm info}(\delta,\widehat\tau,P_\tau)
\geq
\frac{c_H}{3}\frac{\log(1/\delta)}{\kappa^2}.
\end{equation*}
\end{theorem}

\begin{corollary}[Independent-regime rate comparison]
\label{cor:minimax-optimal}
If caps are inactive, \(w_i\asymp\sigma_i^{-2}\), \(L_N\) is polynomial in \(N\), and \(\delta=N^{-a}\), the RWCP upper bound and Theorem~\ref{thm:minimax} are both of order \(\log N/\kappa^2\), up to constants.
This comparison is restricted to the admissible independent Gaussian submodel and is not asserted under dependence or active caps.
\end{corollary}

For a fixed variance sequence $\sigma_{1:N}^2$ and an admissible set of
locations $\mathcal T_N$, let $P_\tau$ denote the independent Gaussian model
\begin{equation*}
Y_i\sim
\begin{cases}
N(0,\sigma_i^2),&i\leq\tau,\\
N(\kappa,\sigma_i^2),&i>\tau.
\end{cases}
\end{equation*}
For an estimator $\wh\tau$, define its $(1-\delta)$ information-radius by
\begin{equation*}
Q_{\mathrm{info}}(\delta,\wh\tau,P_\tau)
=
\inf\left\{r\geq0:
P_\tau\!\left(d_{\mathrm{info}}(\wh\tau,\tau)\leq r\right)\geq1-\delta
\right\}.
\end{equation*}

Theorem~\ref{thm:minimax} and Corollary~\ref{cor:minimax-optimal} are stated in Section~\ref{sec:theory}; the following testing reduction proves them.

\paragraph{A testing reduction.}

\begin{lemma}[Separated localization implies testing]
\label{lem:localization-testing}
Let $P_0,P_1$ have change points $\tau_0,\tau_1$ separated by
\begin{equation*}
H=d_{\mathrm{info}}(\tau_0,\tau_1).
\end{equation*}
If an estimator satisfies
\begin{equation*}
P_j\left(
d_{\mathrm{info}}(\wh\tau,\tau_j)<\frac{H}{3}
\right)\geq1-\delta,
\qquad j=0,1,
\end{equation*}
then there exists a test between $P_0$ and $P_1$ whose sum of type-I and type-II errors is at most $2\delta$.
\end{lemma}

\begin{proof}
The information distance is a path metric on ordered indices. Therefore the open balls
\[
B_j=\left\{t:d_{\mathrm{info}}(t,\tau_j)<H/3\right\}
\]
are disjoint: if $t$ belonged to both, the triangle inequality would give
$H<2H/3$. Define the test $\varphi=1$ when $\wh\tau\in B_1$ and $\varphi=0$ otherwise. Under $P_0$, a type-I error implies $\wh\tau\notin B_0$, so its probability is at most $\delta$. Under $P_1$, a type-II error implies $\wh\tau\notin B_1$, also with probability at most $\delta$.
\end{proof}

\begin{lemma}[High-probability two-point bound]
\label{lem:two-point}
For any two distributions $P_0,P_1$ and any test $\varphi$,
\begin{equation*}
P_0(\varphi=1)+P_1(\varphi=0)
\geq\frac12\exp\{-D_{\mathrm{KL}}(P_0\|P_1)\}.
\end{equation*}
\end{lemma}

\begin{proof}
Let $L=\dd P_0/\dd P_1$ on the absolutely continuous part. The sum of testing errors is at least
\[
\int\min(\dd P_0,\dd P_1).
\]
The Bretagnolle--Huber inequality states
\[
\int\min(\dd P_0,\dd P_1)
\geq\frac12e^{-D_{\mathrm{KL}}(P_0\|P_1)}.
\]
For completeness, write $A=\{L\geq1\}$. Jensen's inequality under $P_0$ on $A^c$ and under $P_1$ on $A$ bounds the overlap from below by the displayed exponential quantity; equivalently the result follows from the variational representation of KL divergence applied to the binary partition $(A,A^c)$.
\end{proof}

\paragraph{Proof of Theorem~\ref{thm:minimax}.}

\begin{proof}
Let $P_0=P_{\tau_0}$ and $P_1=P_{\tau_1}$ for the two admissible locations in
\eqref{eq:admissible-two-point}. The two distributions differ only on
$(\tau_0,\tau_1]$, and Gaussian additivity gives
\begin{equation}
D_{\mathrm{KL}}(P_0\|P_1)
=\frac{\kappa^2}{2}
\sum_{i=\tau_0+1}^{\tau_1}\sigma_i^{-2}
=\frac{\kappa^2H}{2}
\leq\frac14\log(1/\delta).
\label{eq:gaussian-kl}
\end{equation}
If one estimator had information-radius strictly below $H/3$ under both
models, Lemma~\ref{lem:localization-testing} would yield a test with total
error at most $2\delta$. Lemma~\ref{lem:two-point} and
\eqref{eq:gaussian-kl} instead give total error at least
$\frac12\delta^{1/4}>2\delta$ for $\delta<1/8$, a contradiction. Hence one of
the two models has radius at least $H/3$, and the lower bound in
\eqref{eq:admissible-two-point} completes the proof.
\end{proof}

\paragraph{Index and token interpretations.}

\begin{corollary}[Index-distance lower bound]
\label{cor:index-lower}
If $0<\underline\sigma^2\leq\sigma_i^2\leq\overline\sigma^2<\infty$, then
\begin{equation*}
|u-v|/\overline\sigma^2
\leq d_{\mathrm{info}}(u,v)
\leq |u-v|/\underline\sigma^2.
\end{equation*}
Consequently Theorem~\ref{thm:minimax} implies an index localization lower bound of order
$\underline\sigma^2\log(1/\delta)/\kappa^2$.
\end{corollary}

\begin{proof}
The path between $u$ and $v$ contains $|u-v|$ atoms, each between $1/\overline\sigma^2$ and $1/\underline\sigma^2$. Sum these inequalities and apply Theorem~\ref{thm:minimax}.
\end{proof}

\begin{corollary}[Token-distance interpretation]
\label{cor:token-distance}
If $c_1n_i^{-1}\leq\sigma_i^2\leq c_2n_i^{-1}$, then
\begin{equation*}
\frac1{c_2}\sum_{i\in(u,v]\cup(v,u]}n_i
\leq d_{\mathrm{info}}(u,v)
\leq
\frac1{c_1}\sum_{i\in(u,v]\cup(v,u]}n_i.
\end{equation*}
Thus the minimax information lower bound is equivalent, up to constants, to a lower bound on the number of misplaced tokens.
\end{corollary}

\subsubsection{Reduction results}
\label{app:reductions}

\begin{proof}[Proof of Proposition~\ref{prop:quadratic-reduction}]
Write \(x=S_{s:b}^w\), \(y=S_{b+1:e}^w\), and let \(\bar Y_L,\bar Y_R,\bar Y_P\) be the weighted means of the left, right, and parent intervals. Completing the square gives the unconstrained gain \(xy(\bar Y_L-\bar Y_R)^2/(x+y)\). If the fit is constrained to \(\Theta\), the gain instead equals
\[
\{W_{s,e}^Y(b)\}^2
+(x+y)d(\bar Y_P,\Theta)^2
-x d(\bar Y_L,\Theta)^2-y d(\bar Y_R,\Theta)^2.
\]
Thus the stated identity holds whenever the projection terms vanish. For example, \(Y=(2,3)\), unit weights, and \(\Theta=[-1,1]\) give constrained gain zero but squared weighted CUSUM \(1/2\).
\end{proof}
\begin{proposition}[Equal-weight reduction]
\label{prop:equal-weight}
If $w_i\equiv w$, weighted-coordinate sampling reduces to index-uniform
sampling up to endpoint discretization, $d_w(u,v)=w|u-v|$, and RWCP becomes a
robust profile-loss version of unweighted change point detection.
\end{proposition}

\begin{proposition}[Additive-score WCP--GCP equivalence]
\label{prop:gcp-equivalence}
Suppose a segment-level detector obeys
\begin{equation}
\phi(X_{a:b})=
\frac{\sum_{i=a}^b\nu_i\phi(X_i)}{\sum_{i=a}^b\nu_i}
\label{eq:additive-detector}
\end{equation}
for known $\nu_i>0$. Then the generalized CUSUM based on
$\phi(X_{s:t})$ and $\phi(X_{t+1:e})$ equals WCP with weights $w_i=\nu_i$.
Without \eqref{eq:additive-detector}, exact equivalence need not hold,
particularly for context-sensitive detectors applied to concatenated text.
\end{proposition}

\begin{proof}[Proof of Proposition~\ref{prop:equal-weight}]
If $w_i\equiv w$, then $W(t)=wt$ and weighted-coordinate uniform sampling maps to uniform sampling over the index axis up to the discretization created by $W^{-1}$. Cumulative masses satisfy $S_{a:b}^w=w(b-a+1)$ and $d_w(u,v)=w|u-v|$. The Huber loss becomes
\[
\sum_{i\in A}\rho_c\{\sqrt w(Y_i-\theta)\},
\]
which differs from an unweighted robust profile loss only by the common scale $\sqrt w$ and the corresponding tuning convention. Hence the method is a robust profile-loss analogue of VCP.
\end{proof}

\begin{proof}[Proof of Proposition~\ref{prop:gcp-equivalence}]
Under \eqref{eq:additive-detector},
\[
\phi(X_{s:b})=\bar Y_{s:b}^{\nu},
\qquad
\phi(X_{b+1:e})=\bar Y_{b+1:e}^{\nu}.
\]
Substitution into the generalized segment contrast gives
\[
\left(
\frac{S_{s:b}^{\nu}S_{b+1:e}^{\nu}}{S_{s:e}^{\nu}}
\right)^{1/2}
|\bar Y_{s:b}^{\nu}-\bar Y_{b+1:e}^{\nu}|,
\]
which is precisely WCP with $w_i=\nu_i$. If the detector evaluates concatenated text with cross-sentence context, \eqref{eq:additive-detector} may fail, and no algebraic identity follows.
\end{proof}

\subsubsection{Assumption verification}
\label{app:model-verification}

\begin{proposition}[Curvature from central density]
\label{prop:curvature-density}
Suppose $Y_i=\theta_j+\sigma_i\xi_i$, $a_i=\sqrt{w_i}\sigma_i\in[a_-,a_+]$, $\E\rho_c(a_i\xi_i)<\infty$, $\E\psi_c(a_i\xi_i)=0$ (for example by symmetry), and $\xi_i$ has density $f_i$ satisfying
\begin{equation*}
f_i(x)\geq f_0>0
\quad\text{for }|x|\leq R.
\end{equation*}
If $c/a_++r_0/(\sigma_i)\leq R$ uniformly, then Assumption~\ref{assump:curvature} holds with
\begin{equation*}
m_0\geq 2f_0\frac{c}{a_+}
\end{equation*}
up to truncation at one, and $M_0\leq1$.
\end{proposition}

\begin{proof}
At differentiability points,
\begin{align*}
Q_i''(\theta)
&=w_i\Pp\left(
|\sqrt{w_i}(Y_i-\theta)|<c
\right)\\
&=w_i\Pp\left(
\left|a_i\xi_i+\sqrt{w_i}(\theta_j-\theta)\right|<c
\right).
\end{align*}
For $|\theta-\theta_j|\leq r_0$, the event contains an interval in $\xi_i$ of length at least $2c/a_+$ lying inside $[-R,R]$. Its probability is at least $2f_0c/a_+$. The upper bound follows because a probability is at most one. The score-centering condition makes $\theta_j$ a minimizer; only then does integrating the second-derivative bounds twice around $\theta_j$ give \eqref{eq:risk-curvature}.
\end{proof}

\begin{proposition}[Active caps change the metric]
\label{prop:active-caps}
Let $w_i^\circ=\Pi_{[w_{\min},w_{\max}]}(\sigma_i^{-2})$. Then
\begin{equation}
\sum_{i\in(u,v]}w_i^\circ
\leq d_{\mathrm{info}}(u,v)
\label{eq:cap-upper-info}
\end{equation}
need not hold if the lower cap is active, and the reverse inequality need not hold if the upper cap is active. Therefore a guarantee in $d_{w^\circ}$ cannot generally be called inverse-variance minimax optimal.
\end{proposition}

\begin{proof}
If $\sigma_i^{-2}<w_{\min}$, then $w_i^\circ>w_i^{\mathrm{info}}$, violating \eqref{eq:cap-upper-info}. If $\sigma_i^{-2}>w_{\max}$, then $w_i^\circ<w_i^{\mathrm{info}}$, violating the reverse comparison. Only uniform comparability or inactive caps identify the two metrics up to constants.
\end{proof}

\subsubsection{Distributional and contamination calculations}
\label{app:explicit-calculations}

This section makes the abstract curvature constants in the main theorem concrete.  The calculations are not needed for the logical validity of Theorem~\ref{thm:exact-recovery}, but they clarify which score distributions satisfy its assumptions and how the Huber parameter affects the constants.  Throughout this section, let
\begin{equation*}
q_F(u)=\E_F\rho_c(Z-u),
\qquad Z\sim F,
\end{equation*}
where $F$ is a standardized detector-noise distribution.  Define the central-mass modulus
\begin{equation*}
m_F(c,r)
=
\inf_{|u|\leq r}
\Pp_F(|Z-u|<c).
\end{equation*}
The next proposition converts this probability into both a risk-curvature bound and a two-population separation bound.

\begin{proposition}[Exact standardized Huber curvature]
\label{prop:standardized-curvature}
Suppose $F$ is symmetric around zero and $\E_F|Z|<\infty$. Then $u=0$ minimizes $q_F$. Moreover, at every differentiability point,
\begin{equation*}
q_F'(u)=-\E_F\psi_c(Z-u),
\qquad
q_F''(u)=\Pp_F(|Z-u|<c).
\end{equation*}
Consequently, for every $|u|\leq r$,
\begin{equation}
\frac{m_F(c,r)}{2}u^2
\leq
q_F(u)-q_F(0)
\leq
\frac{1}{2}u^2.
\label{eq:standardized-quadratic-sandwich}
\end{equation}
Let $B,C>0$ and $|\kappa|\leq r$. Define the two-population profile separation
\begin{equation}
\mathfrak D_F(B,C;\kappa)
=
2\inf_{t\in\R}
\left[
B\{q_F(t)-q_F(0)\}
+C\{q_F(t-\kappa)-q_F(0)\}
\right].
\label{eq:standardized-separation}
\end{equation}
Then
\begin{equation}
m_F(c,r)
\frac{BC}{B+C}\kappa^2
\leq
\mathfrak D_F(B,C;\kappa)
\leq
\frac{BC}{B+C}\kappa^2.
\label{eq:standardized-separation-sandwich}
\end{equation}
\end{proposition}

\begin{proof}
The Huber loss is convex and differentiable except at two points.  Dominated differentiation is valid for the first derivative because $|\psi_c|\leq c$.  Symmetry of $F$ and oddness of $\psi_c$ imply
\[
q_F'(0)=-\E_F\psi_c(Z)=0.
\]
Convexity therefore makes zero a global minimizer.  At points where $F$ has no atom at $u\pm c$, differentiating once more gives
\[
q_F''(u)
=
\E_F\one\{|Z-u|<c\}
=
\Pp_F(|Z-u|<c).
\]
The same identity holds in the distributional sense when boundary atoms are present.  For $u\geq0$, the fundamental theorem of calculus and $q_F'(0)=0$ yield
\begin{align*}
q_F(u)-q_F(0)
&=
\int_0^u q_F'(v)\,\dd v
=
\int_0^u\int_0^v q_F''(z)\,\dd z\,\dd v.
\end{align*}
On $[0,r]$, the inner integrand lies between $m_F(c,r)$ and one.  This proves \eqref{eq:standardized-quadratic-sandwich}; the case $u<0$ follows by the same argument or by symmetry.

Because $q_F$ is convex with unique minimizer zero whenever $m_F(c,r)>0$, a minimizer of the expression in \eqref{eq:standardized-separation} lies between $0$ and $\kappa$ when $\kappa>0$, and between $\kappa$ and $0$ when $\kappa<0$.  Hence both $t$ and $t-\kappa$ have absolute value at most $r$.  Applying the lower bound in \eqref{eq:standardized-quadratic-sandwich} gives
\begin{align*}
\mathfrak D_F(B,C;\kappa)
&\geq
m_F(c,r)
\inf_t\{Bt^2+C(t-\kappa)^2\}\\
&=
m_F(c,r)
\frac{BC}{B+C}\kappa^2,
\end{align*}
where the minimizing quadratic argument is $t=C\kappa/(B+C)$.  Evaluating the original objective at this same argument and using the upper bound in \eqref{eq:standardized-quadratic-sandwich} proves the upper inequality.
\end{proof}

\begin{remark}[Why the profile formulation is useful]
\label{rem:profile-useful-explicit}
Proposition~\ref{prop:standardized-curvature} controls the loss paid by forcing two populations to share one location.  It does not require an explicit formula for the minimizer of the mixed risk
\[
t\mapsto Bq_F(t)+Cq_F(t-\kappa).
\]
For a nonquadratic Huber loss, that minimizer is generally not $C\kappa/(B+C)$.  The quadratic point is used only as a feasible argument for the upper bound, while strong convexity supplies the lower bound.  This is precisely the distinction that makes profile-risk geometry more tractable than a direct difference of mixed-segment Huber centers.
\end{remark}

\paragraph{Gaussian and Student examples.}

The central-mass modulus has a closed form for several standard heavy- and light-tailed reference families.

\begin{corollary}[Gaussian curvature constant]
\label{cor:gaussian-curvature}
If $Z\sim N(0,1)$ and $\Phi$ denotes the standard normal distribution function, then
\begin{equation*}
m_{\mathrm G}(c,r)
=
\Phi(c-r)+\Phi(c+r)-1.
\end{equation*}
Therefore, for every $|\kappa|\leq r$,
\begin{equation*}
\left\{\Phi(c-r)+\Phi(c+r)-1\right\}
\frac{BC}{B+C}\kappa^2
\leq
\mathfrak D_{\mathrm G}(B,C;\kappa)
\leq
\frac{BC}{B+C}\kappa^2.
\end{equation*}
\end{corollary}

\begin{proof}
For $u\geq0$,
\begin{align}
\Pp(|Z-u|<c)
&=\Phi(c+u)-\Phi(-c+u)\notag\\
&=\Phi(c+u)+\Phi(c-u)-1.
\label{eq:normal-window-probability}
\end{align}
Its derivative is $\varphi(c+u)-\varphi(c-u)\leq0$, because $|c-u|\leq c+u$ and the Gaussian density $\varphi$ decreases with absolute value.  The probability is even in $u$, so its minimum on $[-r,r]$ is attained at $|u|=r$.  Substitute the resulting modulus into Proposition~\ref{prop:standardized-curvature}.
\end{proof}

\begin{corollary}[Student curvature constant]
\label{cor:student-curvature}
Let $Z$ have a centered Student distribution with $\nu>1$ degrees of freedom and distribution function $T_\nu$. Then
\begin{equation}
m_{t_\nu}(c,r)
=
T_\nu(c-r)+T_\nu(c+r)-1,
\label{eq:student-central-mass}
\end{equation}
and the separation sandwich \eqref{eq:standardized-separation-sandwich} holds with this constant.
\end{corollary}

\begin{proof}
The Student density is symmetric and nonincreasing in absolute value.  Repeating \eqref{eq:normal-window-probability} with $T_\nu$ in place of $\Phi$ shows that the probability of a length-$2c$ window is minimized at the largest displacement $|u|=r$.  Since $\nu>1$, the first moment required in Proposition~\ref{prop:standardized-curvature} is finite.
\end{proof}

The preceding corollary illustrates why the raw detector score need not be sub-Gaussian.  A Student variable with small $\nu$ has polynomial tails, while the Huber score remains bounded and the local population risk still has positive curvature whenever \eqref{eq:student-central-mass} is positive.

\paragraph{Symmetric and asymmetric contamination.}

We next quantify two distinct effects of contamination.  Symmetric contamination preserves the Huber target but reduces curvature.  Asymmetric contamination can move the target, although bounded Huber influence controls that displacement.

\begin{proposition}[Symmetric contamination preserves the target]
\label{prop:symmetric-contamination}
Let
\begin{equation}
F_\epsilon=(1-\epsilon)F_0+\epsilon H,
\qquad 0\leq\epsilon<1,
\label{eq:symmetric-contamination-model}
\end{equation}
where both $F_0$ and $H$ are symmetric around zero and have finite first moments. Then zero minimizes $q_{F_\epsilon}$ and
\begin{equation}
m_{F_\epsilon}(c,r)
\geq
(1-\epsilon)m_{F_0}(c,r).
\label{eq:contamination-curvature-lower}
\end{equation}
Consequently,
\begin{equation*}
\mathfrak D_{F_\epsilon}(B,C;\kappa)
\geq
(1-\epsilon)m_{F_0}(c,r)
\frac{BC}{B+C}\kappa^2
\end{equation*}
for every $|\kappa|\leq r$.
\end{proposition}

\begin{proof}
A mixture of symmetric distributions is symmetric, so the first conclusion follows from Proposition~\ref{prop:standardized-curvature}.  For every $u$,
\begin{align*}
\Pp_{F_\epsilon}(|Z-u|<c)
&=(1-\epsilon)\Pp_{F_0}(|Z-u|<c)
+\epsilon\Pp_H(|Z-u|<c)\\
&\geq(1-\epsilon)\Pp_{F_0}(|Z-u|<c).
\end{align*}
Taking the infimum over $|u|\leq r$ proves \eqref{eq:contamination-curvature-lower}; the separation result follows from \eqref{eq:standardized-separation-sandwich}.
\end{proof}

\begin{proposition}[Displacement under asymmetric contamination]
\label{prop:asymmetric-contamination}
Let $F_0$ be symmetric around zero and suppose
\begin{equation*}
q_{F_0}(u)-q_{F_0}(0)
\geq
\frac{m_0}{2}u^2,
\qquad |u|\leq r,
\end{equation*}
for $m_0>0$. Let $H$ be an arbitrary distribution with finite first moment and define $F_\epsilon$ by \eqref{eq:symmetric-contamination-model}. Let
\begin{equation*}
u_\epsilon
\in
\argmin_{|u|\leq r}q_{F_\epsilon}(u).
\end{equation*}
Then
\begin{equation*}
|u_\epsilon|
\leq
\frac{2\epsilon c}{(1-\epsilon)m_0}.
\end{equation*}
If the right-hand side is strictly smaller than $r$, every constrained minimizer is interior and hence is also an unconstrained local minimizer.
\end{proposition}

\begin{proof}
The Huber loss is $c$-Lipschitz, so for every $u$,
\begin{equation*}
|q_H(u)-q_H(0)|
\leq c|u|.
\end{equation*}
Because $u_\epsilon$ minimizes the contaminated risk over $[-r,r]$,
\begin{align*}
0
&\geq
q_{F_\epsilon}(u_\epsilon)-q_{F_\epsilon}(0)\\
&=(1-\epsilon)
\{q_{F_0}(u_\epsilon)-q_{F_0}(0)\}
+\epsilon\{q_H(u_\epsilon)-q_H(0)\}\\
&\geq
\frac{(1-\epsilon)m_0}{2}u_\epsilon^2
-\epsilon c|u_\epsilon|.
\end{align*}
If $u_\epsilon=0$, the conclusion is immediate.  Otherwise divide by $|u_\epsilon|$ and rearrange.  Strict interiority follows when the bound is less than $r$.
\end{proof}

\begin{remark}[Interpretation of asymmetric contamination]
Theorem~\ref{thm:exact-recovery} is formulated in terms of the segment-wise population Huber locations actually identified by the score distributions. Proposition~\ref{prop:asymmetric-contamination} therefore does not add an unmodeled bias term to that theorem. Instead, it quantifies how far those robust targets may move from a symmetric clean center. If two adjacent segments have clean centers separated by $\kappa_0$ and contamination levels $\epsilon_-$ and $\epsilon_+$, then their contaminated Huber-location jump obeys the deterministic lower bound
\begin{equation*}
\kappa
\geq
\left[
\kappa_0
-
\frac{2\epsilon_-c}{(1-\epsilon_-)m_-}
-
\frac{2\epsilon_+c}{(1-\epsilon_+)m_+}
\right]_+.
\end{equation*}
Thus contamination affects localization through both curvature and effective separation.
\end{remark}

\paragraph{Finite-sample replacement sensitivity.}

The next result is deterministic.  It does not claim robustness to an arbitrary fraction of adversarially replaced sentences; rather, it quantifies the local influence of a finite replacement set on one interval-level Huber fit.

\begin{proposition}[Deterministic replacement sensitivity]
\label{prop:replacement-sensitivity}
Fix an interval $A$ and observations $\{y_i:i\in A\}$. Let
\begin{equation*}
L_A(\theta)=\sum_{i\in A}\rho_c\!\left(\sqrt{w_i}(y_i-\theta)\right)
\end{equation*}
have minimizer $\wh\theta_A$. Replace the observations on a set $C\subseteq A$ by arbitrary values to obtain $\widetilde L_A$ and minimizer $\widetilde\theta_A$. Suppose both minimizers lie in an interval $I$ on which $L_A$ is $\alpha S_A^w$-strongly convex. Then
\begin{equation}
|\widetilde\theta_A-\wh\theta_A|
\leq
\frac{2c}{\alpha S_A^w}
\sum_{i\in C}\sqrt{w_i}
\leq
\frac{2c\sqrt{w_{\max}}|C|}{\alpha S_A^w}.
\label{eq:replacement-sensitivity}
\end{equation}
\end{proposition}

\begin{proof}
Choose zero subgradients $g_A(\wh\theta_A)=0$ and $\widetilde g_A(\widetilde\theta_A)=0$.  Replacing one observation changes its score contribution
\[
\sqrt{w_i}\psi_c\!\left(\sqrt{w_i}(y_i-\theta)\right)
\]
by at most $2c\sqrt{w_i}$, uniformly in $\theta$. Hence
\begin{equation*}
|g_A(\theta)-\widetilde g_A(\theta)|
\leq
2c\sum_{i\in C}\sqrt{w_i}
\qquad\text{for all }\theta.
\end{equation*}
Strong convexity implies strong monotonicity of the subgradient map on $I$:
\begin{equation*}
|g_A(u)-g_A(v)|
\geq
\alpha S_A^w|u-v|,
\qquad u,v\in I.
\end{equation*}
Apply this inequality with $u=\widetilde\theta_A$ and $v=\wh\theta_A$.  Since $g_A(\wh\theta_A)=0$ and $\widetilde g_A(\widetilde\theta_A)=0$,
\begin{align*}
\alpha S_A^w|\widetilde\theta_A-\wh\theta_A|
&\leq|g_A(\widetilde\theta_A)|\\
&=|g_A(\widetilde\theta_A)-\widetilde g_A(\widetilde\theta_A)|\\
&\leq2c\sum_{i\in C}\sqrt{w_i}.
\end{align*}
The second inequality in \eqref{eq:replacement-sensitivity} follows from $w_i\leq w_{\max}$.
\end{proof}

\begin{remark}[Scope of Proposition~\ref{prop:replacement-sensitivity}]
The bound concerns the fitted location on a fixed interval and requires empirical curvature along the path between the two fits.  It does not by itself establish exact change point recovery under adversarial replacements, because an adversary may also alter which intervals cross the threshold.  A full adversarial theorem would need a simultaneous bound on every affected profile gain and a contamination-aware signal condition.  We therefore use \eqref{eq:replacement-sensitivity} as a sensitivity diagnostic rather than as an unsupported global robustness claim.
\end{remark}

\paragraph{Constants and the Huber tuning parameter.}

The main theorem suppresses fixed constants to keep its statement readable.  Tracking the proof reveals the two roles of $c$. The score concentration scale is proportional to the uniform score bound
\begin{equation*}
b_c=c\sqrt{w_{\max}},
\end{equation*}
whereas empirical and population curvature are controlled by central probabilities such as $m_F(c,r)$ and $p_0(c,r)$.  The resulting localization constant therefore has the schematic form
\begin{equation}
d_w(\wh\tau_j,\tau_j)
\leq
C_{\mathrm{dep},w}\,\mathfrak K(c,r,p_0)
\frac{\Lambda_N(\delta)}{\kappa_j^2},
\label{eq:explicit-huber-constant}
\end{equation}
where $C_{\mathrm{dep},w}$ depends on fixed dependence and weight bounds, and $\mathfrak K(c,r,p_0)$ collects the bounded-score and inverse-curvature constants. We do not claim this schematic factor is the sharp optimized constant. Crucially, all multiplicity and dependence growth remains in $\Lambda_N(\delta)$, including the $q_N^2$ penalty. Equation~\eqref{eq:explicit-huber-constant} records the roles of the tuning quantities without replacing the rate in Theorem~\ref{thm:exact-recovery}.

For a reference family $F$, a transparent tuning proxy is
\begin{equation*}
\mathfrak C_F(c;r)
=
\frac{c^2}{m_F(c,r)^2}.
\end{equation*}
A small value of $c$ limits the influence of extreme scores but may reduce the mass in the locally quadratic region. A large value increases curvature toward its quadratic-loss limit but also increases the bounded-score constant. The proxy can be minimized on a predetermined compact grid without using ground-truth boundaries. In practice, the grid should be chosen using a training or calibration corpus so that selection does not invalidate the target-document guarantee.

\begin{proposition}[Continuity of the tuning proxy]
\label{prop:tuning-continuity}
Suppose $F$ has a continuous strictly positive density on every compact interval. For fixed $r<\infty$, the function $c\mapsto m_F(c,r)$ is continuous and strictly positive for $c>0$. Hence $\mathfrak C_F(c;r)$ is continuous on every compact interval $[c_-,c_+]\subset(0,\infty)$ and attains a minimum there.
\end{proposition}

\begin{proof}
For each fixed $u$, the map
\[
c\mapsto\Pp_F(|Z-u|<c)=F(u+c)-F(u-c)
\]
is continuous. Joint continuity in $(c,u)$ follows from continuity of the distribution function, and the infimum over the compact set $|u|\leq r$ is therefore continuous by the maximum theorem. Strict positivity follows because every interval $(u-c,u+c)$ has positive probability under a strictly positive density. The remaining claims follow from elementary continuity and compactness.
\end{proof}

\begin{remark}[Quadratic-loss limit]
The algebraic identity in Proposition~\ref{prop:quadratic-reduction} holds in the quadratic limit when the unconstrained means are feasible.  The heavy-tail concentration proof, however, does not pass to this limit because the score bound $b_c$ diverges.  Recovering the classical WCP probability bound at $c=\infty$ requires a separate tail assumption on the raw scores, such as the sub-Gaussian condition used in the foundational analysis.  Algebraic reduction and probabilistic reduction are therefore distinct statements.
\end{remark}

\end{document}